\documentclass[sigconf, screen, pagebackref, breaklinks, nonacm, table]{acmart}
\setcopyright{none}
\renewcommand\footnotetextcopyrightpermission[1]{}

\definecolor{mycolor}{cmyk}{0.1, 0.1, 0, 0}

\AtBeginDocument{%
  }

\setcopyright{acmlicensed}
\copyrightyear{2025}
\acmYear{2025}
\acmDOI{XXXXXXX.XXXXXXX}
\acmConference[MM '25]{Proceedings of the 33nd ACM International Conference on Multimedia}{October 27--31,
  2025}{Dublin, Ireland}
\acmISBN{978-1-4503-XXXX-X/2018/06}

\acmSubmissionID{2103}

\usepackage{multirow}
\usepackage{adjustbox}
\usepackage{subcaption}
\usepackage{pifont} %
\usepackage{xspace}
\usepackage[most]{tcolorbox}
\usepackage{amsmath}
\usepackage{graphicx}
\usepackage{caption}
\newcommand{\green}[1]{\textcolor{green!60!black}{#1}}

\newcommand{\ours}{\textbf{RA-Touch}\xspace}
\newcommand{\ie}{\textit{i.e.},\xspace}
\newcommand{\eg}{\textit{e.g.},\xspace}
\newcommand{\Hquad}{\hspace{0.5em}} 
 
\usepackage{listings}

\begin{document}

\title[RA-Touch: Retrieval-Augmented Touch Understanding with Enriched Visual Data]{RA-Touch: Retrieval-Augmented Touch Understanding \newline with Enriched Visual Data}

\author{Yoorhim Cho}
\authornote{Equal Contribution.}
\email{yourmejo@skku.edu}
\affiliation{
  \institution{Sungkyunkwan University}
  \country{}
}

\author{Hongyeob Kim}
\authornotemark[1]
\email{redleaf.kim@skku.edu}
\affiliation{
  \institution{Sungkyunkwan University}
  \country{}
}

\author{Semin Kim}
\email{serizard1005@g.skku.edu}
\affiliation{
  \institution{Sungkyunkwan University}
  \country{}
}

\author{Youjia Zhang}
\email{zhangyoujia@skku.edu}
\affiliation{
  \institution{Sungkyunkwan University}
  \country{}
}

\author{Yunseok Choi}
\email{ys.choi@skku.edu}
\affiliation{
  \institution{Sungkyunkwan University}
  \country{}
}

\author{Sungeun Hong}
\email{csehong@skku.edu}
\authornote{Corresponding author.}
\affiliation{
  \institution{Sungkyunkwan University}
  \country{}
}

\renewcommand{\shortauthors}{Cho et al.}

\begin{abstract}

Visuo-tactile perception aims to understand an object's tactile properties, such as texture, softness, and rigidity. However, the field remains underexplored because collecting tactile data is costly and labor-intensive. We observe that visually distinct objects can exhibit similar surface textures or material properties. For example, a leather sofa and a leather jacket have different appearances but share similar tactile properties. This implies that tactile understanding can be guided by material cues in visual data, even without direct tactile supervision.
In this paper, we introduce RA-Touch, a retrieval-augmented framework that improves visuo-tactile perception by leveraging visual data enriched with tactile semantics. We carefully recaption a large-scale visual dataset with tactile-focused descriptions, enabling the model to access tactile semantics typically absent from conventional visual datasets. A key challenge remains in effectively utilizing these tactile-aware external descriptions. RA-Touch addresses this by retrieving visual-textual representations aligned with tactile inputs and integrating them to focus on relevant textural and material properties. By outperforming prior methods on the TVL benchmark, our method demonstrates the potential of retrieval-based visual reuse for tactile understanding. Code is available at \hypersetup{urlcolor=magenta}\url{https://aim-skku.github.io/RA-Touch}

\end{abstract}



\maketitle

\section{Introduction}

\begin{figure}[t]
    \centering
        \vspace{2mm}
    \includegraphics[width=0.95\linewidth]{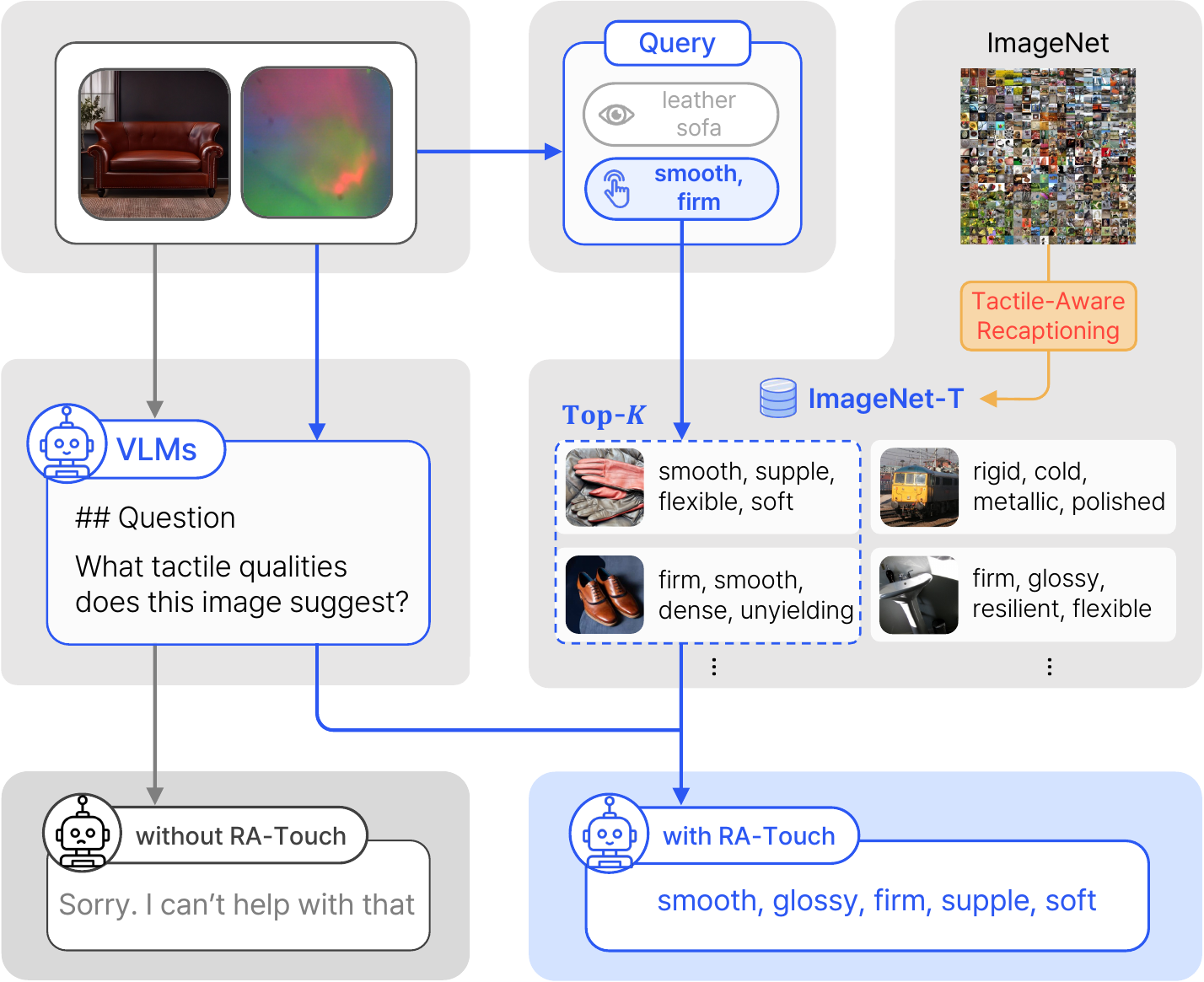}
    \caption{RA-Touch motivation.
Objects with different appearances can share similar tactile properties. RA-Touch leverages this observation by retrieving texture-relevant examples from ImageNet-T, which recaptions existing visual data with tactile-focused descriptions. This enables tactile inference without collecting additional tactile data, even when conventional VLMs fail to provide meaningful responses.
}
    \label{fig:teaser}
    \vspace{-5mm}
\end{figure}

Understanding how an object feels, such as whether it is smooth, rough, soft, or firm, requires combining multiple sensory inputs~\cite{Bresciani2006VisionAT,Stone2015TheCO,Camponogara2020IntegrationOH}. Visuo-tactile perception integrates vision and touch in a complementary way: vision provides global shape and appearance, while touch reveals local surface properties like texture and compliance~\cite{smith20203d,suresh2022shapemap, Lambeta_2020, Calandra2018MoreTA, Yuan2017GelSightHR}. Together, they offer a richer understanding of physical materials than either modality alone. This capability is critical for physical interaction in real-world applications such as robotic manipulation~\cite{Calandra2018MoreTA,chen2022visuo,Calandra2017TheFO, Lambeta_2020}, assistive systems~\cite{zhang2023visual}, and everyday tasks involving deformable or visually occluded objects~\cite{pecyna2022visual,mao2024multimodal}.

Despite its importance, visuo-tactile perception remains underexplored compared to other well-studied multimodal combinations, including vision-language~\cite{Xie2023RACLIPRA,rao2023retrieval} and vision-audio~\cite{Min2024SpeechRG,Elizalde2023CLAPLA}. Especially, while large-scale visual and language datasets have enabled rapid progress in multimodal learning, tactile data remains scarce due to the high cost and complexity of collection~\cite{Calandra2018MoreTA,Li2022SeeHA,Li2023ViHOPEVI}. Furthermore, existing visuo-tactile datasets tend to be small and biased toward specific objects and contact settings~\cite{calandra2017feeling,Li_2019_CVPR}.

Recent work has explored the potential of vision-language models (VLMs) to bridge the gap between vision and touch (\ie tactile)~\cite{Fu2024ATV, Cheng2024Touch100kAL, Yang2024BindingTT, yu2024octopi}. Trained on large image-text pairs, these models possess strong semantic priors and can describe material properties through natural language~\cite{Yang2024BindingTT,yu2024octopi}.  
Several approaches incorporate tactile supervision into VLMs using tri-modal datasets~\cite{cheng2024towards,Cheng2024Touch100kAL,Fu2024ATV}, or align tactile and visual features via auxiliary tasks. However, they still depend on annotated tactile data, which is costly and difficult to scale due to the need for physical contact and specialized sensors. 
This raises a key question: \textit{Can models acquire tactile knowledge and support visuo-tactile perception without large-scale tactile supervision? More fundamentally, is direct tactile sensing the only way to understand the texture of objects?}

To explore these questions, we rethink the role of large-scale visual corpora. Specifically, we revisit datasets like ImageNet~\cite{russakovsky2015imagenet}, examining whether they can be adapted to support tactile learning. Our key observation is that objects with visually distinct appearances can still share similar tactile properties, especially when they are made from the similar materials. For instance, a velvet cushion and a suede slipper may look quite different, yet feel remarkably similar. This insight implies that tactile learning may be feasible without triplet alignment between touch, vision, and language. This is achievable if the model can retrieve instances that are similar in tactile properties, not necessarily in appearance.

In this paper, we propose \ours, a retrieval-augmented framework for visuo-tactile perception. Our method enhances tactile understanding without requiring additional tactile data collection as shown in Figure~\ref{fig:teaser}. Instead, we retrieve visually distinct but tactilely similar examples from visual datasets enriched with descriptive language. These retrieved samples are used to refine tactile representations. Further, to bridge the gap between tactile representation and conventional vision-language data, which are often object-centric and lack material descriptions, we introduce two main modules. \textit{Tactile-Guided Retriever} generates retrieval queries guided by tactile input, helping the model retrieve samples that are aligned with how an object feels. To ensure that only relevant tactile information is integrated, \textit{Texture-Aware Integrator} modulates and fuses the retrieved features with visuo-tactile input, effectively emphasizing texture-specific cues.

To support this framework, we construct {ImageNet-T}, a tactile-centric vision-language dataset built by carefully recaptioning ImageNet~\cite{russakovsky2015imagenet} with descriptions focused on material and texture. We leverage large vision-language models~\cite{liu2023visual,Li2023BLIP2,instructblip,Achiam2023GPT,Hurst2024GPT,zhang2024llamaadapter} to generate captions that highlight tactile attributes such as softness, coarseness, rigidity, and slipperiness. This transforms a conventional visual corpus into a tactile-aware resource, providing a rich external knowledge for retrieval-driven tactile reasoning, without collecting new tactile measurements.

We validate \ours on the Touch-Vision-Language (TVL) benchmark~\cite{Fu2024ATV}, which addresses two key challenges in visuo-tactile learning:
(1) the difficulty of collecting large-scale tactile data, tackled by leveraging natural language as an alternative supervision signal,
(2) limited modality alignment, mitigated by pairing tactile inputs with visual and linguistic descriptions.  
Our method outperforms both tactile-supervised models and recent vision-language baselines, and generalizes well under data scarcity, demonstrating that retrieval-based visuo-tactile learning offers a scalable and data-efficient alternative.
We also provide comprehensive analyses of key design choices, including retrieval query formulation, caption types, and feature integration. These results can offer guidance for the underexplored but emerging domain of visuo-tactile learning.
Our main contributions are as follows:

\vspace{-2mm}
\begin{itemize}
    \item We propose \ours, a retrieval-augmented framework that improves tactile understanding using visual corpora, without relying on costly and hard-to-scale tactile data.
    \item We construct \textit{ImageNet-T}, a recaptioned visual dataset with tactile-focused descriptions, which can serve as a widely applicable resource for the underexplored field of visuo-tactile learning.
    \item We introduce \textit{Tactile-Guided Retriever} and \textit{Texture-Aware Integrator} that align external vision-language cues with tactile input to improve texture-aware representations.
\end{itemize}

\section{Related Works}

\begin{figure*}[t]
    \centering
    \includegraphics[width=1.0\linewidth]{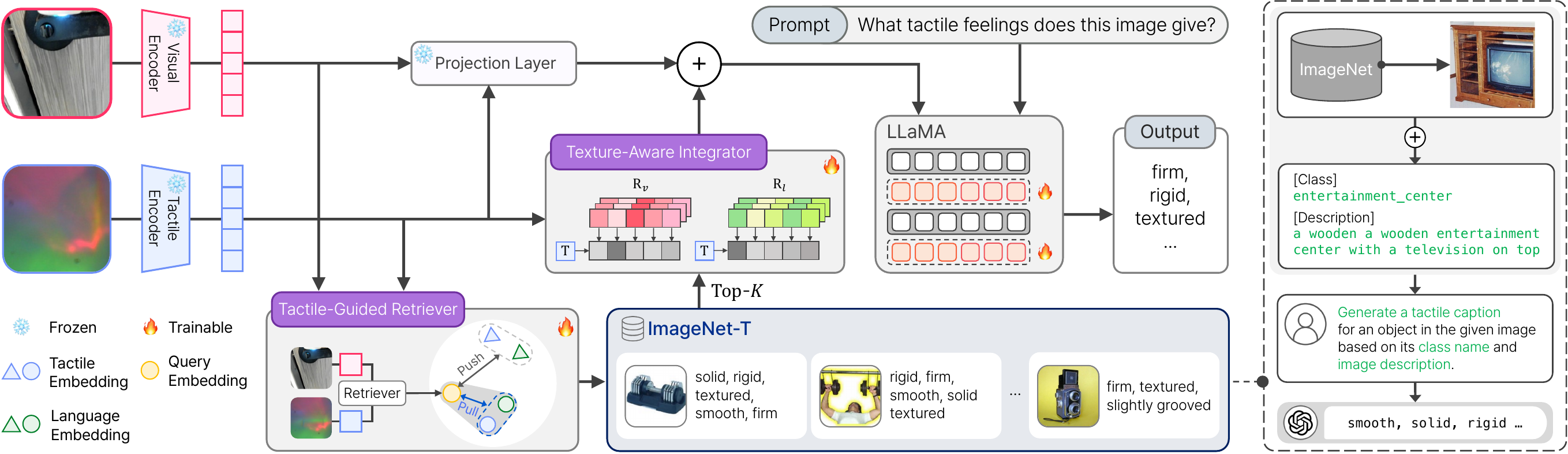}
    \caption{Overview of RA-Touch. We first construct ImageNet-T, a vision-language dataset recaptioned with tactile-focused descriptions using VLMs conditioned on the image, class name, and visual caption. Given RGB and tactile inputs, the Tactile-Guided Retriever selects the top-$K$ relevant samples from ImageNet-T based on visuo-tactile similarity. These samples are processed by the Texture-Aware Integrator, which extracts texture-relevant cues and combines them with the input tactile embedding to produce an augmented representation. This is fused with the original visual prompt to form a retrieval-augmented prompt for LLaMA, enabling tactile description generation in a parameter-efficient manner.
    }
    \label{fig:main_fig}
\end{figure*}

\subsection{Visuo-Tactile Perception}

Integrating vision and touch has become a key direction in robotics and embodied AI, inspired by their complementary roles in human perception~\cite{Bresciani2006VisionAT,Ittyerah2007MemoryFC,Jones2005ACO,Camponogara2020IntegrationOH,Stone2015TheCO}. The development of low-cost, vision-based tactile sensors~\cite{Yamaguchi2016CombiningFV,Yuan2017GelSightHR,liu2023visual,Sferrazza2019DesignMA,Shimonomura2019TactileIS} has enabled precise contact feedback and improved manipulation capabilities~\cite{Xu2024UniTUT,Li2014LocalizationAM,Li2022SeeHA}. Beyond vision and touch, audio cues have also been explored to enhance performance under complex conditions~\cite{Feng2024PlayTT}, while downstream tasks include material classification~\cite{Yang2022TouchAG}, texture recognition~\cite{Ojala2002MultiresolutionGA}, and shape reconstruction~\cite{Gao2022ObjectFolder2A}. Although recent efforts have introduced more diverse and in-the-wild tactile datasets~\cite{Yang2022TouchAG}, most continue to rely on predefined label sets and offer limited multimodal alignment. A few recent studies explore broader semantic grounding across vision, touch, and language~\cite{Yang2024BindingTT,Fu2024ATV}, but large-scale datasets that support diverse and scalable tactile understanding are still lacking.

Constructing datasets that jointly capture vision, language, and touch is challenging due to the labor-intensive nature of tactile data collection and the difficulty of aligning modalities at scale. To address this, we use existing visual datasets such as ImageNet~\cite{russakovsky2015imagenet} as an indirect source of supervision. This is the first approach that carefully recaption visual datasets into a vision-language format and retrieves tactile-relevant information to support scalable visuo-tactile learning without relying on large-scale tactile supervision.

\subsection{Multimodal Retrieval-Augmented Methods}

Growing interest in multimodal applications has led to the development of retrieval systems that operate across diverse modalities. Early work used dual-encoder architectures such as CLIP~\cite{Radford2021Learning} and CLAP~\cite{Elizalde2023CLAPLA} to align vision-language and audio-language inputs~\cite{Liu2022UniversalVD,Wei2023UniIRTA,Min2024SpeechRG}. More recent approaches use large language models to embed multimodal inputs into shared semantic spaces, moving toward universal retrieval across text, vision, and audio~\cite{BehnamGhader2024LLM2VecLL,Lee2024NVEmbedIT}.

Building on these retrieval foundations, retrieval-augmented learning has emerged as a powerful strategy for integrating external knowledge during inference. In NLP, where language models often lack access to domain-specific or up-to-date information, retrieval has been shown to improve factual accuracy, adaptability, and efficiency~\cite{Lewis2020RetrievalAugmentedGF, Guu2020REALMRA}. These advantages have encouraged its extension to multimodal domains. In vision-language tasks, retrieval has been used to enhance Visual Question Answering~\cite{Lin2023FinegrainedLM, Chen2022MuRAGMR}, image captioning~\cite{Ramos2023RetrievalaugmentedIC}, and image generation~\cite{Chen2023ReImagenRT,Blattmann2022SemiParametricNI}, by incorporating relevant image-text pairs from external corpora. Retrieval-based approaches have also been applied to the pre-training of vision-language models~\cite{Xie2023RACLIPRA}, and more recently, to temporal domains such as video understanding~\cite{Wu2024MultiMF,Thawakar2024ComposedVR} and 3D motion synthesis~\cite{Zhang2023ReMoDiffuseRA}, highlighting its versatility across modalities.

Although retrieval-based augmentation has shown strong results in language, vision, and audio, it remains underexplored in the tactile domain, largely due to the high cost of tactile data collection and the scarcity of datasets aligning touch with other modalities~\cite{Yang2022TouchAG,Yang2024BindingTT,Cheng2024Touch100kAL}. To address this, we construct ImageNet-T, a recaptioned version of ImageNet enriched with tactile-focused descriptions such as texture, compliance, and surface feel. This dataset transforms a standard visual corpus into a tactile-aware vision-language resource. Leveraging ImageNet-T, we introduce a tactile-guided retriever that uses both tactile and visual inputs to retrieve semantically aligned samples. These retrieved samples complement tactile inputs by offering texture-relevant cues that are not evident from visual appearance alone.

\subsection{Context-Aware Fusion}

Context is essential for multimodal learning. It includes not only directly observable features, but also implicit knowledge that helps models interpret and relate information across modalities. Prior work has explored context-aware mechanisms across a diverse range of tasks, including emotion recognition~\cite{Lee2019ContextAwareER}, question answering~\cite{Li2020BoostingVQ,Li2023IntentQACV,li2024context}, image captioning~\cite{zha2019context}, image-text retrieval~\cite{Zhang2020ContextAwareAN}, and image segmentation~\cite{wang2022multimodal}. These studies demonstrate that modeling contextual dependencies—whether temporal, semantic, or cross-modal—can enhance a model’s ability to reason over complex multimodal inputs. While incorporating external information can improve performance, its effectiveness depends on how well it aligns with the input. In open-set scenarios, irrelevant or mismatched retrievals can introduce noise and even harm model predictions~\cite{fang2024not}. This challenge becomes more severe in tactile settings, where visual inputs often lack detailed texture information, making models vulnerable to misleading external signals. To address this, we introduce a texture-aware integration module that filters and integrates retrieved features based on tactile input. This allows the model to focus on cues that reflect surface properties such as texture or compliance.

\section{Method}

We aim to enhance visuo-tactile perception using abundant vision-language knowledge without collecting new tactile data.  We build upon TVL-LLaMA~\cite{Fu2024ATV}, which aligns tactile representations with CLIP~\cite{ilharco_2021_5143773} vision-language embeddings and decodes them through a frozen LLaMA-2~\cite{touvron2023llama2openfoundation}. This alignment embeds touch into a shared semantic space, enabling more precise and semantically grounded tactile representation. Specifically, TVL-LLaMA takes visuo-tactile input pairs and extracts visual and tactile features using dedicated encoders. These features are summed and passed through a linear layer to produce a visual prompt embedding.
This embedding is then fed into a frozen LLaMA-2 along with a fixed textual prompt, allowing it to generate open-vocabulary descriptions of tactile properties, without being restricted to a predefined label set.

To enhance this pipeline with visual-language external knowledge, we introduce ImageNet-T, curated with texture-aware captions via a structured recaptioning process. This serves as a semantic bridge, enabling improved tactile understanding without collecting extra tactile samples. 
Figure~\ref{fig:main_fig} presents an overview of our framework based on this foundation with two key components. The tactile-guided retriever selects relevant samples from ImageNet-T using visuo-tactile cues. Then, the texture-aware integrator fuses the retrieved context with visual prompt.
This design allows the system to scale and adapt to diverse touch scenarios by leveraging vision-language knowledge for more context-aware tactile understanding.

\begin{table}
\centering
\footnotesize
\begin{tcolorbox}[colback=gray!10, colframe=black!30, width=0.9\linewidth, boxrule=0.4pt, arc=2mm]
\begin{flushleft}
\#\# Task \\
Create a tactile caption for an object in the given image based on its class name and an image description. \\
Class: \green{\texttt{\{class\_name\}}} \\
Description: \green{\texttt{\{caption\}}} \\
\vspace{0.5em}
\#\# Instructions \\
1. Provide exactly \green{5 adjectives} that refer solely to how the object feels to... \\
... \\
3. Do \green{not} include adjectives related to \green{visual
appearance}, ... or \green{any non-tactile properties}. \\
... \\
\end{flushleft}
\end{tcolorbox}
\caption{Overview of prompt used for recaptioning.}
\label{tab:main_recap_templete}
\vspace{-9mm}
\end{table}

\vspace{-4mm}
\subsection{Tactile-Enriched Image Recaptioning}

Motivated by the observation that visually distinct objects can exhibit similar tactile properties, we aim to expand the range of tactile information available in visual datasets. We use GPT-4o mini~\cite{Hurst2024GPT} to recaption existing datasets with tactile-focused descriptions, thereby addressing the limitations of datasets that primarily capture visual characteristics. Figure~\ref{fig:main_fig} (Right) illustrates the recaptioning process, and Table~\ref{tab:main_recap_templete} presents an example prompt format.

We design prompts that generate captions centered solely on tactile attributes. We exclude unrelated visual or semantic cues such as shape, color, temperature, and weight, and instead emphasize properties like texture, compliance, density, and material. Since visual information alone often fails to convey tactile characteristics, we enrich the context by providing both a class name and a descriptive caption. The combination of class name and descriptive caption offers more comprehensive grounding than either alone, enabling the model to infer tactile features more reliably. By supplying both components, we enable the model to focus on texture cues and overcome the limitations of vision-language models that emphasize appearance over tactile detail.

Once recaptioned, we extract visual features $\mathbf{V} \in \mathbb{R}^{D}$ and text features $\mathbf{L} \in \mathbb{R}^{D}$ using the TVL encoder~\cite{Fu2024ATV}, which places them in the same embedding space as tactile inputs. We set $D = 768$ for all experiments.
These features are used for similarity-based retrieval over the recaptioned dataset. Leveraging large-scale visual data in this way allows us to generate high-quality tactile labels without collecting new data, enabling scalable and cost-efficient dataset enrichment when direct tactile sensing is limited or impractical.

\subsection{Tactile-Guided Retriever}
To retrieve semantically aligned information from external knowledge, we propose a tactile-guided retrieval strategy that addresses the misalignment between visuo-tactile inputs and texture-focused vision-language data.
To this end, we construct a joint query representation that bridges tactile and visual modalities. Unimodal queries (\eg image or tactile features) often miss complementary information: image-only queries may overlook fine-grained tactile cues, while tactile-only queries may lack object-level context.
Instead, we modulate visual features with tactile input to produce a tactile-aware visual query that captures both modalities. This fused representation is used for query-to-text retrieval over the recaptioned external knowledge.

Specifically, Tactile-Guided Retriever takes both visual features $\mathbf{V} \in \mathbb{R}^{D}$ and tactile features $\mathbf{T} \in \mathbb{R}^{D}$, obtained from TVL encoders~\cite{Fu2024ATV}, to generate tactile-specific query. First, both features are passed through the multi-head self-attention (SA) to enhance intra-modality relationships. We denote the refined outputs $\mathbf{V'} \in \mathbb{R}^{D}$ and $\mathbf{T'} \in \mathbb{R}^{D}$ as follows:
\begin{align}
    &\mathbf{V'} = \mathbf{V} + \text{SA}(\mathbf{V}, \mathbf{V}, \mathbf{V}), \\
    &\mathbf{T'} = \mathbf{T} + \text{SA}(\mathbf{T}, \mathbf{T}, \mathbf{T}).
\end{align}

Then, we generate a tactile-specific query feature $\mathbf{q} \in \mathbb{R}^{D}$ through multi-head cross-attention (CA), using the refined tactile feature $\mathbf{T’}$ as the query and the refined visual features $\mathbf{V’}$ as the key and value. This design is based on the fact that tactile inputs capture only local contact regions within a global visual scene. This allows the tactile signal to selectively attend to relevant visual context and extract texture-relevant information grounded in the visual modality.
Finally, a linear projection is applied to obtain the final query $\mathcal{Q} \in \mathbb{R}^{D}$, ensuring semantic alignment with the textual embedding:
\begin{gather}
    \mathbf{q} = \text{CA}(\mathbf{T'}, \mathbf{V'}, \mathbf{V'}), \quad
    \mathcal{Q} = \mathbf{q} + \text{Linear}(\mathbf{q}).
\end{gather}

Tactile-Guided Retriever is offline-trained and applied in a frozen manner to downstream tasks without any further fine-tuning. Using this aligned query $\mathcal{Q}$, Tactile-Guided Retriever selects the top-$K$ most relevant vision-language pre-computed feature pairs $\{r_{v},r_{l}\} \in \mathbb{R}^{D}$ from the external recaptioned knowledge. Concretely, we measure the cosine similarity between the query $\mathcal{Q}$ and the text embeddings $r_{l}$ of all candidates in ImageNet-T, and retrieve the top-$K$ most similar pairs. Note that all features in external knowledge are pre-computed with the same frozen TVL encoder~\cite{Fu2024ATV}.

\subsection{Texture-Aware Integrator}

To effectively leverage the retrieved samples, we introduce a texture-aware knowledge integration module. The retrieved features lie in the CLIP embedding space, which reflects vision-language semantics and mainly encodes object-level information rather than texture-specific cues. This is because CLIP is trained on large-scale image-caption pairs that emphasize object identity or scene context over fine-grained tactile properties such as surface texture or material. Since our goal is to infer tactile attributes like softness or roughness, it is important to selectively aggregate texture-relevant information. The proposed module attends to and integrate tactile-aligned representations from the retrieved features by adaptively re-weighting them to mitigate misaligned background or object-centric representations.

Given an input tactile embedding $\mathbf{T}$ and a set of retrieved image-caption feature pairs $\{r^{k}_{v}, r^{k}_{l}\}^{K}_{k=1}$, the module applies cross-attention to extract tactile-relevant contextual features $\mathbf{a}^V \in \mathbb{R}^{D}$ and $\mathbf{a}^L \in \mathbb{R}^{D}$ from the retrieved samples. These tactile-aware features are then integrated into the prompt to enrich it with fine-grained, texture-sensitive information. Finally, the visual prompt embedding $\mathbf{p} \in \mathbb{R}^{D'}$ is computed by combining the enriched prompt with the input tactile and visual embeddings, following the fusion strategy used in TVL-LLaMA~\cite{Fu2024ATV}. Note that $D'=4096$ throughout all experiments

Concretely, we first compute two cross-attention outputs, $\mathbf{a}^V$ and $\mathbf{a}^L$, both using the input tactile embedding $\mathbf{T}$ as the query. The retrieved image features $\mathbf{R}_{v} = \{r^k_{v}\}^{K}_{k=1}$ and text features $\mathbf{R}_{l} = \{r^k_{l}\}^{K}_{k=1}$ are used as token-wise key and value for each attention, respectively. The image-based attention $\mathbf{a}^V$ is designed to aggregate tactile-relevant cuas from object-centric visual representations, enhancing material-specific signals within complex visual contexts. Meanwhile, the text-based attention $\mathbf{a}^L$ further reinforces tactile semantics embedded in the recaptioned textual descriptions. These two cross-attention steps are formulated as follows:
\begin{align}
    &\mathbf{a}^{V}= \text{CA}(\mathbf{T}, \mathbf{R}_v, \mathbf{R}_v), \\
    &\mathbf{a}^{L}= \text{CA}(\mathbf{T}, \mathbf{R}_l, \mathbf{R}_l).
\end{align}

The outputs, $\mathbf{a}^V$ and $\mathbf{a}^L$, are summed and passed through a linear projection to form a fused context representation, which integrates visual and textual cues conditioned on the tactile input. This fused representation is then further processed by a feedforward network, which has residual connection, and added to the original visual prompt embedding $\mathbf{p}$ to produce the final context-aware prompt embedding $\mathbf{p}’ \in \mathbb{R}^{D'}$ as follows:
\begin{gather}
    \mathbf{p}' = \mathbf{p} + \text{FFN}(\text{Linear}(\mathbf{a}^V + \mathbf{a}^L)).
\end{gather}

Finally, $\textbf{p}'$ is used as input to the LLaMA-2~\cite{touvron2023llama2openfoundation} to generate tactile descriptions, allowing the model to attend over visual information that has been semantically aligned with tactile context.

\subsection{Training Framework}
TVL dataset~\cite{Fu2024ATV} provides aligned tri-modal samples comprising visual, tactile, and language modalities, enabling us to supervise the Tactile-Guided Retriever and Texture-Aware Integrator with textual description targets.

Given this setup, we need to ensure that the generated query embedding $\mathcal{Q}$, which comes from the Tactile-Guided Retriever module, aligns well with the intended semantics. To achieve this, we employ a loss function composed of two parts: alignment loss and stability loss. The alignment loss $\mathcal{L}_{align}$ encourages the query to be close to the ground-truth text embedding $\mathbf{L}$. It also includes an auxiliary term that aligns the query with the associated tactile feature $\mathbf{T}$ to preserve tactile-relevant semantics. The balance between the two is controlled by a small weighting factor $\lambda_{1}$:
\begin{align}
     & \mathcal{L}_{align}= (1 - \text{sim}(\mathcal{Q}, \mathbf{L})) + \lambda_{1}\cdot (1 - \text{sim}(\mathcal{Q}, \mathbf{T})).
\end{align}

To prevent collapse to a trivial solution, which is a common failure mode in cosine similarity-based losses where representations converge to a mean embedding, we incorporate the stability loss: 
\begin{align}
   &\mathcal{L}_{stability} = \lambda_{2} \cdot \mathcal{L}_{mse} + \lambda_{3} \cdot (\mathcal{L}_{div} + \mathcal{L}_{nce}),\\
   &\mathcal{L}_{mse} = \textstyle \sum_i \|\mathcal{Q}_i-\mathbf{L}_i\|^2, \quad \mathcal{L}_{div} = \textstyle \sum_i \sum_{j \ne i} \mathcal{C}_{ij}, \\
   &\mathcal{L}_{nce} = \textstyle  -\sum_i \log \frac{\exp(\text{sim}(\mathcal{Q}_i, \mathbf{T}_i)/\tau)}{\sum_j \exp(\text{sim}(\mathcal{Q}_i, \mathbf{T}_j)/\tau)},
\end{align}
where $\mathcal{L}_{mse}$ encourages absolute alignment between the query and the ground-truth text embedding using the mean squared error. Motivated by~\cite{zbontar2021barlow,lin2023relaxing}, the second component, $\mathcal{L}_{\text{div}}$, suppresses redundancy among queries to maintain diversity by penalizing off-diagonal similarities. The last InfoNCE loss $\mathcal{L}_{nce}$ mitigates collapse while preserving consistency with tactile semantic in the CLIP embedding space, without forcing absolute similarity with it. Note that the $\text{sim}(\cdot, \cdot)$ denotes cosine similarity and query-query similarity matrix computed by simply matrix multiplication $\mathcal{C}=\mathbf{Q}^\top \mathbf{Q}$. We set $\lambda_1 = 0.2$, $\lambda_2 = 10$, and $\lambda_3 = 0.1$ in all experiments.

The final objective for the Tactile-Guided Retriever is defined as:
\begin{gather}
    \mathcal{L} = \mathcal{L}_{align} + \mathcal{L}_{stability}.
\end{gather}

After training the retriever, we freeze its parameters and train the Texture-Aware Integrator, while updating LLaMA-2~\cite{touvron2023llama2openfoundation} in a parameter-efficient manner. The integrator enriches the visual prompt $\mathbf{p}$ from TVL-LLaMA's~\cite{Fu2024ATV} projection layer with texture-aware cues, producing an augmented embedding $\mathbf{p}’$, which is then fed into LLaMA-2~\cite{touvron2023llama2openfoundation}.

Following TVL-LLaMA~\cite{Fu2024ATV}, we use a set of semantically similar prompts such as ``This image gives tactile feelings of?'' to guide the model. Based on these prompts, the language model generates open-vocabulary descriptions of object texture (\eg `soft,' `fuzzy,' `deformable'). The model is trained using a standard cross-entropy loss to maximize the likelihood of the ground-truth caption given the multimodal embedding.

\section{Experimental Setup}
\begin{table*}[ht]
  \centering
  \footnotesize
  \renewcommand{\arraystretch}{1.1}
  \begin{adjustbox}{width=0.95\textwidth}
  \begin{tabular}{
    l |
    >{\centering\arraybackslash}m{1.2cm}
    >{\centering\arraybackslash}m{1.2cm}
    >{\centering\arraybackslash}m{1.2cm}
    | >{\raggedright\arraybackslash}m{1.3cm}
    >{\raggedright\arraybackslash}m{1.3cm}
    >{\raggedright\arraybackslash}m{1.3cm}
    | >{\centering\arraybackslash}m{1.5cm}
  }
    \specialrule{1pt}{0pt}{0pt}
     & \multicolumn{3}{c|}{\textbf{Encoder Pre-training Modalities}} & \multicolumn{3}{c|}{\textbf{Score (1--10)}} & \textbf{$\boldsymbol{p}$-value} \\
    \cline{2-4} \cline{5-7}
    & Vision & Tactile & Language & \multicolumn{1}{c}{\begin{tabular}[c]{@{}c@{}}SSVTP\end{tabular}} & \multicolumn{1}{c}{\begin{tabular}[c]{@{}c@{}}HCT\end{tabular}} & \multicolumn{1}{c|}{\begin{tabular}[c]{@{}c@{}}TVL\end{tabular}} & (d.f. = 401) \\
    \specialrule{1pt}{0pt}{0pt}
    LLaVA-1.5 7B~\cite{liu2024improved} & \checkmark & -- & \checkmark & 3.64 & 3.55 & 3.56 & $1.21 \times 10^{-9}$ \\
    LLaVA-1.5 13B~\cite{liu2024improved} & \checkmark & -- & \checkmark & 3.55 & 3.63 & 3.62 & $1.49 \times 10^{-8}$ \\
    ViP-LLaVA 7B~\cite{cai2024vipllava} & \checkmark & -- & \checkmark & 2.72 & 3.44 & 3.36 & $8.77 \times 10^{-14}$ \\
    ViP-LLaVA 13B~\cite{cai2024vipllava} & \checkmark & -- & \checkmark & 4.10 & 3.76 & 3.83 & $1.72 \times 10^{-6}$ \\
    LLaMA-Adapter~\cite{zhang2024llamaadapter} & \checkmark & -- & \checkmark & 2.56 & 3.08 & 3.02 & $2.68 \times 10^{-17}$ \\
    BLIP-2 Opt-6.7B~\cite{Li2023BLIP2} & \checkmark & -- & \checkmark & 2.02 & 2.72 & 2.64 & $1.92 \times 10^{-31}$ \\
    InstructBLIP 7B~\cite{instructblip}& \checkmark & -- & \checkmark & 1.40 & 1.71 & 1.44 & $1.07 \times 10^{-84}$ \\
    InstructBLIP 13B~\cite{instructblip} & \checkmark & -- & \checkmark & 1.44 & 1.21 & 1.24 & $4.64 \times 10^{-88}$ \\
    GPT-4V~\cite{Achiam2023GPT} & \checkmark & -- & \checkmark & 5.02 & 4.42 & 4.49 & -- \\
    GPT-4-Turbo~\cite{Achiam2023GPT} & \checkmark & -- & \checkmark & 4.91 & 5.00 & 4.99 & $1.25 \times 10^{-5}$ \\
    GPT-4o~\cite{Hurst2024GPT} & \checkmark & -- & \checkmark & 4.44 & 4.59 & 4.57 & 0.4532 \\
    GPT-4o mini~\cite{Hurst2024GPT}& \checkmark & -- & \checkmark & 4.02 & 4.72 & 4.64 & 0.2101 \\
    \specialrule{0.1pt}{0pt}{0pt}
    TVL-LLaMA~\cite{Fu2024ATV} (ViT-Tiny) & \checkmark & \checkmark & \checkmark & 6.09 & 4.79 & 4.94 & $4.24 \times 10^{-5}$ \\
    \rowcolor{mycolor}
    \quad \textbf{+ RA-Touch (ImageNet-T 10k)} & \checkmark & \checkmark & \checkmark
      & 6.21 (\textcolor{blue}{+0.12})
      & 5.09 (\textcolor{blue}{+0.30})
      & 5.22 (\textcolor{blue}{+0.28}) & $1.13 \times 10^{-13}$ \\
    \rowcolor{mycolor}
    \quad \textbf{+ RA-Touch (ImageNet-T 150k)} & \checkmark & \checkmark & \checkmark
      & 6.27 (\textcolor{blue}{+0.18})
      & 5.11 (\textcolor{blue}{+0.32})
      & 5.24 (\textcolor{blue}{+0.30}) & $1.08 \times 10^{-13}$ \\
    TVL-LLaMA~\cite{Fu2024ATV} (ViT-Small) & \checkmark & \checkmark & \checkmark & 5.81 & 4.77 & 4.89 & $6.02 \times 10^{-4}$ \\
    \rowcolor{mycolor}
    \quad \textbf{+ RA-Touch (ImageNet-T 10k)} & \checkmark & \checkmark & \checkmark
      & 6.13 (\textcolor{blue}{+0.32})
      & 5.07 (\textcolor{blue}{+0.30})
      & 5.19 (\textcolor{blue}{+0.30}) & $7.52 \times 10^{-12}$ \\
    \rowcolor{mycolor}
    \quad \textbf{+ RA-Touch (ImageNet-T 150k)} & \checkmark & \checkmark & \checkmark
      & 6.21 (\textcolor{blue}{+0.40})
      & 5.13 (\textcolor{blue}{+0.36})
      & 5.26 (\textcolor{blue}{+0.37}) & $2.89 \times 10^{-13}$ \\
    TVL-LLaMA~\cite{Fu2024ATV} (ViT-Base) & \checkmark & \checkmark & \checkmark & 6.16 & 4.89 & 5.03 & $3.46 \times 10^{-6}$ \\
    \rowcolor{mycolor}
    \quad \textbf{+ RA-Touch (ImageNet-T 10k)} & \checkmark & \checkmark & \checkmark
      & 6.73 (\textcolor{blue}{+0.57})
      & 5.13 (\textcolor{blue}{+0.24})
      & 5.32 (\textcolor{blue}{+0.29}) & $2.31 \times 10^{-14}$ \\
    \rowcolor{mycolor}
    \quad \textbf{+ RA-Touch (ImageNet-T 150k)} & \checkmark & \checkmark & \checkmark
      & \textbf{6.83} (\textcolor{blue}{+0.67})
      & \textbf{5.17} (\textcolor{blue}{+0.28})
      & \textbf{5.36} (\textcolor{blue}{+0.33}) & $7.15 \times 10^{-16}$ \\
    \specialrule{1pt}{0pt}{0pt}
  \end{tabular}
  \end{adjustbox}
  \caption{TVL Benchmark Performance. Note that scores range from 1 to 10. $p$-values are two-sided paired $t$-tests comparing each model to GPT-4V~\cite{Achiam2023GPT} on the tactile-semantic task.}
  \label{tab:main_results}
  \vspace{-5mm}
\end{table*}

\subsection{Datasets and Evaluation Metric}
\noindent \textbf{Datasets.} We conduct experiments using the TVL dataset~\cite{Fu2024ATV}, which contains 43,548 visuo-tactile pairs annotated with open-vocabulary language descriptions. The dataset combines SSVTP~\cite{kerr2022ssvtp}, comprising 4,587 samples from structured robotic settings, and HCT~\cite{Fu2024ATV}, consisting of 38,961 samples collected in the wild using a DIGIT~\cite{kerr2022ssvtp} sensor. In these tactile images, color does not directly measure force. Instead, it reflects how the gel surface deforms under colored (RGB) lights, allowing the sensor to infer contact angle, depth, and shape from changes in brightness and shading. The TVL benchmark comprises 402 test samples in total, including 46 from SSVTP and 356 from HCT. Each sample consists of a visual image paired with a tactile image. Natural language annotations were obtained through a mixture of human annotation and GPT-4V-based~\cite{Achiam2023GPT} labeling. 

\vspace{0.15em}
\noindent \textbf{External Knowledge Source.} We introduce our tactile-centric external knowledge source, \textbf{ImageNet-T}, derived from the ImageNet~\cite{russakovsky2015imagenet}. Since original ImageNet lacks descriptive captions, we incorporated captions from ImageNet-1K-VL-Enriched~\cite{VisualLayer2024ImageNetEnriched}, which enhances ImageNet with captions generated by BLIP-2~\cite{Li2023BLIP2}. We observed that tactile adjectives tend to repeat frequently within object categories, causing redundancy in the dataset. To mitigate this while balancing tactile diversity with computational efficiency, we performed stratified random sampling across object categories and created several curated subsets of different sizes. These subsets optimize computational resources through the use of representative samples rather than the entire dataset.

\vspace{0.15em}
\noindent \textbf{Evaluation Metrics.} Each sample in the TVL test set consists of a visual image, a cropped visual region centered on the tactile contact point, and a corresponding tactile image. Given these inputs, the model is prompted to describe the tactile properties of the object using no more than five adjectives. To ensure consistency during inference, we use a fixed language prompt across all samples. For evaluation, we follow the protocol introduced in the TVL-LLaMA benchmark~\cite{Fu2024ATV}, which itself builds on prior works~\cite{liu2023visual,vicuna2023}. Specifically, a text-only version of GPT-4~\cite{Achiam2023GPT} is prompted to rate the similarity between the model-generated description and the human-annotated ground-truth labels. It assigns a score from 1 to 10 based on instruction adherence and semantic alignment. In addition to the numerical score, GPT-4 also provides a natural language explanation justifying its decision. This automatic evaluation setup enables scalable and interpretable comparisons across models.

\subsection{Implementation Details}
Given that vision, tactile, and language features are extracted using TVL encoder~\cite{Fu2024ATV}, where vision and language encoders are initialized from OpenCLIP~\cite{ilharco_2021_5143773} and remain frozen. All extracted features are 768-dimensional and serve as input to downstream modules unless otherwise specified. The Tactile-Guided Retriever receives these 768-dimensional visuo-tactile embedding pairs and is trained for 60 epochs with a batch size of 256. We use a learning rate of 3e-4, weight decay of 0.02, and apply a warm-up for the first 10 epochs. For experiments, we use TVL-LLaMA~\cite{Fu2024ATV} model and train the texture-aware integrator. The integrator is built on top of LLaMA-2-7B~\cite{touvron2023llama2openfoundation} with 32 LoRA-injected layers, layers are updated in a parameter-efficient manner during training. To align with LLaMA's input space, the encoder features are projected to 4096 dimensions by a learnable linear layer. Training is conducted for one epoch with a batch size of 1. The learning rate is set to 1e-3, and weight decay to 0.02. We use AdamW for optimization and trained on four NVIDIA RTX A6000 GPUs.

\vspace{-2mm}
\subsection{Experimental Results}
\noindent \textbf{Main Result.} As shown in Table~\ref{tab:main_results}, open-source vision-language models (VLMs)~\cite{cai2024vipllava, liu2023visual,Li2023BLIP2,instructblip,zhang2024llamaadapter} generally underperform compared to GPT-4V~\cite{Achiam2023GPT}, which itself struggles on tasks requiring fine-grained texture reasoning. This performance gap suggests a mismatch between the visual-centric knowledge encoded in large-scale VLMs and the type of semantic grounding required for tactile perception. In contrast, TVL-LLaMA~\cite{Fu2024ATV}, fine-tuned specifically for tactile understanding, achieves stronger performance, demonstrating the importance of tactile-aware adaptation for texture-focused tasks. Building on TVL-LLaMA, \ours further improves performance by augmenting the model with ImageNet-T retrievals, which are recaptioned to reflect tactile semantics. Notably, it achieves the highest scores across all datasets, with the ViT-Base variant showing improvements of 0.33 on TVL~\cite{Fu2024ATV}. This demonstrates the advantage of using vision-language external knowledge, recaptioned to reflect tactile semantics, for improving fine-grained texture understanding. All subsequent experiments in this section are conducted with ViT-Base backbone and ImageNet-T subset size of 10k. Additional qualitative results are provided in the supplementary Figure~\ref{fig:generation_examples}.

\newcommand{\cmark}{\ding{51}}
\newcommand{\xmark}{\ding{55}}

\begin{table}[t]
\centering
    \renewcommand{\arraystretch}{1.1} %
    \begin{adjustbox}{width=\linewidth}
        \begin{tabular}{
          >{\raggedright\arraybackslash}p{1.5cm} |
          >{\centering\arraybackslash}p{1.4cm}   
          >{\centering\arraybackslash}p{1.4cm} | 
          >{\raggedright\arraybackslash}p{1.5cm}
          >{\raggedright\arraybackslash}p{1.5cm}
          >{\raggedright\arraybackslash}p{1.5cm}
        }
            \toprule
            \textbf{Backbone} & \textbf{Retriever} & \textbf{Integrator} & \multicolumn{1}{c}{\textbf{\begin{tabular}[c]{@{}c@{}}SSVTP\end{tabular}}} & \multicolumn{1}{c}{\textbf{\begin{tabular}[c]{@{}c@{}}HCT\end{tabular}}} & \multicolumn{1}{c}{\textbf{\begin{tabular}[c]{@{}c@{}}TVL\end{tabular}}} \\
            \midrule
            ViT-Tiny & \xmark & \xmark & 6.09 & 4.79 & 4.94 \\
            & \cmark & \xmark & 6.12 (\textcolor{blue}{+0.03}) & 4.87 (\textcolor{blue}{+0.08}) & 4.99 (\textcolor{blue}{+0.05}) \\
            \rowcolor{mycolor} & \cmark & \cmark & \textbf{6.21} (\textcolor{blue}{+0.12}) & \textbf{5.09} (\textcolor{blue}{+0.30}) & \textbf{5.22} (\textcolor{blue}{+0.28}) \\

            \midrule
            ViT-Small & \xmark & \xmark & 5.81 & 4.77 & 4.89 \\
            & \cmark & \xmark & 6.10 (\textcolor{blue}{+0.29}) & 4.92 (\textcolor{blue}{+0.15}) & 5.05 (\textcolor{blue}{+0.16}) \\
            \rowcolor{mycolor} & \cmark & \cmark & \textbf{6.13} (\textcolor{blue}{+0.32}) & \textbf{5.07} (\textcolor{blue}{+0.30}) & \textbf{5.19} (\textcolor{blue}{+0.30}) \\

            \midrule
            ViT-Base & \xmark & \xmark & 6.16 & 4.89 & 5.03 \\
            & \cmark & \xmark & 6.36 (\textcolor{blue}{+0.20}) & 4.95 (\textcolor{blue}{+0.06}) & 5.11 (\textcolor{blue}{+0.08}) \\
            \rowcolor{mycolor} & \cmark & \cmark & \textbf{6.73} (\textcolor{blue}{+0.57}) & \textbf{5.13} (\textcolor{blue}{+0.24}) & \textbf{5.32} (\textcolor{blue}{+0.29}) \\
            \bottomrule
              \multicolumn{6}{c}{\small
              ~\textbf{Retriever}: Tactile-Guided Retriever \quad  
              \textbf{Integrator}: Texture-Aware Integrator}
        \end{tabular}
    \end{adjustbox}
\caption{Ablation study of our proposed method. The integrator is not ablated independently, as it relies on the retriever for relevant input.}
\vspace{-10mm}
\label{tab:ablation}
\end{table}

\vspace{0.5mm}
\noindent \textbf{Ablation Study.}
We evaluated our proposed modules in Table~\ref{tab:ablation}. The baseline model is TVL-LLaMA~\cite{Fu2024ATV} with ViT backbones ranging from Tiny to Base. The Tactile-Guided Retriever enhances the model by injecting tactile-relevant external vision-language knowledge, allowing it to reason beyond its internal representation. Meanwhile, the Texture-Aware Integration selectively filters out object-centric and noise from retrieved vision-language features, enabling the model to focus on texture-relevant cues critical for tactile reasoning. 

We observe consistent performance improvements as modules are introduced. In particular, focusing on the ViT-Base variant, the Tactile-Guided Retriever alone improves the SSVTP~\cite{kerr2022ssvtp} score from 6.16 to 6.36, while the addition of Texture-Aware Integration further boosts the performance to 6.73. Similar trends are observed in HCT~\cite{Lambeta_2020}  and TVL~\cite{Fu2024ATV}, confirming the complementary benefits of both modules. Note that without the Texture-Aware Integration module, we apply a simple summation to integrate retrieval information and linear operation to match the feature dimensions.

\section{Further Analysis}

\begin{table}[t]
\centering
\footnotesize
\renewcommand{\arraystretch}{1.1}
\begin{adjustbox}{width=0.99\linewidth}
    \begin{tabular}{
      >{\raggedright\arraybackslash}m{1.4cm} |
      >{\centering\arraybackslash}m{2.1cm} |
      >{\raggedright\arraybackslash}p{1.3cm}
      >{\raggedright\arraybackslash}p{1.3cm}
      >{\raggedright\arraybackslash}p{1.3cm}
    }
    \toprule
    \textbf{Method} & \textbf{Caption Type} & \multicolumn{1}{c}{\textbf{\begin{tabular}[c]{@{}c@{}}SSVTP\end{tabular}}} & \multicolumn{1}{c}{\textbf{\begin{tabular}[c]{@{}c@{}}HCT\end{tabular}}} & \multicolumn{1}{c}{\textbf{\begin{tabular}[c]{@{}c@{}}TVL\end{tabular}}} \\
    \midrule
    TVL-LLaMA & - & 6.16  & 4.89 & 5.03 \\
    \Hquad +RA-Touch & Class Name & 6.48 (\textcolor{blue}{+0.32}) & 4.98 (\textcolor{blue}{+0.09}) & 5.15 (\textcolor{blue}{+0.12}) \\
    \Hquad +RA-Touch & Visual Description & 6.50 (\textcolor{blue}{+0.34}) & 5.07 (\textcolor{blue}{+0.18}) & 5.23 (\textcolor{blue}{+0.20}) \\
    \rowcolor{mycolor} 
    \Hquad \textbf{+RA-Touch} & \textbf{Tactile Description} & \textbf{6.73} (\textcolor{blue}{+0.57}) & \textbf{5.13} (\textcolor{blue}{+0.24}) & \textbf{5.32} (\textcolor{blue}{+0.29}) \\
    \bottomrule
    \end{tabular}
\end{adjustbox}
\caption{Performance comparison on different caption types.}
\label{tab:impact_tab_desc}
\vspace{-6mm}
\end{table}

\begin{table}[t]
\centering
\renewcommand{\arraystretch}{1.1}
\begin{adjustbox}{width=0.99\linewidth}
    \begin{tabular}{
      >{\centering\arraybackslash}m{2.5cm} |
      >{\centering\arraybackslash}m{2.5cm} |
      >{\centering\arraybackslash}p{1.3cm}
      >{\centering\arraybackslash}p{1.3cm}
      >{\centering\arraybackslash}p{1.3cm}
    }
    \toprule
    \textbf{Retrieval Query} & \textbf{Retrieval Key} & \textbf{SSVTP} & \textbf{HCT} & \textbf{TVL} \\
    \midrule
    Image & Image & 6.55 & 5.01 & 5.19 \\
    Image & Text & 6.52 & 5.05 & 5.22 \\
    Tactile & Image & 6.55 & 5.10 & 5.26 \\
    Tactile & Text & 6.54 & 5.07 & 5.24 \\
    \rowcolor{mycolor} \textbf{Query} & \textbf{Text} & \textbf{6.73} & \textbf{5.13} & \textbf{5.32} \\
    \bottomrule
    \end{tabular}
\end{adjustbox}
\caption{Performance comparison of retrieval method.}
\label{tab:retrieval}
\vspace{-8mm}
\end{table}

\subsection{Impact of Texture Descriptions}

To examine the impact of caption types on tactile understanding, we compare \ours using various forms of external knowledge, as shown in Table~\ref{tab:impact_tab_desc}. Performance improves progressively from class-level labels to visual descriptions and finally to texture-focused captions, suggesting that relevant semantic information leads to better tactile grounding. While both class names and visual descriptions lack explicit tactile semantics, they still lead to noticeable performance gains over the non-retrieval baseline. This suggests that our training framework effectively maintains tactile grounding in the CLIP~\cite{oord2018representation} embedding space through semantic alignment. This demonstrates the potential of RA-Touch to enhance tactile understanding using existing visual data without requiring additional tactile annotations. However, the most substantial gains come from the tactile descriptions, which is ImageNet-T dataset, tailored to better align vision-language knowledge with visuo-tactile inputs. This underscores the need of recaptioning in bridging the modality gap and supporting fine-grained tactile understanding.

\subsection{Effect of External Knowledge Source Scale}
\begin{figure}[t]
    \centering
    \vspace{-2mm}
    \includegraphics[width=0.85\linewidth]{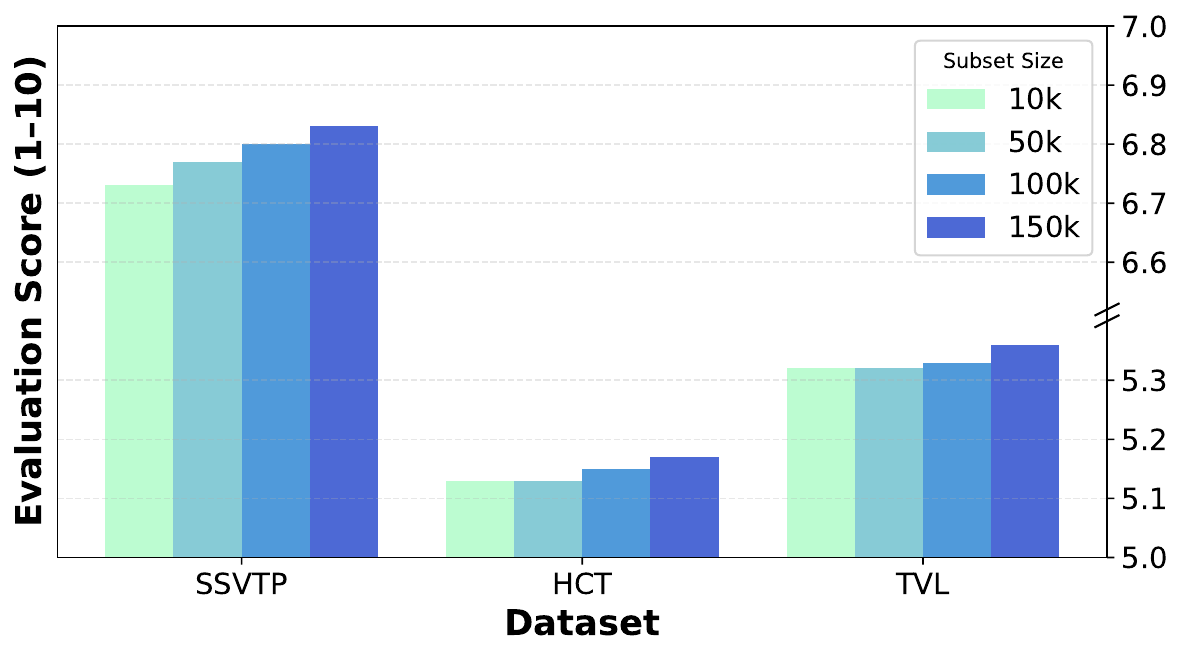}
    \vspace{-4mm}
    \caption{Performance comparisons across different subset sizes of ImageNet-T (10k, 50k, 100k, 150k) on three datasets: SSVTP, HCT, and TVL.}
    \label{fig:subset_size_comp}
    \vspace{-4mm}
\end{figure}

To evaluate the effect of retrieval scale, we compare model performance across different subset sizes of ImageNet-T (\ie 10k, 50k, 100k, 150k) on three benchmark datasets: SSVTP, HCT, and TVL. As shown in Figure~\ref{fig:subset_size_comp}, increasing the subset size consistently leads to performance gains across all datasets. This trend indicates that expanding the pool of vision-language knowledge with tactile-relevant descriptions can lead to better tactile understanding.

\subsection{Exploration of Tactile-Guided Retriever}

\noindent We evaluate the effectiveness of our Tactile-Guided Retrieval by comparing it with retrieval using image or tactile features itself as queries. As shown in Table~\ref{tab:retrieval}, both alternatives outperform the non-retrieval baseline but consistently fall short of our approach. 

\begin{figure}[t]
    \centering
    \begin{subfigure}[t]{0.99\linewidth}
        \centering
        \includegraphics[width=\linewidth]{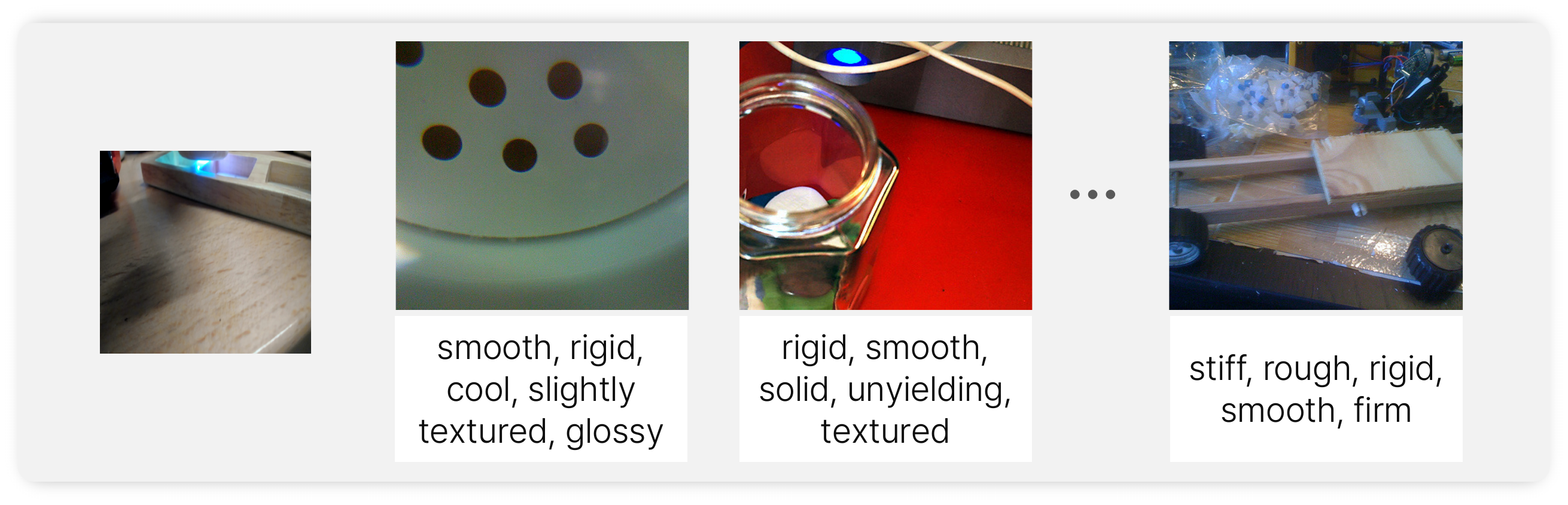}
        \vspace{-5mm}
        \subcaption{Image-to-Image Retrieval}
        \label{fig:retrieval_img2img}
    \end{subfigure}
    \vspace{2mm} %
    \begin{subfigure}[t]{0.99\linewidth}
        \centering
        \includegraphics[width=\linewidth]{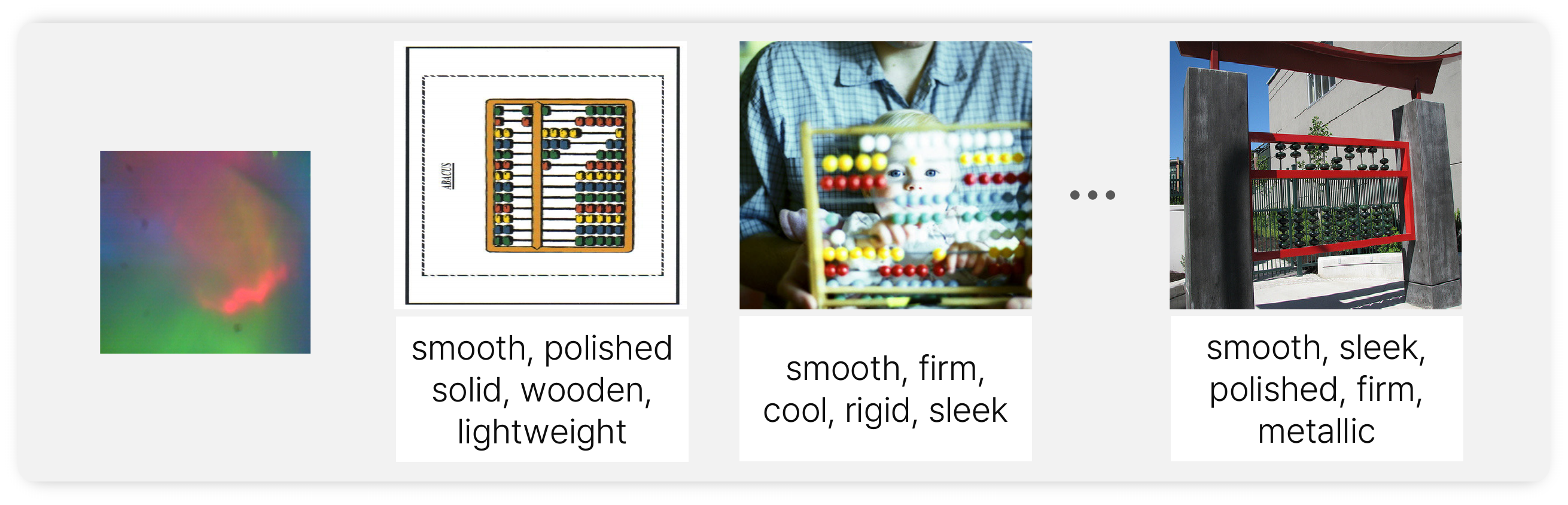}
        \vspace{-5mm}
        \subcaption{Tactile-to-Text Retrieval}
        \label{fig:retrieval_tac2txt}
    \end{subfigure}
    \vspace{-5mm}
    \caption{Retrieval results with visual or tactile features.  (a) Image-to-Image retrieves polished surface objects but lacks physical texture. (b) Tactile-to-Text focuses on text alone, retrieving a drawing of an abacus as Top-1.}
    \label{fig:main_retrieval_comp}
    \vspace{-4mm}
\end{figure}

\begin{figure}[t]
    \centering
    \begin{subfigure}[t]{\linewidth}
        \centering
        \includegraphics[width=0.99\linewidth]{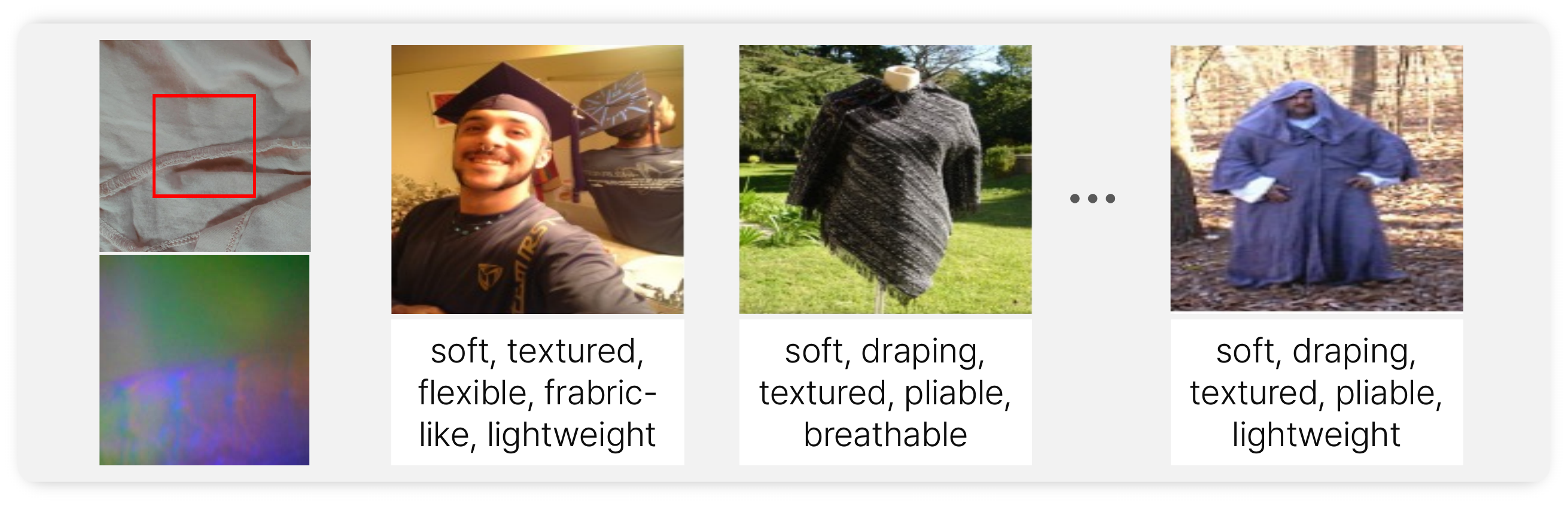}
        \vspace{-2mm}
        \subcaption{SSVTP}
        \label{fig:retrieval_ssvtp}
    \end{subfigure}
    \vspace{2mm} %
    \begin{subfigure}[t]{\linewidth}
        \centering
        \includegraphics[width=0.99\linewidth]{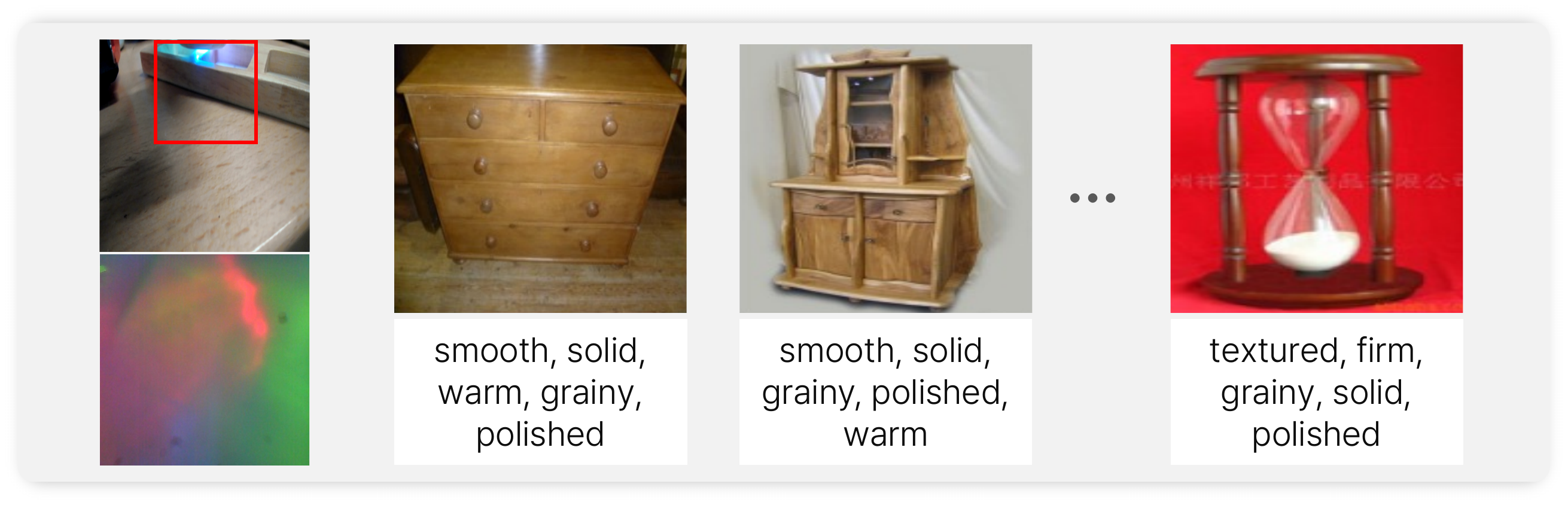}
        \vspace{-2mm}
        \subcaption{HCT}
        \label{fig:retrieval_hct}
    \end{subfigure}
    \vspace{-5mm}
    \caption{Example of retrieval samples from (a) SSVTP and (b) HCT with given inputs. The red bounding box indicates the region of contact sensed by the tactile sensor. Although five samples were retrieved, only three are shown for clarity.}
    \label{fig:retrieval_case}
    \vspace{-2mm}
\end{figure}

This performance gap arises from how each modality encodes tactile semantics. Image-based queries often retrieve visually similar samples—such as polished wooden surfaces—due to CLIP~\cite{Radford2021Learning} embeddings favoring appearance over material properties (Figure~\ref{fig:retrieval_img2img}). Tactile-to-text retrievals better reflect material-level cues like softness or roughness, but may return visually misleading results. For instance, the top-1 result in Figure~\ref{fig:retrieval_tac2txt} is a drawing of an abacus that shares shape but lacks relevant texture. In contrast, our method integrates both tactile and visual cues to retrieve examples that are not only relevant but also grounded in tactile meaning. As illustrated in Figure~\ref{fig:retrieval_case}, it successfully retrieves visually diverse yet materially similar objects. This highlights the model's ability to retrieve based on tactile semantics, not just visual resemblance.

\subsection{Analysis of Loss Strategies for Retriever}

\begin{figure}[t]
    \centering

    \begin{minipage}[b]{0.9\linewidth}
        \centering
        \includegraphics[width=0.75\textwidth]{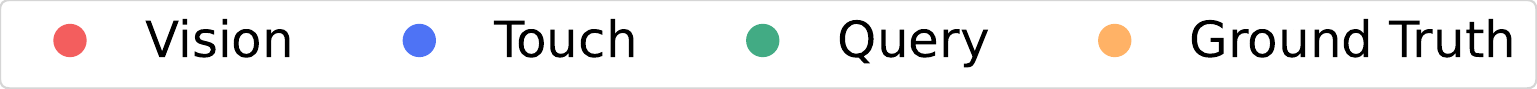}
    \end{minipage}
    \vspace{6mm}
    \begin{minipage}[b]{0.495\linewidth}
        \centering
        \includegraphics[width=0.9\textwidth]{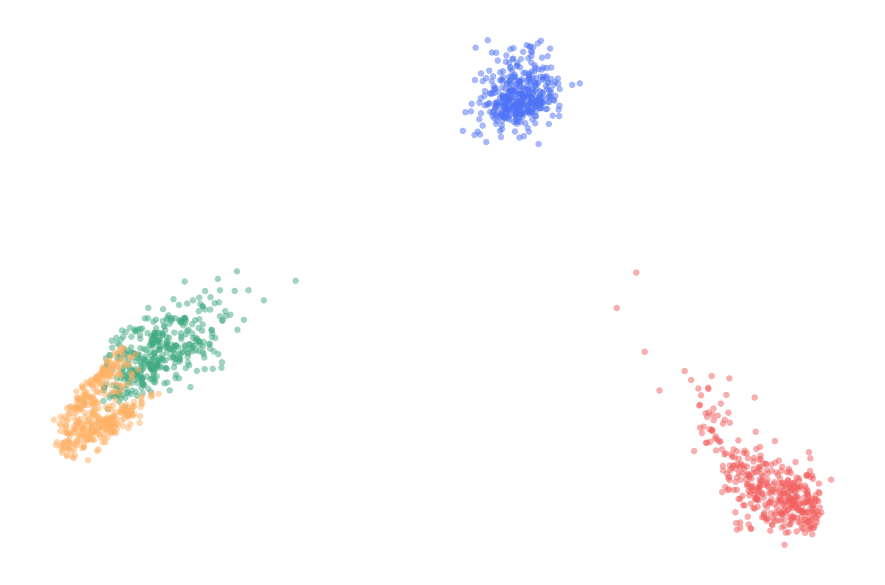}
        \subcaption{{\footnotesize $\mathcal{L}_{align}$}} \label{fig:embed_a}
    \end{minipage}
    \hfill
    \hfill
    \begin{minipage}[b]{0.495\linewidth}
        \centering
        \includegraphics[width=0.9\textwidth]{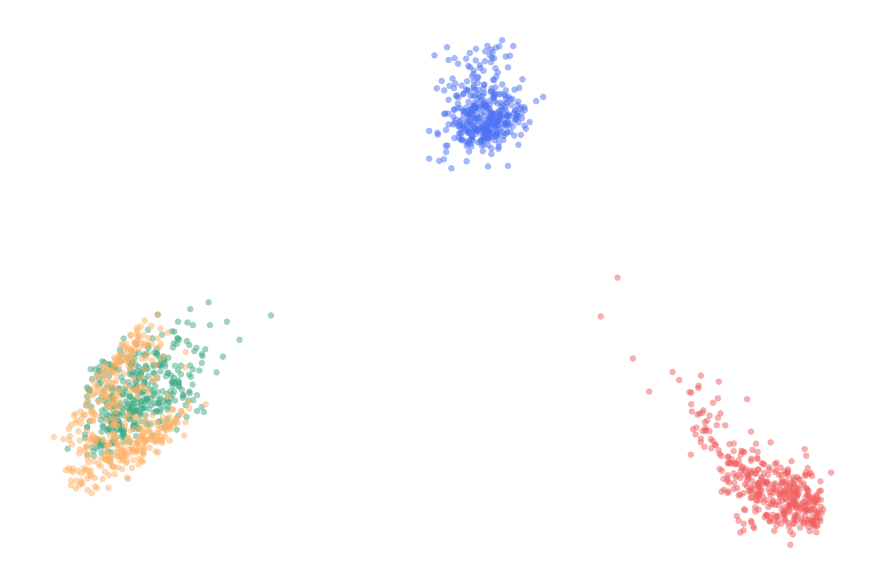}
        \subcaption{{\footnotesize $\mathcal{L}_{align} + \mathcal{L}_{stability}$}} \label{fig:embed_b}
    \end{minipage}
    \vspace{-8mm}
    \caption{Feature visualization of query embeddings. (a) shows results with alignment loss only, while (b) includes both alignment and stability losses.}
    \label{fig:embed_feats}
    \vspace{-5mm}
\end{figure}

To better understand how different loss strategies affect the semantic alignment of query embeddings, we visualize their distributions using PCA in Figure~\ref{fig:embed_feats}. Training with alignment loss $\mathcal{L}_{align}$ alone (Figure~\ref{fig:embed_a}) produces queries that are directionally aligned with ground truth but remain dispersed, reflecting limited semantic cohesion. In contrast, applying both alignment and stability losses jointly (Figure~\ref{fig:embed_b}) results in tighter, more coherent clusters that closely match the ground-truth. This indicates that the two losses play complementary roles: the alignment loss encourages proximity to the target semantics, while the stability loss promotes structural consistency and prevents representational collapse. We use PCA instead of t-SNE to ensure a globally consistent projection space across all settings for fair comparison.

\subsection{Effect of Top-$K$ Retrieval on Performance}
\begin{figure}[t]
    \centering
    \includegraphics[width=0.8\linewidth]{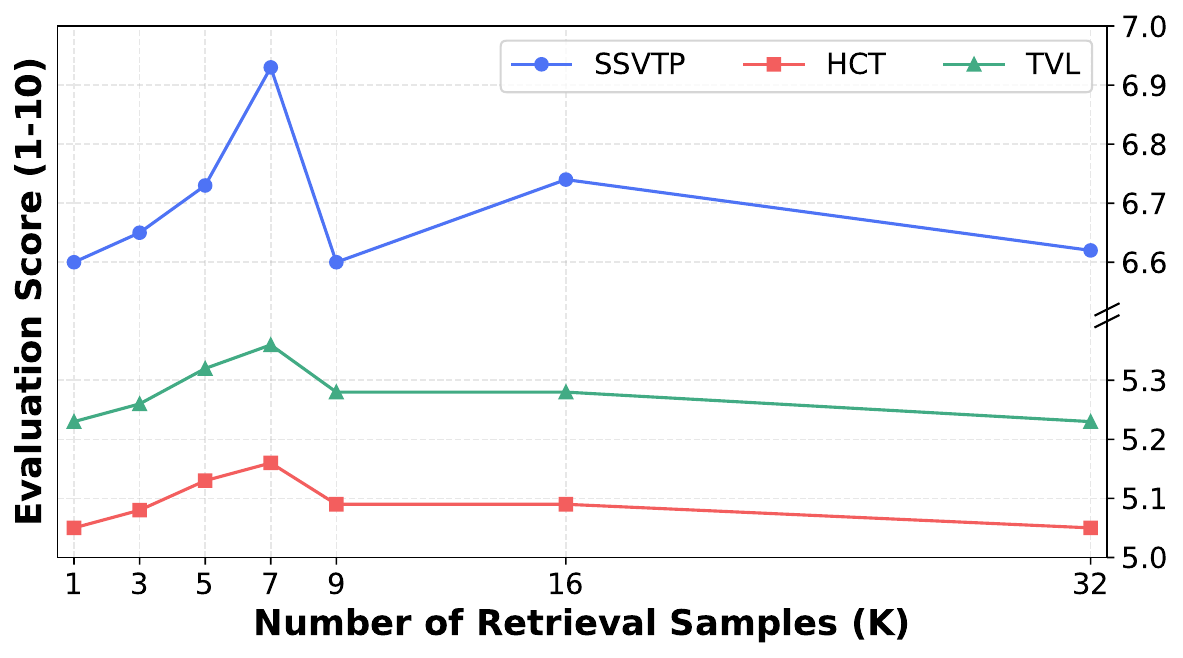}
    \caption{Effect of the numbers of retrieved samples ($K$) on three benchmarks.} 
    \label{fig:topk_comparison}
    \vspace{-2mm}
\end{figure}

We explore how the number of retrieved samples ($K$) influences the model’s capacity for tactile understanding. All experiments are conducted using \ours with ViT-Base and a default retrieval size of $K{=}5$. As shown in Figure~\ref{fig:topk_comparison}, performance generally improves as more relevant samples are retrieved, peaking at $K{=}7$, beyond which it degrades. This indicates that a moderately sized yet focused retrieval set offers the most useful context, while excessive retrieval may introduce redundancy or noise. We attribute this early performance drop to potential misalignment within the recaptioned vision-language dataset, which although curated for tactile understanding, may contain semantically distant or irrelevant samples. As $K$ increases, the retrieval pool broadens, which may lead to semantic inconsistencies, especially when the external retrieval source has limited diversity, increasing the chance of retrieving poorly aligned examples. This can be mitigated by using larger and more diverse retrieval datasets, as evidenced by the performance gains in Figure~\ref{fig:subset_size_comp}. These findings highlight the importance of balancing diversity and semantic relevance in retrieval size selection.

\subsection{Design Choices of Knowledge Integration}
We analyze how different retrieval modalities influence the performance of integration module. Specifically, we compare three configurations: image-only, text-only, and image-text combined retrieval features. The last setting corresponds to our method. 

As shown in Table~\ref{tab:integration_design}, both image-only and text-only retrievals contribute to performance gains over the no-retrieval baseline, indicating the usefulness of ImageNet-T. Text features alone often underperform compared to image features, likely due to their tendency to redundantly describe similar textures using overlapping language. In contrast, image features offer more diverse visual cues related to surface appearance and environmental context, enabling richer representations. Combining both modalities consistently yields the best performance across all benchmarks, highlighting their complementary nature and validating our design choice to integrate them for enhanced tactile understanding.

\begin{table}[t]
\centering
\small
\renewcommand{\arraystretch}{1.1}
\begin{adjustbox}{max width=0.9\linewidth}
    \begin{tabular}{cc|ccc}
    \toprule
    \textbf{Image} & \textbf{Text} & \textbf{SSVTP} & \textbf{HCT} & \textbf{TVL} \\
    \midrule
    \checkmark & & 6.52 & 5.05 & 5.22 \\
    & \checkmark & 6.48 & 5.00 & 5.17 \\
    \checkmark & \checkmark & \textbf{6.73} & \textbf{5.13} & \textbf{5.32} \\
    \bottomrule
    \end{tabular}
\end{adjustbox}
\caption{Analysis of Tactile-Aware Integration design choices.}
\label{tab:integration_design}
\vspace{-6mm}
\end{table}

\vspace{3mm}
\section{Conclusion}

We present \ours, a novel framework that rethinks tactile perception by leveraging vision-language data in a retrieval-augmented setting. Instead of relying on costly and labor-intensive tactile supervision, RA-Touch identifies semantically aligned samples from recaptioned visual corpora, enabling fine-grained texture reasoning from limited tactile input. By integrating a tactile-guided retrieval strategy with a texture-aware fusion module, our method consistently outperforms baseline models across multiple benchmarks. These results establish RA-Touch as a scalable and data-efficient solution for visuo-tactile learning, particularly in scenarios with limited touch data. We believe this approach opens up new directions for multimodal grounding and semantic alignment in low-resource sensory domains.

\clearpage
\bibliographystyle{ACM-Reference-Format}
\bibliography{main}


\begin{thebibliography}{78}


\ifx \showCODEN    \undefined \def \showCODEN     #1{\unskip}     \fi
\ifx \showISBNx    \undefined \def \showISBNx     #1{\unskip}     \fi
\ifx \showISBNxiii \undefined \def \showISBNxiii  #1{\unskip}     \fi
\ifx \showISSN     \undefined \def \showISSN      #1{\unskip}     \fi
\ifx \showLCCN     \undefined \def \showLCCN      #1{\unskip}     \fi
\ifx \shownote     \undefined \def \shownote      #1{#1}          \fi
\ifx \showarticletitle \undefined \def \showarticletitle #1{#1}   \fi
\ifx \showURL      \undefined \def \showURL       {\relax}        \fi
\providecommand\bibfield[2]{#2}
\providecommand\bibinfo[2]{#2}
\providecommand\natexlab[1]{#1}
\providecommand\showeprint[2][]{arXiv:#2}

\bibitem[Achiam et~al\mbox{.}(2023)]%
        {Achiam2023GPT}
\bibfield{author}{\bibinfo{person}{Josh Achiam}, \bibinfo{person}{Steven Adler}, \bibinfo{person}{Sandhini Agarwal}, \bibinfo{person}{Lama Ahmad}, \bibinfo{person}{Ilge Akkaya}, \bibinfo{person}{Florencia~Leoni Aleman}, \bibinfo{person}{Diogo Almeida}, \bibinfo{person}{Janko Altenschmidt}, \bibinfo{person}{Sam Altman}, \bibinfo{person}{Shyamal Anadkat}, {et~al\mbox{.}}} \bibinfo{year}{2023}\natexlab{}.
\newblock \showarticletitle{Gpt-4 technical report}.
\newblock \bibinfo{journal}{\emph{arXiv preprint arXiv:2303.08774}} (\bibinfo{year}{2023}).
\newblock


\bibitem[BehnamGhader et~al\mbox{.}(2024)]%
        {BehnamGhader2024LLM2VecLL}
\bibfield{author}{\bibinfo{person}{Parishad BehnamGhader}, \bibinfo{person}{Vaibhav Adlakha}, \bibinfo{person}{Marius Mosbach}, \bibinfo{person}{Dzmitry Bahdanau}, \bibinfo{person}{Nicolas Chapados}, {and} \bibinfo{person}{Siva Reddy}.} \bibinfo{year}{2024}\natexlab{}.
\newblock \showarticletitle{LLM2Vec: Large Language Models Are Secretly Powerful Text Encoders}. In \bibinfo{booktitle}{\emph{COLM}}.
\newblock


\bibitem[Blattmann and Robin~Rombach(2022)]%
        {Blattmann2022SemiParametricNI}
\bibfield{author}{\bibinfo{person}{Andreas Blattmann} {and} \bibinfo{person}{Jonas Müller Björn~Ommer Robin~Rombach, Kaan~Oktay}.} \bibinfo{year}{2022}\natexlab{}.
\newblock \showarticletitle{Semi-Parametric Neural Image Synthesis}. In \bibinfo{booktitle}{\emph{NeurIPS}}.
\newblock


\bibitem[Bresciani et~al\mbox{.}(2006)]%
        {Bresciani2006VisionAT}
\bibfield{author}{\bibinfo{person}{J. Bresciani}, \bibinfo{person}{Franziska Dammeier}, {and} \bibinfo{person}{M. Ernst}.} \bibinfo{year}{2006}\natexlab{}.
\newblock \showarticletitle{Vision and touch are automatically integrated for the perception of sequences of events}. In \bibinfo{booktitle}{\emph{Journal of Vision}}.
\newblock


\bibitem[Cai et~al\mbox{.}(2024)]%
        {cai2024vipllava}
\bibfield{author}{\bibinfo{person}{Mu Cai}, \bibinfo{person}{Haotian Liu}, \bibinfo{person}{Siva~Karthik Mustikovela}, \bibinfo{person}{Gregory~P. Meyer}, \bibinfo{person}{Yuning Chai}, \bibinfo{person}{Dennis Park}, {and} \bibinfo{person}{Yong~Jae Lee}.} \bibinfo{year}{2024}\natexlab{}.
\newblock \showarticletitle{Making Large Multimodal Models Understand Arbitrary Visual Prompts}. In \bibinfo{booktitle}{\emph{CVPR}}.
\newblock


\bibitem[Calandra et~al\mbox{.}(2018)]%
        {Calandra2018MoreTA}
\bibfield{author}{\bibinfo{person}{R. Calandra}, \bibinfo{person}{Andrew Owens}, \bibinfo{person}{Dinesh Jayaraman}, \bibinfo{person}{Justin Lin}, \bibinfo{person}{Wenzhen Yuan}, \bibinfo{person}{Jitendra Malik}, \bibinfo{person}{E. Adelson}, {and} \bibinfo{person}{S. Levine}.} \bibinfo{year}{2018}\natexlab{}.
\newblock \showarticletitle{More Than a Feeling: Learning to Grasp and Regrasp Using Vision and Touch}.
\newblock \bibinfo{journal}{\emph{IEEE Robotics and Automation Letters}} \bibinfo{volume}{3}, \bibinfo{number}{4} (\bibinfo{year}{2018}), \bibinfo{pages}{3300--3307}.
\newblock


\bibitem[Calandra et~al\mbox{.}(2017)]%
        {Calandra2017TheFO}
\bibfield{author}{\bibinfo{person}{R. Calandra}, \bibinfo{person}{Andrew Owens}, \bibinfo{person}{M. Upadhyaya}, \bibinfo{person}{Wenzhen Yuan}, \bibinfo{person}{Justin Lin}, \bibinfo{person}{E. Adelson}, {and} \bibinfo{person}{S. Levine}.} \bibinfo{year}{2017}\natexlab{}.
\newblock \showarticletitle{The Feeling of Success: Does Touch Sensing Help Predict Grasp Outcomes?}. In \bibinfo{booktitle}{\emph{CoRL}}.
\newblock


\bibitem[Camponogara and Volcic(2020)]%
        {Camponogara2020IntegrationOH}
\bibfield{author}{\bibinfo{person}{I. Camponogara} {and} \bibinfo{person}{R. Volcic}.} \bibinfo{year}{2020}\natexlab{}.
\newblock \showarticletitle{Integration of haptics and vision in human multisensory grasping}. In \bibinfo{booktitle}{\emph{Cortex}}.
\newblock


\bibitem[Chen et~al\mbox{.}(2022a)]%
        {Chen2022MuRAGMR}
\bibfield{author}{\bibinfo{person}{Wenhu Chen}, \bibinfo{person}{Hexiang Hu}, \bibinfo{person}{Xi Chen}, \bibinfo{person}{Pat Verga}, {and} \bibinfo{person}{William~W. Cohen}.} \bibinfo{year}{2022}\natexlab{a}.
\newblock \showarticletitle{MuRAG: Multimodal Retrieval-Augmented Generator for Open Question Answering over Images and Text}. In \bibinfo{booktitle}{\emph{EMNLP}}.
\newblock


\bibitem[Chen et~al\mbox{.}(2023)]%
        {Chen2023ReImagenRT}
\bibfield{author}{\bibinfo{person}{Wenhu Chen}, \bibinfo{person}{Hexiang Hu}, \bibinfo{person}{Chitwan Saharia}, {and} \bibinfo{person}{William~W. Cohen}.} \bibinfo{year}{2023}\natexlab{}.
\newblock \showarticletitle{Re-Imagen: Retrieval-Augmented Text-to-Image Generator}. In \bibinfo{booktitle}{\emph{ICLR}}.
\newblock


\bibitem[Chen et~al\mbox{.}(2022b)]%
        {chen2022visuo}
\bibfield{author}{\bibinfo{person}{Yizhou Chen}, \bibinfo{person}{Andrea Sipos}, \bibinfo{person}{Mark Van~der Merwe}, {and} \bibinfo{person}{Nima Fazeli}.} \bibinfo{year}{2022}\natexlab{b}.
\newblock \showarticletitle{Visuo-tactile transformers for manipulation}. In \bibinfo{booktitle}{\emph{CoRL}}.
\newblock


\bibitem[Cheng et~al\mbox{.}(2024a)]%
        {Cheng2024Touch100kAL}
\bibfield{author}{\bibinfo{person}{Ning Cheng}, \bibinfo{person}{Changhao Guan}, \bibinfo{person}{Jing Gao}, \bibinfo{person}{Weihao Wang}, \bibinfo{person}{You Li}, \bibinfo{person}{Fandong Meng}, \bibinfo{person}{Jie Zhou}, \bibinfo{person}{Bin Fang}, \bibinfo{person}{Jinan Xu}, {and} \bibinfo{person}{Wenjuan Han}.} \bibinfo{year}{2024}\natexlab{a}.
\newblock \showarticletitle{Touch100k: A Large-Scale Touch-Language-Vision Dataset for Touch-Centric Multimodal Representation}.
\newblock \bibinfo{journal}{\emph{arXiv preprint arXiv:2406.03813}} (\bibinfo{year}{2024}).
\newblock


\bibitem[Cheng et~al\mbox{.}(2024b)]%
        {cheng2024towards}
\bibfield{author}{\bibinfo{person}{Ning Cheng}, \bibinfo{person}{You Li}, \bibinfo{person}{Jing Gao}, \bibinfo{person}{Bin Fang}, \bibinfo{person}{Jinan Xu}, {and} \bibinfo{person}{Wenjuan Han}.} \bibinfo{year}{2024}\natexlab{b}.
\newblock \showarticletitle{Towards Comprehensive Multimodal Perception: Introducing the Touch-Language-Vision Dataset}, In \bibinfo{booktitle}{ICIC}.
\newblock \bibinfo{journal}{\emph{arXiv preprint arXiv:2403.09813}}.
\newblock


\bibitem[Chiang et~al\mbox{.}(2023)]%
        {vicuna2023}
\bibfield{author}{\bibinfo{person}{Wei-Lin Chiang}, \bibinfo{person}{Zhuohan Li}, \bibinfo{person}{Zi Lin}, \bibinfo{person}{Ying Sheng}, \bibinfo{person}{Zhanghao Wu}, \bibinfo{person}{Hao Zhang}, \bibinfo{person}{Lianmin Zheng}, \bibinfo{person}{Siyuan Zhuang}, \bibinfo{person}{Yonghao Zhuang}, \bibinfo{person}{Joseph~E. Gonzalez}, \bibinfo{person}{Ion Stoica}, {and} \bibinfo{person}{Eric~P. Xing}.} \bibinfo{year}{2023}\natexlab{}.
\newblock \bibinfo{title}{Vicuna: An Open-Source Chatbot Impressing GPT-4 with 90\%* ChatGPT Quality}.
\newblock


\bibitem[Dai et~al\mbox{.}(2023)]%
        {instructblip}
\bibfield{author}{\bibinfo{person}{Wenliang Dai}, \bibinfo{person}{Junnan Li}, \bibinfo{person}{Dongxu Li}, \bibinfo{person}{Anthony Meng~Huat Tiong}, \bibinfo{person}{Junqi Zhao}, \bibinfo{person}{Weisheng Wang}, \bibinfo{person}{Boyang Li}, \bibinfo{person}{Pascale Fung}, {and} \bibinfo{person}{Steven Hoi}.} \bibinfo{year}{2023}\natexlab{}.
\newblock \showarticletitle{InstructBLIP: Towards General-purpose Vision-Language Models with Instruction Tuning}. In \bibinfo{booktitle}{\emph{NeurIPS}}.
\newblock


\bibitem[Elizalde et~al\mbox{.}(2023)]%
        {Elizalde2023CLAPLA}
\bibfield{author}{\bibinfo{person}{Benjamin Elizalde}, \bibinfo{person}{Soham Deshmukh}, \bibinfo{person}{Mahmoud~Al Ismail}, {and} \bibinfo{person}{Huaming Wang}.} \bibinfo{year}{2023}\natexlab{}.
\newblock \showarticletitle{CLAP Learning Audio Concepts from Natural Language Supervision}. In \bibinfo{booktitle}{\emph{ICASSP}}. \bibinfo{pages}{1--5}.
\newblock


\bibitem[Fang et~al\mbox{.}(2024)]%
        {fang2024not}
\bibfield{author}{\bibinfo{person}{Xiang Fang}, \bibinfo{person}{Wanlong Fang}, \bibinfo{person}{Daizong Liu}, \bibinfo{person}{Xiaoye Qu}, \bibinfo{person}{Jianfeng Dong}, \bibinfo{person}{Pan Zhou}, \bibinfo{person}{Renfu Li}, \bibinfo{person}{Zichuan Xu}, \bibinfo{person}{Lixing Chen}, \bibinfo{person}{Panpan Zheng}, {et~al\mbox{.}}} \bibinfo{year}{2024}\natexlab{}.
\newblock \showarticletitle{Not all inputs are valid: Towards open-set video moment retrieval using language}. In \bibinfo{booktitle}{\emph{ACM MM}}. \bibinfo{pages}{28--37}.
\newblock


\bibitem[Feng et~al\mbox{.}(2024)]%
        {Feng2024PlayTT}
\bibfield{author}{\bibinfo{person}{Ruoxuan Feng}, \bibinfo{person}{Di Hu}, \bibinfo{person}{Wenke Ma}, {and} \bibinfo{person}{Xuelong Li}.} \bibinfo{year}{2024}\natexlab{}.
\newblock \showarticletitle{Play to the Score: Stage-Guided Dynamic Multi-Sensory Fusion for Robotic Manipulation}. In \bibinfo{booktitle}{\emph{CoRL}}.
\newblock


\bibitem[Fu et~al\mbox{.}(2024)]%
        {Fu2024ATV}
\bibfield{author}{\bibinfo{person}{Letian Fu}, \bibinfo{person}{Gaurav Datta}, \bibinfo{person}{Huang Huang}, \bibinfo{person}{Will Panitch}, \bibinfo{person}{Jaimyn Drake}, \bibinfo{person}{Joseph Ortiz}, \bibinfo{person}{Mustafa Mukadam}, \bibinfo{person}{Mike Lambeta}, \bibinfo{person}{Roberto Calandra}, {and} \bibinfo{person}{Ken Goldberg}.} \bibinfo{year}{2024}\natexlab{}.
\newblock \showarticletitle{A Touch, Vision, and Language Dataset for Multimodal Alignment}. In \bibinfo{booktitle}{\emph{ICML}}.
\newblock


\bibitem[Gao et~al\mbox{.}(2022)]%
        {Gao2022ObjectFolder2A}
\bibfield{author}{\bibinfo{person}{Ruohan Gao}, \bibinfo{person}{Zilin Si}, \bibinfo{person}{Yen-Yu Chang}, \bibinfo{person}{Samuel Clarke}, \bibinfo{person}{Jeannette Bohg}, \bibinfo{person}{Li Fei-Fei}, \bibinfo{person}{Wenzhen Yuan}, {and} \bibinfo{person}{Jiajun Wu}.} \bibinfo{year}{2022}\natexlab{}.
\newblock \showarticletitle{ObjectFolder 2.0: A Multisensory Object Dataset for Sim2Real Transfer}. In \bibinfo{booktitle}{\emph{CVPR}}.
\newblock


\bibitem[Guu et~al\mbox{.}(2020)]%
        {Guu2020REALMRA}
\bibfield{author}{\bibinfo{person}{Kelvin Guu}, \bibinfo{person}{Kenton Lee}, \bibinfo{person}{Zora Tung}, \bibinfo{person}{Panupong Pasupat}, {and} \bibinfo{person}{Ming-Wei Chang}.} \bibinfo{year}{2020}\natexlab{}.
\newblock \showarticletitle{REALM: Retrieval-Augmented Language Model Pre-Training}. In \bibinfo{booktitle}{\emph{ICML}}.
\newblock


\bibitem[Hurst et~al\mbox{.}(2024)]%
        {Hurst2024GPT}
\bibfield{author}{\bibinfo{person}{Aaron Hurst}, \bibinfo{person}{Adam Lerer}, \bibinfo{person}{Adam~P Goucher}, \bibinfo{person}{Adam Perelman}, \bibinfo{person}{Aditya Ramesh}, \bibinfo{person}{Aidan Clark}, \bibinfo{person}{AJ Ostrow}, \bibinfo{person}{Akila Welihinda}, \bibinfo{person}{Alan Hayes}, \bibinfo{person}{Alec Radford}, {et~al\mbox{.}}} \bibinfo{year}{2024}\natexlab{}.
\newblock \showarticletitle{Gpt-4o system card}.
\newblock \bibinfo{journal}{\emph{arXiv preprint arXiv:2410.21276}} (\bibinfo{year}{2024}).
\newblock


\bibitem[Ilharco et~al\mbox{.}(2021)]%
        {ilharco_2021_5143773}
\bibfield{author}{\bibinfo{person}{Gabriel Ilharco}, \bibinfo{person}{Mitchell Wortsman}, \bibinfo{person}{Nicholas Carlini}, \bibinfo{person}{Rohan Taori}, \bibinfo{person}{Achal Dave}, \bibinfo{person}{Vaishaal Shankar}, \bibinfo{person}{Hongseok Namkoong}, \bibinfo{person}{John Miller}, \bibinfo{person}{Hannaneh Hajishirzi}, \bibinfo{person}{Ali Farhadi}, {and} \bibinfo{person}{Ludwig Schmidt}.} \bibinfo{year}{2021}\natexlab{}.
\newblock \bibinfo{booktitle}{\emph{OpenCLIP}}.
\newblock


\bibitem[Ittyerah and Marks(2007)]%
        {Ittyerah2007MemoryFC}
\bibfield{author}{\bibinfo{person}{M. Ittyerah} {and} \bibinfo{person}{L. Marks}.} \bibinfo{year}{2007}\natexlab{}.
\newblock \showarticletitle{Memory for curvature of objects: haptic touch vs. vision}. In \bibinfo{booktitle}{\emph{British Journal of Psychology}}.
\newblock


\bibitem[Jones et~al\mbox{.}(2005)]%
        {Jones2005ACO}
\bibfield{author}{\bibinfo{person}{M.~G. Jones}, \bibinfo{person}{Alexandra Bokinsky}, \bibinfo{person}{T. Tretter}, {and} \bibinfo{person}{Atsuko Negishi}.} \bibinfo{year}{2005}\natexlab{}.
\newblock \showarticletitle{A comparison of learning with haptic and visual modalities}.
\newblock


\bibitem[Kerr et~al\mbox{.}(2022)]%
        {kerr2022ssvtp}
\bibfield{author}{\bibinfo{person}{Justin Kerr}, \bibinfo{person}{Huang Huang}, \bibinfo{person}{Albert Wilcox}, \bibinfo{person}{Ryan Hoque}, \bibinfo{person}{Jeffrey Ichnowski}, \bibinfo{person}{Roberto Calandra}, {and} \bibinfo{person}{Ken Goldberg}.} \bibinfo{year}{2022}\natexlab{}.
\newblock \showarticletitle{Self-supervised visuo-tactile pretraining to locate and follow garment features}.
\newblock \bibinfo{journal}{\emph{arXiv preprint arXiv:2209.13042}} (\bibinfo{year}{2022}).
\newblock


\bibitem[Lambeta et~al\mbox{.}(2020)]%
        {Lambeta_2020}
\bibfield{author}{\bibinfo{person}{Mike Lambeta}, \bibinfo{person}{Po-Wei Chou}, \bibinfo{person}{Stephen Tian}, \bibinfo{person}{Brian Yang}, \bibinfo{person}{Benjamin Maloon}, \bibinfo{person}{Victoria~Rose Most}, \bibinfo{person}{Dave Stroud}, \bibinfo{person}{Raymond Santos}, \bibinfo{person}{Ahmad Byagowi}, \bibinfo{person}{Gregg Kammerer}, {et~al\mbox{.}}} \bibinfo{year}{2020}\natexlab{}.
\newblock \showarticletitle{Digit: A novel design for a low-cost compact high-resolution tactile sensor with application to in-hand manipulation}.
\newblock \bibinfo{journal}{\emph{IEEE Robotics and Automation Letters}} \bibinfo{volume}{5}, \bibinfo{number}{3} (\bibinfo{year}{2020}), \bibinfo{pages}{3838--3845}.
\newblock


\bibitem[Layer(2024)]%
        {VisualLayer2024ImageNetEnriched}
\bibfield{author}{\bibinfo{person}{Visual Layer}.} \bibinfo{year}{2024}\natexlab{}.
\newblock \bibinfo{title}{imagenet-1k-vl-enriched}.
\newblock \bibinfo{howpublished}{\url{https://huggingface.co/datasets/visual-layer/imagenet-1k-vl-enriched}}.
\newblock


\bibitem[Lee et~al\mbox{.}(2025)]%
        {Lee2024NVEmbedIT}
\bibfield{author}{\bibinfo{person}{Chankyu Lee}, \bibinfo{person}{Rajarshi Roy}, \bibinfo{person}{Mengyao Xu}, \bibinfo{person}{Jonathan Raiman}, \bibinfo{person}{Mohammad Shoeybi}, \bibinfo{person}{Bryan Catanzaro}, {and} \bibinfo{person}{Wei Ping}.} \bibinfo{year}{2025}\natexlab{}.
\newblock \showarticletitle{NV-Embed: Improved Techniques for Training LLMs as Generalist Embedding Models}. In \bibinfo{booktitle}{\emph{ICLR}}.
\newblock


\bibitem[Lee et~al\mbox{.}(2019)]%
        {Lee2019ContextAwareER}
\bibfield{author}{\bibinfo{person}{Jiyoung Lee}, \bibinfo{person}{Seungryong Kim}, \bibinfo{person}{Sunok Kim}, \bibinfo{person}{Jungin Park}, {and} \bibinfo{person}{K. Sohn}.} \bibinfo{year}{2019}\natexlab{}.
\newblock \showarticletitle{Context-Aware Emotion Recognition Networks}. In \bibinfo{booktitle}{\emph{ICCV}}.
\newblock


\bibitem[Lee et~al\mbox{.}(2024)]%
        {lee2024fleur}
\bibfield{author}{\bibinfo{person}{Yebin Lee}, \bibinfo{person}{Imseong Park}, {and} \bibinfo{person}{Myungjoo Kang}.} \bibinfo{year}{2024}\natexlab{}.
\newblock \showarticletitle{FLEUR: An Explainable Reference-Free Evaluation Metric for Image Captioning Using a Large Multimodal Model}. In \bibinfo{booktitle}{\emph{ACL}}.
\newblock


\bibitem[Lewis et~al\mbox{.}(2020)]%
        {Lewis2020RetrievalAugmentedGF}
\bibfield{author}{\bibinfo{person}{Patrick Lewis}, \bibinfo{person}{Ethan Perez}, \bibinfo{person}{Aleksandra Piktus}, \bibinfo{person}{Fabio Petroni}, \bibinfo{person}{Vladimir Karpukhin}, \bibinfo{person}{Naman Goyal}, \bibinfo{person}{Heinrich Küttler}, \bibinfo{person}{Mike Lewis}, \bibinfo{person}{Wen tau Yih}, \bibinfo{person}{Tim Rocktäschel}, \bibinfo{person}{Sebastian Riedel}, {and} \bibinfo{person}{Douwe Kiela}.} \bibinfo{year}{2020}\natexlab{}.
\newblock \showarticletitle{Retrieval-Augmented Generation for Knowledge-Intensive NLP Tasks}. In \bibinfo{booktitle}{\emph{NeurIPS}}.
\newblock


\bibitem[Li et~al\mbox{.}(2020)]%
        {Li2020BoostingVQ}
\bibfield{author}{\bibinfo{person}{Guohao Li}, \bibinfo{person}{Xin Wang}, {and} \bibinfo{person}{Wenwu Zhu}.} \bibinfo{year}{2020}\natexlab{}.
\newblock \showarticletitle{Boosting Visual Question Answering with Context-aware Knowledge Aggregation}.
\newblock \bibinfo{journal}{\emph{ACM MM}} (\bibinfo{year}{2020}), \bibinfo{pages}{1227--1235}.
\newblock


\bibitem[Li et~al\mbox{.}(2023a)]%
        {Li2023ViHOPEVI}
\bibfield{author}{\bibinfo{person}{Hongyu Li}, \bibinfo{person}{Snehal Dikhale}, \bibinfo{person}{Soshi Iba}, {and} \bibinfo{person}{Nawid Jamali}.} \bibinfo{year}{2023}\natexlab{a}.
\newblock \showarticletitle{ViHOPE: Visuotactile In-Hand Object 6D Pose Estimation With Shape Completion}. In \bibinfo{booktitle}{\emph{IEEE Robotics and Automation Letters}}.
\newblock


\bibitem[Li et~al\mbox{.}(2022)]%
        {Li2022SeeHA}
\bibfield{author}{\bibinfo{person}{Hao Li}, \bibinfo{person}{Yizhi Zhang}, \bibinfo{person}{Junzhe Zhu}, \bibinfo{person}{Shaoxiong Wang}, \bibinfo{person}{Michelle~A. Lee}, \bibinfo{person}{Huazhe Xu}, \bibinfo{person}{E. Adelson}, \bibinfo{person}{Li Fei-Fei}, \bibinfo{person}{Ruohan Gao}, {and} \bibinfo{person}{Jiajun Wu}.} \bibinfo{year}{2022}\natexlab{}.
\newblock \showarticletitle{See, Hear, and Feel: Smart Sensory Fusion for Robotic Manipulation}. In \bibinfo{booktitle}{\emph{CoRL}}.
\newblock


\bibitem[Li et~al\mbox{.}(2023b)]%
        {Li2023BLIP2}
\bibfield{author}{\bibinfo{person}{Junnan Li}, \bibinfo{person}{Dongxu Li}, \bibinfo{person}{Silvio Savarese}, {and} \bibinfo{person}{Steven Hoi}.} \bibinfo{year}{2023}\natexlab{b}.
\newblock \showarticletitle{BLIP-2: Bootstrapping Language-Image Pre-training with Frozen Image Encoders and Large Language Models}. In \bibinfo{booktitle}{\emph{ICML}}.
\newblock


\bibitem[Li et~al\mbox{.}(2023c)]%
        {Li2023IntentQACV}
\bibfield{author}{\bibinfo{person}{Jiapeng Li}, \bibinfo{person}{Ping Wei}, \bibinfo{person}{Wenjuan Han}, {and} \bibinfo{person}{Lifeng Fan}.} \bibinfo{year}{2023}\natexlab{c}.
\newblock \showarticletitle{IntentQA: Context-aware Video Intent Reasoning}. In \bibinfo{booktitle}{\emph{ICCV}}.
\newblock


\bibitem[Li et~al\mbox{.}(2014)]%
        {Li2014LocalizationAM}
\bibfield{author}{\bibinfo{person}{Rui Li}, \bibinfo{person}{Robert~W. Platt}, \bibinfo{person}{Wenzhen Yuan}, \bibinfo{person}{A.~T. Pas}, \bibinfo{person}{Nathan Roscup}, \bibinfo{person}{M. Srinivasan}, {and} \bibinfo{person}{E. Adelson}.} \bibinfo{year}{2014}\natexlab{}.
\newblock \showarticletitle{Localization and manipulation of small parts using GelSight tactile sensing}. In \bibinfo{booktitle}{\emph{IROS}}.
\newblock


\bibitem[Li et~al\mbox{.}(2024)]%
        {li2024context}
\bibfield{author}{\bibinfo{person}{Shengdong Li}, \bibinfo{person}{Chen Gong}, \bibinfo{person}{Yuqing Zhu}, \bibinfo{person}{Chuanwen Luo}, \bibinfo{person}{Yi Hong}, {and} \bibinfo{person}{Xueqiang Lv}.} \bibinfo{year}{2024}\natexlab{}.
\newblock \showarticletitle{Context-aware multi-level question embedding fusion for visual question answering}.
\newblock \bibinfo{journal}{\emph{Information Fusion}}  \bibinfo{volume}{102} (\bibinfo{year}{2024}), \bibinfo{pages}{102000}.
\newblock


\bibitem[Li et~al\mbox{.}(2019)]%
        {Li_2019_CVPR}
\bibfield{author}{\bibinfo{person}{Yunzhu Li}, \bibinfo{person}{Jun-Yan Zhu}, \bibinfo{person}{Russ Tedrake}, {and} \bibinfo{person}{Antonio Torralba}.} \bibinfo{year}{2019}\natexlab{}.
\newblock \showarticletitle{Connecting Touch and Vision via Cross-Modal Prediction}. In \bibinfo{booktitle}{\emph{CVPR}}.
\newblock


\bibitem[Lin et~al\mbox{.}(2023b)]%
        {Lin2023FinegrainedLM}
\bibfield{author}{\bibinfo{person}{Weizhe Lin}, \bibinfo{person}{Jinghong Chen}, \bibinfo{person}{Jingbiao Mei}, \bibinfo{person}{Alexandru Coca}, {and} \bibinfo{person}{Bill Byrne}.} \bibinfo{year}{2023}\natexlab{b}.
\newblock \showarticletitle{Fine-grained Late-interaction Multi-modal Retrieval for Retrieval Augmented Visual Question Answering}. In \bibinfo{booktitle}{\emph{NeurIPS}}.
\newblock


\bibitem[Lin et~al\mbox{.}(2023a)]%
        {lin2023relaxing}
\bibfield{author}{\bibinfo{person}{Zudi Lin}, \bibinfo{person}{Erhan Bas}, \bibinfo{person}{Kunwar~Yashraj Singh}, \bibinfo{person}{Gurumurthy Swaminathan}, {and} \bibinfo{person}{Rahul Bhotika}.} \bibinfo{year}{2023}\natexlab{a}.
\newblock \showarticletitle{Relaxing contrastiveness in multimodal representation learning}. In \bibinfo{booktitle}{\emph{WACV}}. \bibinfo{pages}{2227--2236}.
\newblock


\bibitem[Liu et~al\mbox{.}(2024)]%
        {liu2024improved}
\bibfield{author}{\bibinfo{person}{Haotian Liu}, \bibinfo{person}{Chunyuan Li}, \bibinfo{person}{Yuheng Li}, {and} \bibinfo{person}{Yong~Jae Lee}.} \bibinfo{year}{2024}\natexlab{}.
\newblock \showarticletitle{Improved baselines with visual instruction tuning}. In \bibinfo{booktitle}{\emph{CVPR}}. \bibinfo{pages}{26296--26306}.
\newblock


\bibitem[Liu et~al\mbox{.}(2023a)]%
        {liu2023visual}
\bibfield{author}{\bibinfo{person}{Haotian Liu}, \bibinfo{person}{Chunyuan Li}, \bibinfo{person}{Qingyang Wu}, {and} \bibinfo{person}{Yong~Jae Lee}.} \bibinfo{year}{2023}\natexlab{a}.
\newblock \showarticletitle{Visual instruction tuning}. In \bibinfo{booktitle}{\emph{NeurIPS}}, Vol.~\bibinfo{volume}{36}. \bibinfo{pages}{34892--34916}.
\newblock


\bibitem[Liu et~al\mbox{.}(2023b)]%
        {Liu2022UniversalVD}
\bibfield{author}{\bibinfo{person}{Zhenghao Liu}, \bibinfo{person}{Chenyan Xiong}, \bibinfo{person}{Yuanhuiyi Lv}, \bibinfo{person}{Zhiyuan Liu}, {and} \bibinfo{person}{Ge Yu}.} \bibinfo{year}{2023}\natexlab{b}.
\newblock \showarticletitle{Universal Vision-Language Dense Retrieval: Learning A Unified Representation Space for Multi-Modal Retrieval}. In \bibinfo{booktitle}{\emph{ICLR}}.
\newblock


\bibitem[Lygerakis et~al\mbox{.}(2024)]%
        {calandra2017feeling}
\bibfield{author}{\bibinfo{person}{Fotios Lygerakis}, \bibinfo{person}{Vedant Dave}, {and} \bibinfo{person}{Elmar Rueckert}.} \bibinfo{year}{2024}\natexlab{}.
\newblock \showarticletitle{M2CURL: Sample-Efficient Multimodal Reinforcement Learning via Self-Supervised Representation Learning for Robotic Manipulation}. In \bibinfo{booktitle}{\emph{International Conference on Ubiquitous Robots}}. \bibinfo{pages}{490--497}.
\newblock


\bibitem[Mao et~al\mbox{.}(2024)]%
        {mao2024multimodal}
\bibfield{author}{\bibinfo{person}{Qian Mao}, \bibinfo{person}{Zijian Liao}, \bibinfo{person}{Jinfeng Yuan}, {and} \bibinfo{person}{Rong Zhu}.} \bibinfo{year}{2024}\natexlab{}.
\newblock \showarticletitle{Multimodal tactile sensing fused with vision for dexterous robotic housekeeping}.
\newblock \bibinfo{journal}{\emph{Nature Communications}} \bibinfo{volume}{15}, \bibinfo{number}{1} (\bibinfo{year}{2024}), \bibinfo{pages}{6871}.
\newblock


\bibitem[Min et~al\mbox{.}(2025)]%
        {Min2024SpeechRG}
\bibfield{author}{\bibinfo{person}{Do~June Min}, \bibinfo{person}{Karel Mundnich}, \bibinfo{person}{Andy Lapastora}, \bibinfo{person}{Erfan Soltanmohammadi}, \bibinfo{person}{S. Ronanki}, {and} \bibinfo{person}{Kyu~J Han}.} \bibinfo{year}{2025}\natexlab{}.
\newblock \showarticletitle{Speech Retrieval-Augmented Generation without Automatic Speech Recognition}. In \bibinfo{booktitle}{\emph{ICASSP}}.
\newblock


\bibitem[Ojala et~al\mbox{.}(2002)]%
        {Ojala2002MultiresolutionGA}
\bibfield{author}{\bibinfo{person}{T. Ojala}, \bibinfo{person}{M. Pietikäinen}, {and} \bibinfo{person}{Topi Mäenpää}.} \bibinfo{year}{2002}\natexlab{}.
\newblock \showarticletitle{Multiresolution Gray-Scale and Rotation Invariant Texture Classification with Local Binary Patterns}. In \bibinfo{booktitle}{\emph{IEEE TPAMI}}.
\newblock


\bibitem[Oord et~al\mbox{.}(2018)]%
        {oord2018representation}
\bibfield{author}{\bibinfo{person}{Aaron van~den Oord}, \bibinfo{person}{Yazhe Li}, {and} \bibinfo{person}{Oriol Vinyals}.} \bibinfo{year}{2018}\natexlab{}.
\newblock \showarticletitle{Representation learning with contrastive predictive coding}.
\newblock \bibinfo{journal}{\emph{arXiv preprint arXiv:1807.03748}} (\bibinfo{year}{2018}).
\newblock


\bibitem[Pecyna et~al\mbox{.}(2022)]%
        {pecyna2022visual}
\bibfield{author}{\bibinfo{person}{Leszek Pecyna}, \bibinfo{person}{Siyuan Dong}, {and} \bibinfo{person}{Shan Luo}.} \bibinfo{year}{2022}\natexlab{}.
\newblock \showarticletitle{Visual-tactile multimodality for following deformable linear objects using reinforcement learning}. In \bibinfo{booktitle}{\emph{IROS}}. IEEE, \bibinfo{pages}{3987--3994}.
\newblock


\bibitem[Radford et~al\mbox{.}(2021)]%
        {Radford2021Learning}
\bibfield{author}{\bibinfo{person}{Alec Radford}, \bibinfo{person}{Jong~Wook Kim}, \bibinfo{person}{Chris Hallacy}, \bibinfo{person}{Aditya Ramesh}, \bibinfo{person}{Gabriel Goh}, \bibinfo{person}{Sandhini Agarwal}, \bibinfo{person}{Girish Sastry}, \bibinfo{person}{Amanda Askell}, \bibinfo{person}{Pamela Mishkin}, \bibinfo{person}{Jack Clark}, {et~al\mbox{.}}} \bibinfo{year}{2021}\natexlab{}.
\newblock \showarticletitle{Learning transferable visual models from natural language supervision}. In \bibinfo{booktitle}{\emph{ICML}}.
\newblock


\bibitem[Ramos et~al\mbox{.}(2023)]%
        {Ramos2023RetrievalaugmentedIC}
\bibfield{author}{\bibinfo{person}{Rita Ramos}, \bibinfo{person}{Desmond Elliott}, {and} \bibinfo{person}{Bruno Martins}.} \bibinfo{year}{2023}\natexlab{}.
\newblock \showarticletitle{Retrieval-augmented Image Captioning}. In \bibinfo{booktitle}{\emph{ACL}}.
\newblock


\bibitem[Rao et~al\mbox{.}(2023)]%
        {rao2023retrieval}
\bibfield{author}{\bibinfo{person}{Jiahua Rao}, \bibinfo{person}{Zifei Shan}, \bibinfo{person}{Longpo Liu}, \bibinfo{person}{Yao Zhou}, {and} \bibinfo{person}{Yuedong Yang}.} \bibinfo{year}{2023}\natexlab{}.
\newblock \showarticletitle{Retrieval-based knowledge augmented vision language pre-training}. In \bibinfo{booktitle}{\emph{ACM MM}}. \bibinfo{pages}{5399--5409}.
\newblock


\bibitem[Russakovsky et~al\mbox{.}(2015)]%
        {russakovsky2015imagenet}
\bibfield{author}{\bibinfo{person}{Olga Russakovsky}, \bibinfo{person}{Jia Deng}, \bibinfo{person}{Hao Su}, \bibinfo{person}{Jonathan Krause}, \bibinfo{person}{Sanjeev Satheesh}, \bibinfo{person}{Sean Ma}, \bibinfo{person}{Zhiheng Huang}, \bibinfo{person}{Andrej Karpathy}, \bibinfo{person}{Aditya Khosla}, \bibinfo{person}{Michael Bernstein}, \bibinfo{person}{Alexander~C. Berg}, {and} \bibinfo{person}{Li Fei-Fei}.} \bibinfo{year}{2015}\natexlab{}.
\newblock \showarticletitle{ImageNet Large Scale Visual Recognition Challenge}.
\newblock \bibinfo{journal}{\emph{International Journal of Computer Vision}}  \bibinfo{volume}{115} (\bibinfo{year}{2015}), \bibinfo{pages}{211--252}.
\newblock


\bibitem[Sferrazza and D’Andrea(2019)]%
        {Sferrazza2019DesignMA}
\bibfield{author}{\bibinfo{person}{Carmelo Sferrazza} {and} \bibinfo{person}{R. D’Andrea}.} \bibinfo{year}{2019}\natexlab{}.
\newblock \showarticletitle{Design, Motivation and Evaluation of a Full-Resolution Optical Tactile Sensor}. In \bibinfo{booktitle}{\emph{Sensors}}.
\newblock


\bibitem[Shimonomura(2019)]%
        {Shimonomura2019TactileIS}
\bibfield{author}{\bibinfo{person}{K. Shimonomura}.} \bibinfo{year}{2019}\natexlab{}.
\newblock \showarticletitle{Tactile Image Sensors Employing Camera: A Review}. In \bibinfo{booktitle}{\emph{Sensors}}.
\newblock


\bibitem[Smith et~al\mbox{.}(2020)]%
        {smith20203d}
\bibfield{author}{\bibinfo{person}{Edward Smith}, \bibinfo{person}{Roberto Calandra}, \bibinfo{person}{Adriana Romero}, \bibinfo{person}{Georgia Gkioxari}, \bibinfo{person}{David Meger}, \bibinfo{person}{Jitendra Malik}, {and} \bibinfo{person}{Michal Drozdzal}.} \bibinfo{year}{2020}\natexlab{}.
\newblock \showarticletitle{3D Shape Reconstruction from Vision and Touch}.
\newblock \bibinfo{journal}{\emph{NeurIPS}}  \bibinfo{volume}{33} (\bibinfo{year}{2020}), \bibinfo{pages}{14193--14206}.
\newblock


\bibitem[Stone and Gonzalez(2015)]%
        {Stone2015TheCO}
\bibfield{author}{\bibinfo{person}{K. Stone} {and} \bibinfo{person}{Claudia L.~R. Gonzalez}.} \bibinfo{year}{2015}\natexlab{}.
\newblock \showarticletitle{The contributions of vision and haptics to reaching and grasping}. In \bibinfo{booktitle}{\emph{Frontiers in Psychology}}.
\newblock


\bibitem[Suresh et~al\mbox{.}(2022)]%
        {suresh2022shapemap}
\bibfield{author}{\bibinfo{person}{Sudharshan Suresh}, \bibinfo{person}{Zilin Si}, \bibinfo{person}{Joshua~G Mangelson}, \bibinfo{person}{Wenzhen Yuan}, {and} \bibinfo{person}{Michael Kaess}.} \bibinfo{year}{2022}\natexlab{}.
\newblock \showarticletitle{ShapeMap 3-D: Efficient shape mapping through dense touch and vision}. In \bibinfo{booktitle}{\emph{ICRA}}. IEEE, \bibinfo{pages}{7073--7080}.
\newblock


\bibitem[Thawakar et~al\mbox{.}(2024)]%
        {Thawakar2024ComposedVR}
\bibfield{author}{\bibinfo{person}{Omkar Thawakar}, \bibinfo{person}{Muzammal Naseer}, \bibinfo{person}{Rao~Muhammad Anwer}, \bibinfo{person}{Salman Khan}, \bibinfo{person}{Michael Felsberg}, \bibinfo{person}{Mubarak Shah}, {and} \bibinfo{person}{Fahad~Shahbaz Khan}.} \bibinfo{year}{2024}\natexlab{}.
\newblock \showarticletitle{Composed Video Retrieval via Enriched Context and Discriminative Embeddings}. In \bibinfo{booktitle}{\emph{CVPR}}.
\newblock


\bibitem[Touvron et~al\mbox{.}(2023)]%
        {touvron2023llama2openfoundation}
\bibfield{author}{\bibinfo{person}{Hugo Touvron}, \bibinfo{person}{Louis Martin}, \bibinfo{person}{Kevin Stone}, \bibinfo{person}{Peter Albert}, \bibinfo{person}{Amjad Almahairi}, \bibinfo{person}{Yasmine Babaei}, \bibinfo{person}{Nikolay Bashlykov}, \bibinfo{person}{Soumya Batra}, \bibinfo{person}{Prajjwal Bhargava}, \bibinfo{person}{Shruti Bhosale}, {et~al\mbox{.}}} \bibinfo{year}{2023}\natexlab{}.
\newblock \showarticletitle{Llama 2: Open foundation and fine-tuned chat models}.
\newblock \bibinfo{journal}{\emph{arXiv preprint arXiv:2307.09288}} (\bibinfo{year}{2023}).
\newblock


\bibitem[Wang et~al\mbox{.}(2022)]%
        {wang2022multimodal}
\bibfield{author}{\bibinfo{person}{Xue Wang}, \bibinfo{person}{Zhanshan Li}, \bibinfo{person}{Yongping Huang}, {and} \bibinfo{person}{Yingying Jiao}.} \bibinfo{year}{2022}\natexlab{}.
\newblock \showarticletitle{Multimodal medical image segmentation using multi-scale context-aware network}.
\newblock \bibinfo{journal}{\emph{Neurocomputing}}  \bibinfo{volume}{486} (\bibinfo{year}{2022}), \bibinfo{pages}{135--146}.
\newblock


\bibitem[Wei et~al\mbox{.}(2024)]%
        {Wei2023UniIRTA}
\bibfield{author}{\bibinfo{person}{Cong Wei}, \bibinfo{person}{Yang Chen}, \bibinfo{person}{Haonan Chen}, \bibinfo{person}{Hexiang Hu}, \bibinfo{person}{Ge Zhang}, \bibinfo{person}{Jie Fu}, \bibinfo{person}{Alan Ritter}, {and} \bibinfo{person}{Wenhu Chen}.} \bibinfo{year}{2024}\natexlab{}.
\newblock \showarticletitle{Uniir: Training and benchmarking universal multimodal information retrievers}. In \bibinfo{booktitle}{\emph{ECCV}}. Springer, \bibinfo{pages}{387--404}.
\newblock


\bibitem[Wu et~al\mbox{.}(2024)]%
        {Wu2024MultiMF}
\bibfield{author}{\bibinfo{person}{Guanfeng Wu}, \bibinfo{person}{Abbas Haider}, \bibinfo{person}{Ivor Spence}, {and} \bibinfo{person}{Hui Wang}.} \bibinfo{year}{2024}\natexlab{}.
\newblock \showarticletitle{Multi Modal Fusion for Video Retrieval based on CLIP Guide Feature Alignment}. In \bibinfo{booktitle}{\emph{MVRMLM '24: Proceedings of 2024 ACM ICMR Workshop on Multimodal Video Retrieval}}.
\newblock


\bibitem[Xie et~al\mbox{.}(2023)]%
        {Xie2023RACLIPRA}
\bibfield{author}{\bibinfo{person}{Chen-Wei Xie}, \bibinfo{person}{Siyang Sun}, \bibinfo{person}{Xiong Xiong}, \bibinfo{person}{Yun Zheng}, \bibinfo{person}{Deli Zhao}, {and} \bibinfo{person}{Jingren Zhou}.} \bibinfo{year}{2023}\natexlab{}.
\newblock \showarticletitle{RA-CLIP: Retrieval Augmented Contrastive Language-Image Pre-Training}. In \bibinfo{booktitle}{\emph{CVPR}}.
\newblock


\bibitem[Xu et~al\mbox{.}(2024)]%
        {Xu2024UniTUT}
\bibfield{author}{\bibinfo{person}{Zhengtong Xu}, \bibinfo{person}{Raghava Uppuluri}, \bibinfo{person}{Xinwei Zhang}, \bibinfo{person}{Cael Fitch}, \bibinfo{person}{Philip~Glen Crandall}, \bibinfo{person}{Wan Shou}, \bibinfo{person}{Dongyi Wang}, {and} \bibinfo{person}{Yu She}.} \bibinfo{year}{2024}\natexlab{}.
\newblock \showarticletitle{UniT: Unified Tactile Representation for Robot Learning}. In \bibinfo{booktitle}{\emph{arXiv.org}}.
\newblock


\bibitem[Yamaguchi and Atkeson(2016)]%
        {Yamaguchi2016CombiningFV}
\bibfield{author}{\bibinfo{person}{Akihiko Yamaguchi} {and} \bibinfo{person}{C. Atkeson}.} \bibinfo{year}{2016}\natexlab{}.
\newblock \showarticletitle{Combining finger vision and optical tactile sensing: Reducing and handling errors while cutting vegetables}. In \bibinfo{booktitle}{\emph{IEEE-RAS International Conference on Humanoid Robots}}.
\newblock


\bibitem[Yang et~al\mbox{.}(2024)]%
        {Yang2024BindingTT}
\bibfield{author}{\bibinfo{person}{Fengyu Yang}, \bibinfo{person}{Chao Feng}, \bibinfo{person}{Ziyang Chen}, \bibinfo{person}{Hyoungseob Park}, \bibinfo{person}{Daniel Wang}, \bibinfo{person}{Yiming Dou}, \bibinfo{person}{Ziyao Zeng}, \bibinfo{person}{Xien Chen}, \bibinfo{person}{Rit Gangopadhyay}, \bibinfo{person}{Andrew Owens}, {and} \bibinfo{person}{Alex Wong}.} \bibinfo{year}{2024}\natexlab{}.
\newblock \showarticletitle{Binding Touch to Everything: Learning Unified Multimodal Tactile Representations}. In \bibinfo{booktitle}{\emph{CVPR}}.
\newblock


\bibitem[Yang et~al\mbox{.}(2022)]%
        {Yang2022TouchAG}
\bibfield{author}{\bibinfo{person}{Fengyu Yang}, \bibinfo{person}{Chenyang Ma}, \bibinfo{person}{Jiacheng Zhang}, \bibinfo{person}{Jing Zhu}, \bibinfo{person}{Wenzhen Yuan}, {and} \bibinfo{person}{Andrew Owens}.} \bibinfo{year}{2022}\natexlab{}.
\newblock \showarticletitle{Touch and Go: Learning from Human-Collected Vision and Touch}. In \bibinfo{booktitle}{\emph{NeurIPS}}.
\newblock


\bibitem[Yu et~al\mbox{.}(2024)]%
        {yu2024octopi}
\bibfield{author}{\bibinfo{person}{Samson Yu}, \bibinfo{person}{Kelvin Lin}, \bibinfo{person}{Anxing Xiao}, \bibinfo{person}{Jiafei Duan}, {and} \bibinfo{person}{Harold Soh}.} \bibinfo{year}{2024}\natexlab{}.
\newblock \showarticletitle{Octopi: Object property reasoning with large tactile-language models}.
\newblock \bibinfo{journal}{\emph{arXiv preprint arXiv:2405.02794}} (\bibinfo{year}{2024}).
\newblock


\bibitem[Yuan et~al\mbox{.}(2017)]%
        {Yuan2017GelSightHR}
\bibfield{author}{\bibinfo{person}{Wenzhen Yuan}, \bibinfo{person}{Siyuan Dong}, {and} \bibinfo{person}{E. Adelson}.} \bibinfo{year}{2017}\natexlab{}.
\newblock \showarticletitle{GelSight: High-Resolution Robot Tactile Sensors for Estimating Geometry and Force}. In \bibinfo{booktitle}{\emph{Sensors}}.
\newblock


\bibitem[Zbontar et~al\mbox{.}(2021)]%
        {zbontar2021barlow}
\bibfield{author}{\bibinfo{person}{Jure Zbontar}, \bibinfo{person}{Li Jing}, \bibinfo{person}{Ishan Misra}, \bibinfo{person}{Yann LeCun}, {and} \bibinfo{person}{St{\'e}phane Deny}.} \bibinfo{year}{2021}\natexlab{}.
\newblock \showarticletitle{Barlow twins: Self-supervised learning via redundancy reduction}. In \bibinfo{booktitle}{\emph{ICML}}. PMLR, \bibinfo{pages}{12310--12320}.
\newblock


\bibitem[Zha et~al\mbox{.}(2019)]%
        {zha2019context}
\bibfield{author}{\bibinfo{person}{Zheng-Jun Zha}, \bibinfo{person}{Daqing Liu}, \bibinfo{person}{Hanwang Zhang}, \bibinfo{person}{Yongdong Zhang}, {and} \bibinfo{person}{Feng Wu}.} \bibinfo{year}{2019}\natexlab{}.
\newblock \showarticletitle{Context-aware visual policy network for fine-grained image captioning}.
\newblock \bibinfo{journal}{\emph{IEEE TPAMI}} \bibinfo{volume}{44}, \bibinfo{number}{2} (\bibinfo{year}{2019}), \bibinfo{pages}{710--722}.
\newblock


\bibitem[Zhang and Demiris(2023)]%
        {zhang2023visual}
\bibfield{author}{\bibinfo{person}{Fan Zhang} {and} \bibinfo{person}{Yiannis Demiris}.} \bibinfo{year}{2023}\natexlab{}.
\newblock \showarticletitle{Visual-tactile learning of garment unfolding for robot-assisted dressing}.
\newblock \bibinfo{journal}{\emph{IEEE Robotics and Automation Letters}} \bibinfo{volume}{8}, \bibinfo{number}{9} (\bibinfo{year}{2023}), \bibinfo{pages}{5512--5519}.
\newblock


\bibitem[Zhang et~al\mbox{.}(2023)]%
        {Zhang2023ReMoDiffuseRA}
\bibfield{author}{\bibinfo{person}{Mingyuan Zhang}, \bibinfo{person}{Xinying Guo}, \bibinfo{person}{Liang Pan}, \bibinfo{person}{Zhongang Cai}, \bibinfo{person}{Fangzhou Hong}, \bibinfo{person}{Huirong Li}, \bibinfo{person}{Lei Yang}, {and} \bibinfo{person}{Ziwei Liu}.} \bibinfo{year}{2023}\natexlab{}.
\newblock \showarticletitle{ReMoDiffuse: Retrieval-Augmented Motion Diffusion Model}. In \bibinfo{booktitle}{\emph{ICCV}}.
\newblock


\bibitem[Zhang et~al\mbox{.}(2020)]%
        {Zhang2020ContextAwareAN}
\bibfield{author}{\bibinfo{person}{Qi Zhang}, \bibinfo{person}{Zhen Lei}, \bibinfo{person}{Zhaoxiang Zhang}, {and} \bibinfo{person}{S. Li}.} \bibinfo{year}{2020}\natexlab{}.
\newblock \showarticletitle{Context-Aware Attention Network for Image-Text Retrieval}. In \bibinfo{booktitle}{\emph{CVPR}}.
\newblock


\bibitem[Zhang et~al\mbox{.}(2024)]%
        {zhang2024llamaadapter}
\bibfield{author}{\bibinfo{person}{Renrui Zhang}, \bibinfo{person}{Jiaming Han}, \bibinfo{person}{Chris Liu}, \bibinfo{person}{Aojun Zhou}, \bibinfo{person}{Pan Lu}, \bibinfo{person}{Yu Qiao}, \bibinfo{person}{Hongsheng Li}, {and} \bibinfo{person}{Peng Gao}.} \bibinfo{year}{2024}\natexlab{}.
\newblock \showarticletitle{{LL}a{MA}-Adapter: Efficient Fine-tuning of Large Language Models with Zero-initialized Attention}. In \bibinfo{booktitle}{\emph{ICLR}}.
\newblock


\end{thebibliography}

\clearpage

\appendix

\renewcommand\thefigure{\Alph{figure}}    
\setcounter{figure}{0}  
\renewcommand\thetable{\Alph{table}}
\setcounter{table}{0} 

\title[Supplementary for RA-Touch: Retrieval-Augmented Touch Understanding with Enriched Visual Data]{Supplementary for RA-Touch: Retrieval-Augmented Touch Understanding with Enriched Visual Data}

\section{Exploration of Captioning Models}
We examined how the choice of captioning model for constructing an external knowledge source affects the performance of the model and the quality of the generated captions. Specifically, we trained our model with captions generated by various captioning models, including BLIP-2 Opt-6.7B~\cite{Li2023BLIP2}, InstructBLIP 13B~\cite{instructblip}, LLaVA-1.5 13B~\cite{liu2024improved}, and GPT-4o mini~\cite{Hurst2024GPT}. We then evaluated each variant on the TVL benchmark~\cite{Fu2024ATV} to determine which captioning model produced the most effective knowledge source.

In addition to TVL benchmark scores, we conducted a tactile relevance evaluation to directly assess the intrinsic quality and tactile alignment. For this evaluation, we employed the GPT-4o~\cite{Hurst2024GPT} model to rate on a 1-10 scale how effectively each caption described the tactile attributes of the depicted object, given both image and context. The evaluation prompt is a modified version of the one proposed in FLEUR~\cite{lee2024fleur}, with the complete prompt provided in Table~\ref{tab:gpt_eval_templete}. As shown in Table~\ref{tab:caption_comparison} and Table~\ref{tab:caption_quality}, GPT-4o mini achieves the best performance across all recaption models, indicating its superior ability to leverage tactile-grounded information. These results suggest that not all VLMs are equally capable of interpreting non-visual cues embedded in the descriptions, and careful selection of the captioning model is crucial for downstream tactile perception tasks. Note that we conduct the TVL benchmark evaluation on a subset size of 10k, while the tactile relevance evaluation on a subset size of 1k.

\begin{figure}
    \centering
    \includegraphics[width=1.0\linewidth]{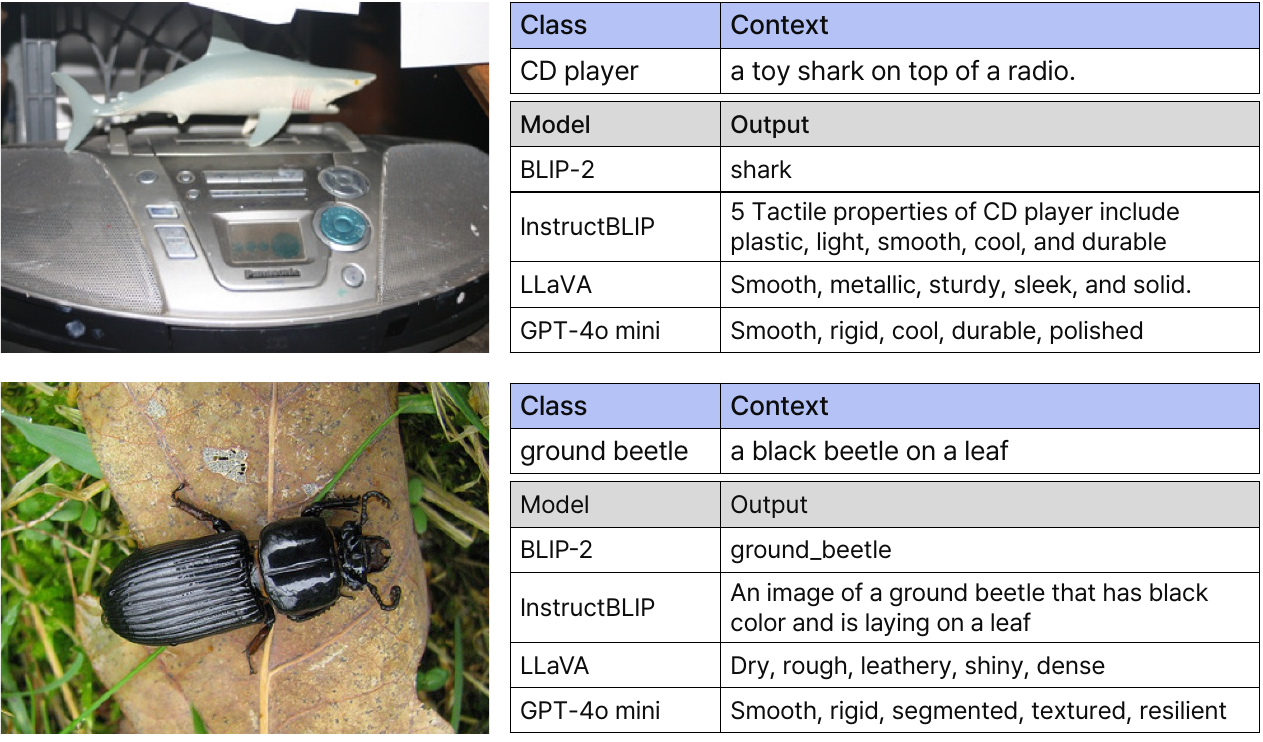}
    \vspace{-4mm}
    \caption{Recaptioning qualitative comparison with VLMs: BLIP-2 Opt-6.7B, InstructBLIP, and LLaVA-1.5 13B.}
    \label{fig:sup_caption_qualitative}
    \vspace{-4mm}
\end{figure}

To further support our quantitative findings, we present qualitative comparisons of the generated tactile-centric captions in Figure~\ref{fig:sup_caption_qualitative}. While the overall structure of captions remains similar across models, the subtle differences in the richness and specificity of tactile expressions highlight each model's varying degrees of tactile understanding. Notably, the captions generated by GPT-4o mini~\cite{Hurst2024GPT}, which achieved the highest scores in Table~\ref{tab:caption_comparison} and Table~\ref{tab:caption_quality}, also exhibit clearer and more coherent tactile semantics. This qualitative alignment with the quantitative results suggests that performance improvements are not merely numerical but also correspond to more accurate and meaningful tactile descriptions.

\vspace{-2mm}
\section{Illustration of Retriever and Integrator}

\begin{figure}[t]
    \centering
    \begin{minipage}[b]{1.0\linewidth}
        \centering
        \includegraphics[width=1.0\textwidth]{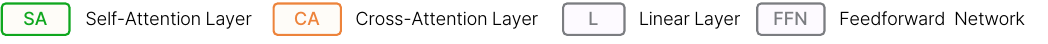}
    \end{minipage}
    \vspace{-5mm}
    \begin{minipage}[b]{0.495\linewidth}
        \centering
        \includegraphics[width=1.0\textwidth]{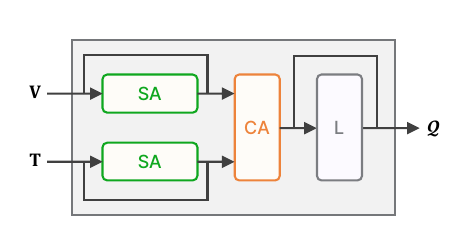}
        \subcaption{{\footnotesize Tactile-Guided Retriever}} \label{fig:sup_retriever}
    \end{minipage}
    \hfill
    \hfill
    \begin{minipage}[b]{0.495\linewidth}
        \centering
        \includegraphics[width=1.0\textwidth]{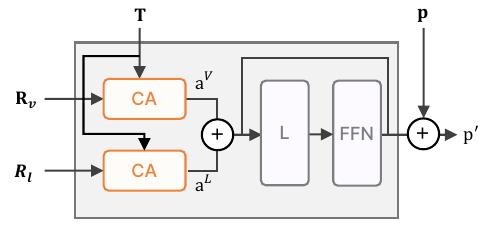}
        \subcaption{{\footnotesize Texture-Aware Integrator}} \label{fig:sup_integrator} 
    \end{minipage}
    \vspace{0.5mm}
    \caption{Illustraion of proposed modules. (a) The Tactile-Guided Retriever and (b) The Texture-Aware Integrator.}
    \vspace{-4mm}
    \label{fig:module_arch}
\end{figure}

\noindent \textbf{Tactile-Guided Retriever} The Tactile-Guided Retriever, in Figure~\ref{fig:sup_retriever}, integrates visual $\mathbf{V}$ and tactile $\mathbf{T}$ features to generate retrieval queries. Each modality first passes through a self-attention (SA) module to capture intra-modal dependencies. The resulting features are then fused via a cross-attention (CA) mechanism, enabling the tactile input to guide the integration with visual cues. Finally, a lightweight projection layer (L) transforms the fused representation into a retrieval query vector $\mathcal{Q}$. This query is then used to retrieve semantically relevant $K$ visual featuers $\mathbf{R}_{v} = \{r^k_{v}\}^{K}_{k=1}$ and and text features $\mathbf{R}_{l} = \{r^k_{l}\}^{K}_{k=1}$ from ImageNet-T, a vision-language knowledge dataset recaptioned with a tactile-centric perspective.

\begin{table}
\centering
\footnotesize
\begin{tcolorbox}[colback=gray!10, colframe=black!30, width=0.9\linewidth, boxrule=0.4pt, arc=2mm]
\begin{flushleft}
Your task is to evaluate and rate the tactile caption \green{on a scale of 1.0 to 10.0} based on the given Grading Criteria. \\
\vspace{0.5em}
Grading Criteria:\\
1.0: The caption does not describe any tactile feelings of the object in the image at all. \\
10.0: The caption accurately and clearly describes the tactile feelings of the object in the image. \\
\vspace{0.5em}
Class: \green{\texttt{\{class\_name\}}} \\
Caption: \green{\texttt{\{caption\}}} \\
\vspace{0.5em}
Score(Choose a rating from 1.0 to 10.0, provide only the number):
\end{flushleft}
\end{tcolorbox}
\caption{Evaluation prompt to measure tactile relevance.}
\label{tab:gpt_eval_templete}
\vspace{-6mm}
\end{table}

\begin{table}[t]
\centering
\footnotesize
\renewcommand{\arraystretch}{1.2}
\begin{adjustbox}{width=0.99\linewidth}
    \begin{tabular}{
      >{\raggedright\arraybackslash}m{2.2cm} |
      >{\centering\arraybackslash}p{1.2cm}
      >{\centering\arraybackslash}p{1.2cm}
      >{\centering\arraybackslash}p{1.2cm}
    }
    \toprule
    \textbf{Model} & \textbf{SSVTP} & \textbf{HCT} & \textbf{TVL} \\
    \midrule
    BLIP-2 Opt-6.7B~\cite{Li2023BLIP2} & 6.42 & 5.01 & 5.17 \\
    InstructBLIP 13B~\cite{instructblip} & 6.60 & 5.04 & 5.22 \\
    LLaVA-1.5 13B~\cite{liu2024improved} & 6.60 & 5.06 & 5.23 \\
    \rowcolor{mycolor} \textbf{GPT-4o mini}~\cite{Hurst2024GPT} & \textbf{6.73} & \textbf{5.13} & \textbf{5.32} \\
    \bottomrule
    \end{tabular}
\end{adjustbox}
\caption{Performance comparison of captioning models.}
\label{tab:caption_comparison}
\vspace{-6mm}
\end{table}

\begin{table}[t]
\centering
\footnotesize
\renewcommand{\arraystretch}{1.2}
\begin{adjustbox}{width=0.99\linewidth}
    \begin{tabular}{
      >{\raggedright\arraybackslash}m{1.2cm} |
      >{\centering\arraybackslash}p{1.2cm}
      >{\centering\arraybackslash}p{1.2cm}
      >{\centering\arraybackslash}p{1.2cm}
      >{\centering\arraybackslash}p{1.5cm}
    }
    \toprule
    \textbf{Model} & \textbf{BLIP-2} & \textbf{InstructBLIP} & \textbf{LLaVA} & \textbf{GPT-4o mini} \\
    \midrule
    Score (1-10) & 1.36 & 3.11 & 6.31 & \textbf{6.84} \\
    \bottomrule
    \end{tabular}
\end{adjustbox}
\caption{Performance comparison of VLMs for tactile relevance: BLIP-2 Opt-6.7B, InstructBLIP, and LLaVA1.5 13B.}
\label{tab:caption_quality}
\vspace{-6mm}
\end{table}

\vspace{0.5mm}
\noindent \textbf{Texture-Aware Integrator} The Texture-Aware Integrator, in Figure~\ref{fig:sup_integrator}, is designed to enrich tactile understanding by selectively integrating information from retrieved features $\mathbf{R}_v$ and $\mathbf{R}_l$. Given tactile features $\mathbf{T}$ as the core signal, the integrator extracts texture-relevant features $\mathbf{a}^V$ and $\mathbf{a}^L$ from the retrieved visual and language representations $\mathbf{R}_v$ and $\mathbf{R}_l$, respectively. Specifically, it employs cross-modal attention mechanisms (CA) conditioned on $\mathbf{T}$ to emphasize tactile-aligned semantics. It allows the model to selectively attend to texture-relevant cues while filtering out modality-specific noise such as background information or irrelevant textual information. The resulting fused feature is then passed through a linear layer and a feedforward network to align with the visual prompt $\mathbf{p}$ from the Projection Layer in TVL-LLaMA~\cite{Fu2024ATV}. Finally, it is combined with $\mathbf{p}$ to form an enhanced prompt $\mathbf{p}'$, which is used as the input to the LLaMA-2~\cite{touvron2023llama2openfoundation} for generating open-vocabulary texture descriptions.

\section{Qualitative Results of Retriever}

\begin{figure*}[t]
    \centering
    \begin{subfigure}[b]{0.92\textwidth}
        \centering
        \includegraphics[width=0.92\textwidth]{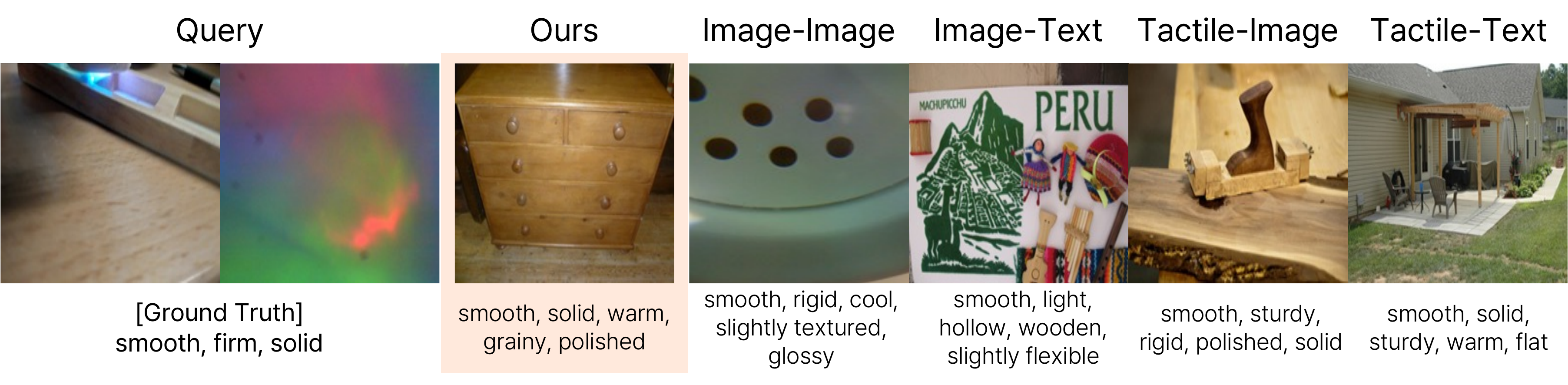}
        \vspace{-2mm}
        \subcaption{Case 1}
        \label{fig:sup_retrieval_comp_1}
    \end{subfigure}
    \begin{subfigure}[b]{0.92\textwidth}
        \centering
        \includegraphics[width=0.92\textwidth]{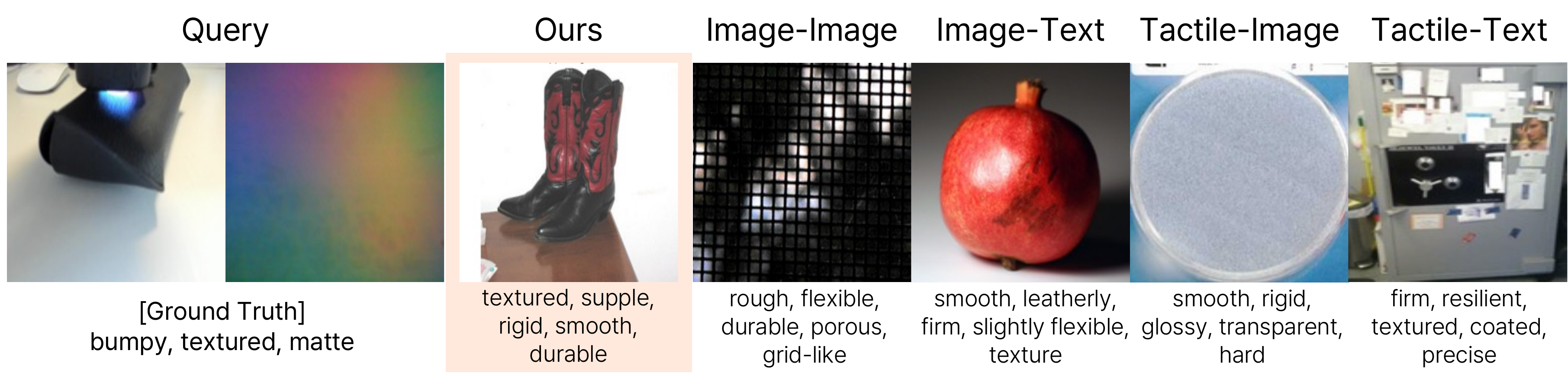}
        \vspace{-2mm}
        \subcaption{Case 2}
        \label{fig:sup_retrieval_comp_2}
    \end{subfigure}
    \begin{subfigure}[b]{0.92\textwidth}
        \centering
        \includegraphics[width=0.92\textwidth]{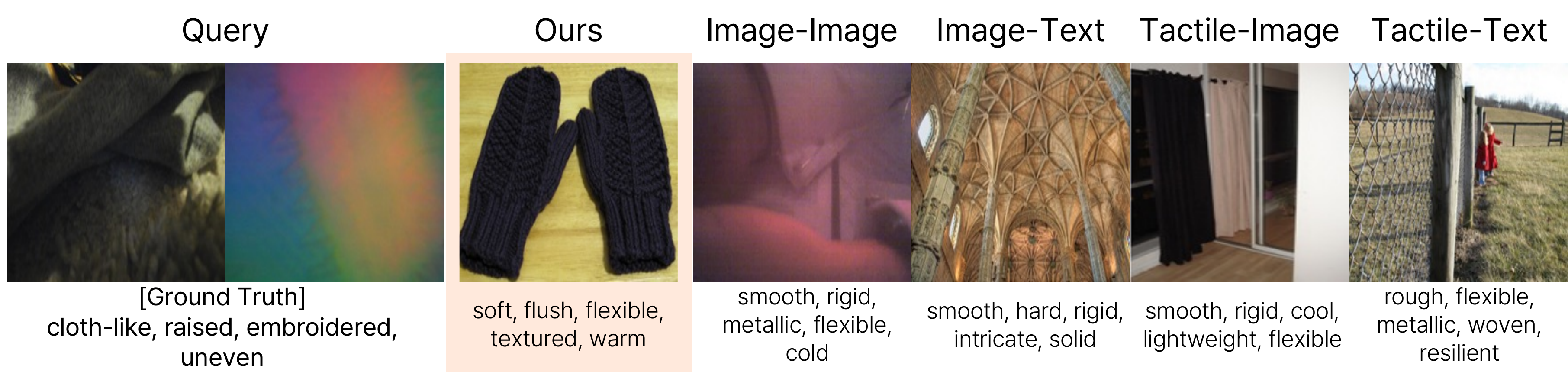}
        \vspace{-2mm}
        \subcaption{Case 3}
        \label{fig:sup_retrieval_comp_3}
    \end{subfigure}
    \vspace{-3mm}
    \caption{Qualitative comparison of retrieval methods.}
    \label{fig:sup_retrieval_comp}
    \vspace{-3mm}
\end{figure*}

\subsection{Compare to Alternative Retrieval Methods}
To qualitatively assess the effectiveness of Tactile-Guided Retriever, we visualize the Top-1 retrieved samples from TVL dataset~\cite{Fu2024ATV} for three randomly selected visuo-tactile input pairs in Figure~\ref{fig:sup_retrieval_comp}. We compare the performance of the retriever against five alternative settings: Image-Image, Image-Text, Tactile-Image, and Tactile-Text. We present the most semantically relevant vision-language sample pair for each setting retrieved from the ImageNet-T.

\vspace{0.5mm}
\noindent \textbf{Case 1.} As shown in Figure~\ref{fig:sup_retrieval_comp_1}, Image-Image relies on visual glossiness, retrieving a shiny but texture-irrelevant object, while Tactile-Image performs slightly better by retrieving a wooden object with some tactile similarity. On the other hand, Image-Text and Tactile-Text retrieve background-heavy samples with minimal tactile alignment, as they rely solely on unimodal cues. In contrast, our method retrieves a visually dissimilar but texturally aligned wooden drawer.

\vspace{0.5mm}
\noindent \textbf{Case 2.} Also in the second sample, Figure~\ref{fig:sup_retrieval_comp_2}, Image-Image retrieves a visually similar object in color scale with a grid-like texture, focusing on surface pattern rather than material. Tactile-Image retrieves a flat, smooth object, which partially matches the tactile feel but fails to capture the material characteristics of leather. Meanwhile, Image-Text and Tactile-Text result in semantically distant samples, such as a fruit or a vault, due to limited cross-modal grounding capabilities. In contrast, our method retrieves a pair of leather boots that share similar material properties with the query.

\vspace{0.5mm}
\noindent \textbf{Case 3.} Lastly, as shown in Figure~\ref{fig:sup_retrieval_comp_3}, 
Image-Image retrieves a visually similar fabric object, capturing some resemblance in material category, but it fails to reflect the detailed knitted texture of the query. Tactile-Image retrieves a flat curtain, but the result is dominated by background context and lacks tactile grounding. Interestingly, Image-Text and Tactile-Text retrieve samples that resemble the knitted pattern to some extent, but they lack visual or material grounding, resulting in semantically unrelated scenes. In contrast, our proposed method retrieves a pair of knitted mittens that closely match the query's material and texture.

\begin{figure*}[t]
    \centering
    \begin{subfigure}[b]{0.92\textwidth}
        \centering
        \includegraphics[width=0.92\textwidth]{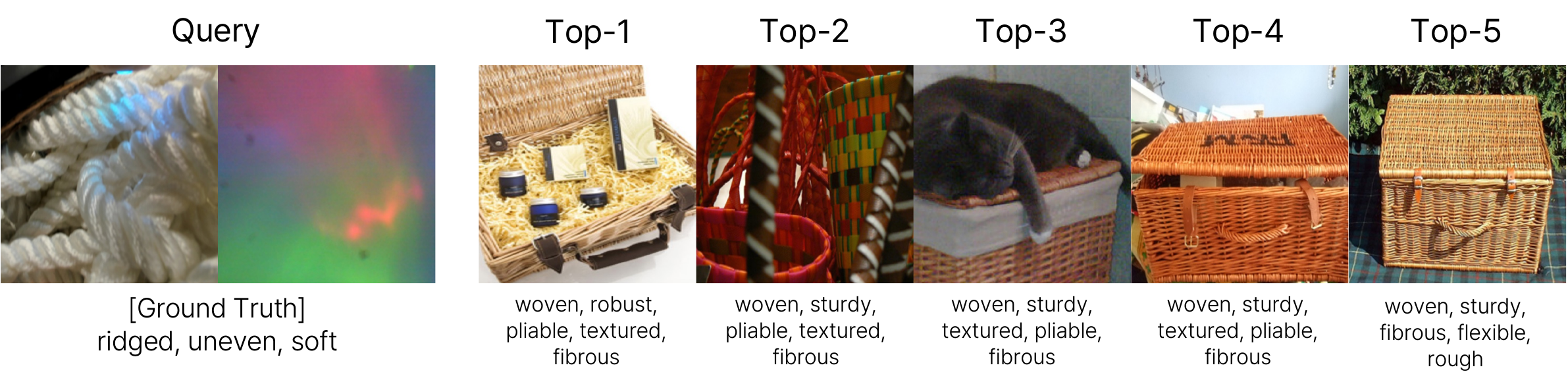}
        \vspace{-2mm}
        \subcaption{}
        \label{fig:sup_retrieval_valid_1}
    \end{subfigure}
    \begin{subfigure}[b]{0.92\textwidth}
        \centering
        \includegraphics[width=0.92\textwidth]{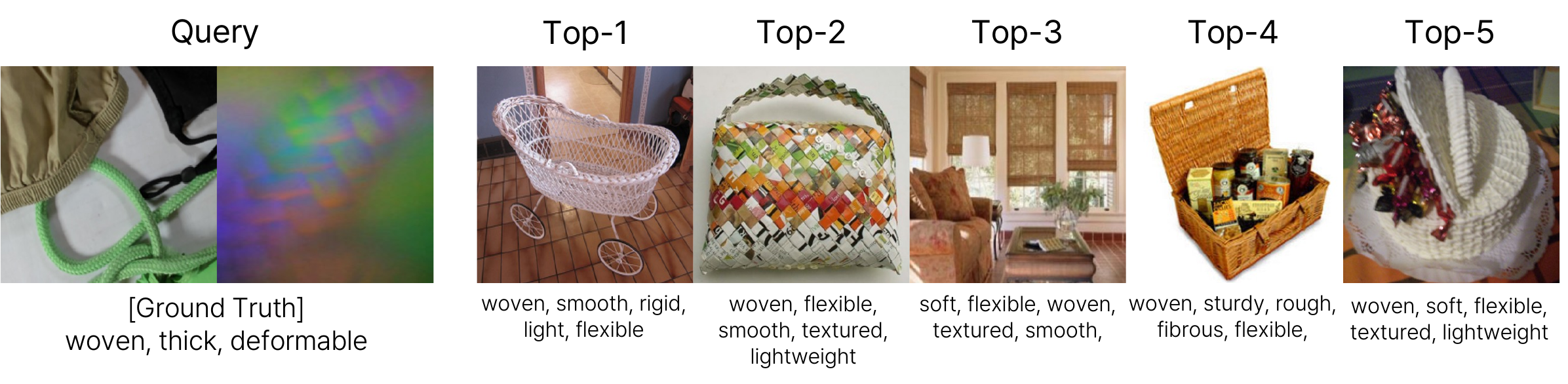}
        \vspace{-2mm}
        \subcaption{}
        \label{fig:sup_retrieval_valid_2}
    \end{subfigure}
    \vspace{-3mm}
    \caption{Valid cases of retriever from TVL dataset.}
    \label{fig:sup_retrieval_valid}
    \vspace{-4mm}
\end{figure*}

\begin{figure*}[t]
    \centering
    \begin{subfigure}[b]{0.95\textwidth}
        \centering
        \includegraphics[width=0.95\textwidth]{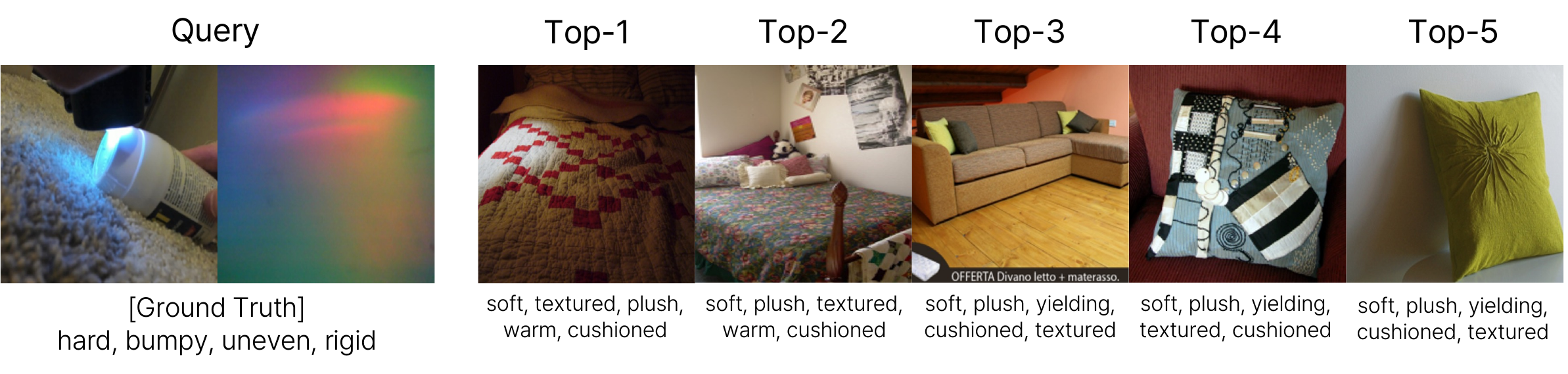}
        \vspace{-2mm}
        \subcaption{}
        \label{fig:sup_retrieval_failure_1}
    \end{subfigure}
    \begin{subfigure}[b]{0.95\textwidth}
        \centering
        \includegraphics[width=0.95\textwidth]{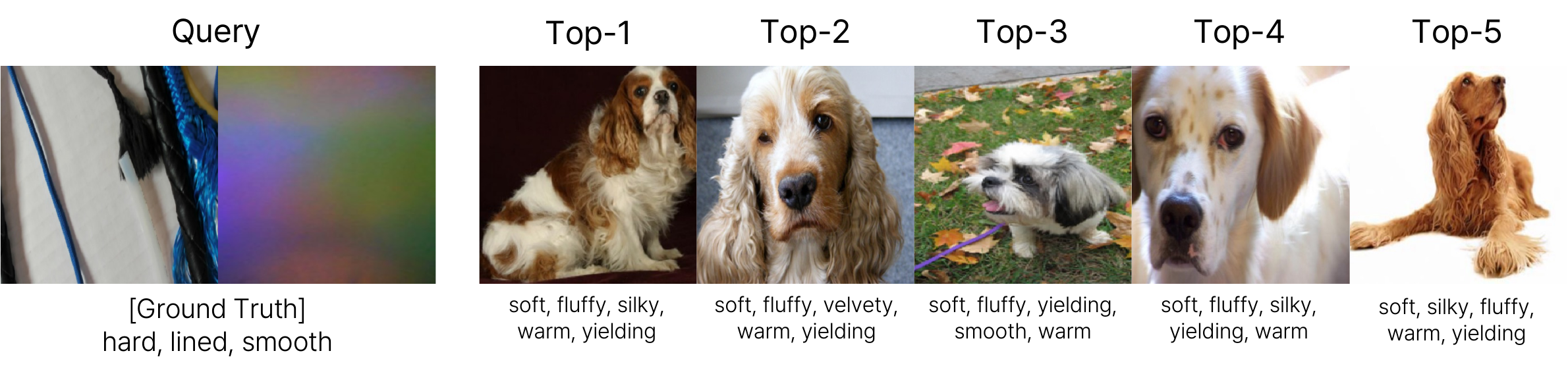}
        \vspace{-2mm}
        \subcaption{}
        \label{fig:sup_retrieval_failure_2}
    \end{subfigure}
    \vspace{-3mm}
    \caption{Failure cases of retriever from TVL dataset.}
    \label{fig:sup_retrieval_failure}
    \vspace{-3mm}
\end{figure*}

\subsection{Valid Cases}
As shown in Figure~\ref{fig:sup_retrieval_valid}, the Tactile-Guided Retriever successfully retrieves top-5 samples that share key material properties and texture patterns with the query inputs. In the first example, in Figure~\ref{fig:sup_retrieval_valid_1}, where the query consists of intertwined ropes, the model retrieves objects such as hampers and baskets that exhibit coarse, fibrous, and woven textures. These results reflect tactile properties like rigidity, pliability, and surface roughness that align closely with the query's physical characteristics.

In the second example, in Figure~\ref{fig:sup_retrieval_valid_2}, the query depicts a soft, stretchable textile string. The retrieved items, such as bassinets, purses, and woven shades, consistently exhibit deformable and flexible properties. The associated descriptions frequently contain tactile-relevant terms such as `woven,' `smooth,' `textured,' and `lightweight,' demonstrating the model's ability to retrieve semantically meaningful and physically grounded results. Together, these examples highlight the effectiveness of our tactile-guided approach in retrieving samples that go beyond visual similarity and reflect material-aware semantics.

\subsection{Failure Cases}
We also analyze failure cases, as illustrated in Figure~\ref{fig:sup_retrieval_failure}, where the Tactile-Guided Retriever fails to retrieve texture-relevant samples. Although the retriever is designed to leverage tactile signals to guide the retrieval, it occasionally focuses on dominant visual patterns such as background information, especially when the object is visually small or not clearly localized. The tactile input may have emphasized surface roughness or grain in this example. Still, the corresponding region in the visual input lacks saliency, which leads the retriever to get a background-centric sample in both cases. This illustrates a remaining challenge in grounding tactile semantics when the visual cue is weak or spatially ambiguous.

\subsection{Insights from Qualitative Evaluation}
These qualitative comparisons demonstrate that our proposed tactile guided retriever effectively captures the material and texture semantics of the query, going beyond superficial visual similarity or unimodal cues. Compared to alternative unimodal baselines, our method consistently retrieves samples that better reflect the physical characteristics of the input, particularly in terms of texture, flexibility, and material composition. Although the retriever performs reliably across diverse scenarios, such as those involving coarse woven surfaces and deformable fabrics, it may occasionally fail when the object of interest is visually small or overshadowed by dominant background elements. In conclusion, the overall results highlight the effectiveness of our retrieval strategy in leveraging both visual and tactile features to ground retrievals in semantically rich and physically meaningful ways.

\section{ImageNet-T Dataset}
\subsection{Recaptioning Template}
We used the prompt shown in Table~\ref{tab:sup_recaption_template} to generate tactile-centric captions of ImageNet-T. This prompt is carefully designed to instruct vision-language models to focus solely on describing tactile attributes such as texture, material feel, and surface patterns relevant to touch. The expected output also aligns with the TVL~\cite{Fu2024ATV} caption style, which consists of five tactile adjectives separated by commas, to reduce the domain gap.
\begin{table}
\centering
\footnotesize
\begin{tcolorbox}[colback=gray!10, colframe=black!30, width=0.95\linewidth, boxrule=0.4pt, arc=2mm]
\begin{flushleft}
\#\# Task \\
Create a tactile caption for an object in the given image based on its class name and an image description. \\
Class: \green{\texttt{\{class\_name\}}} \\
Description: \green{\texttt{\{caption\}}} \\
\vspace{0.5em}

\#\# Instructions \\
1. Provide exactly \green{5 adjectives} that refer solely to how the object feels to the touch--focusing on texture, flexibility, density, and material properties. \\
2. Try to include more varied and nuanced tactile descriptors. \\
3. Do \green{not} include adjectives related to \green{visual
appearance, shape, color, temperature, sound, weight}, or any non-tactile properties or \green{any non-tactile properties}. \\
4. Respond using the exact format: "adj1, adj2, adj3, adj4, adj5". \\
\vspace{0.5em}
Remember: Your ENTIRE response must be ONLY 5 adjectives separated by commas. \\

\end{flushleft}
\end{tcolorbox}
\caption{Overview of prompt used for recaptioning.}
\label{tab:sup_recaption_template}
\vspace{-3mm}
\end{table}

\subsection{Distribution of Vocabulary Words}
\begin{figure}[t]
    \centering
    \includegraphics[width=0.98\linewidth]{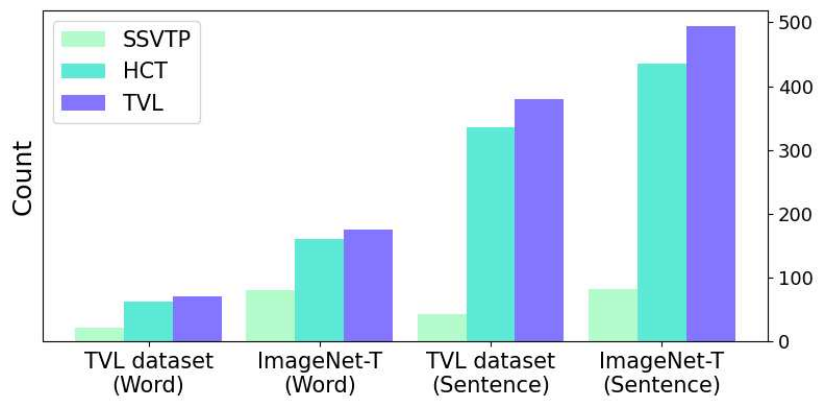}
    \vspace{-2mm}
    \caption{Comparison of unique word and sentence counts across TVL dataset (ground truth) and ImageNet-T (retrieved captions) for SSVTP, HCT, and TVL test sets.}
    \label{fig:word_count}
\end{figure}

\begin{figure*}[t]
    \centering
    \includegraphics[height=0.425\textheight]{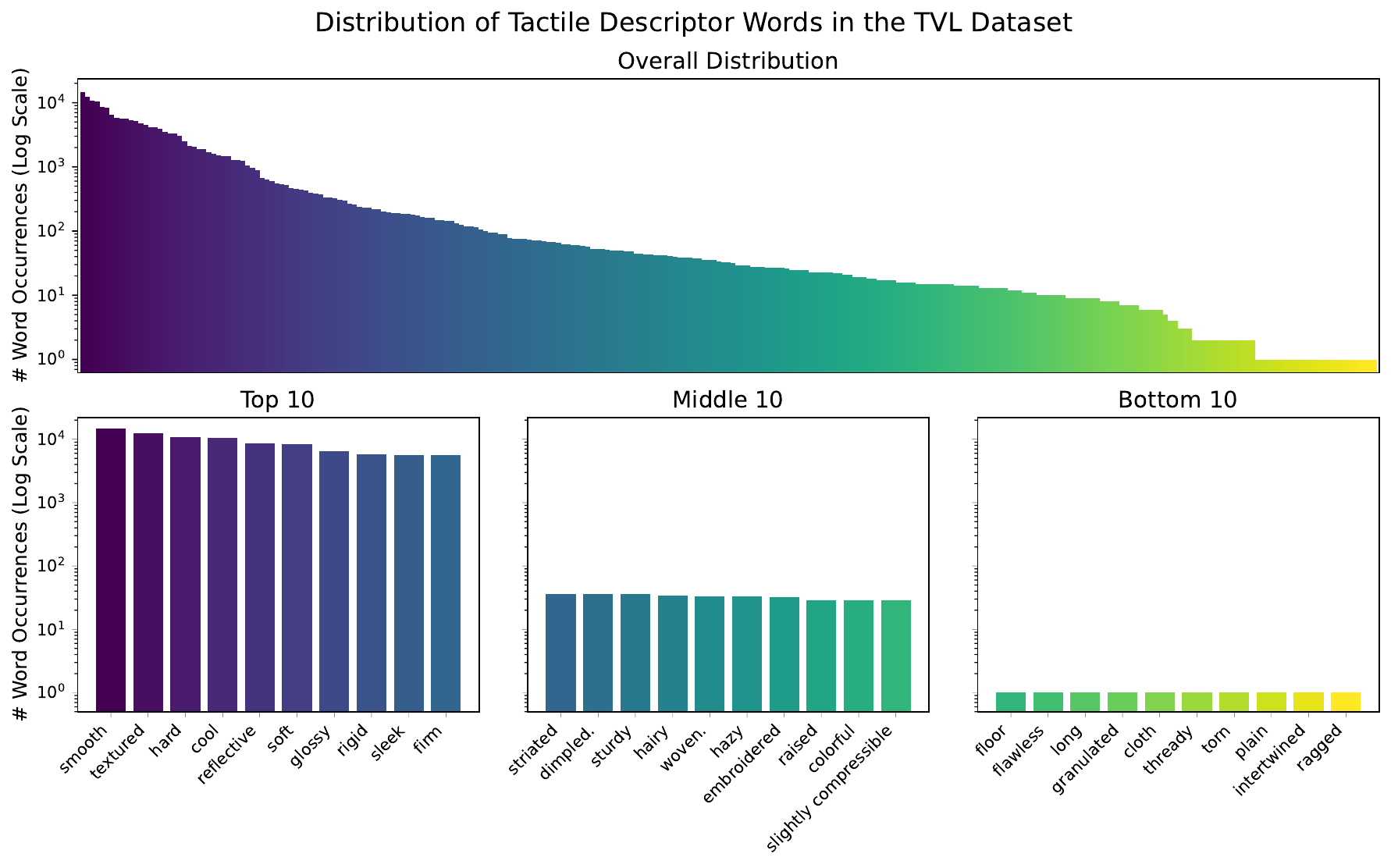}
    \vspace{-4mm}
    \caption{Distribution of words of TVL Dataset.}
    \label{fig:tvl_44k_word_hist}
\end{figure*}

\begin{figure*}[t]
    \centering
    \hspace{-0.8cm}
    \includegraphics[height=0.495\textheight]{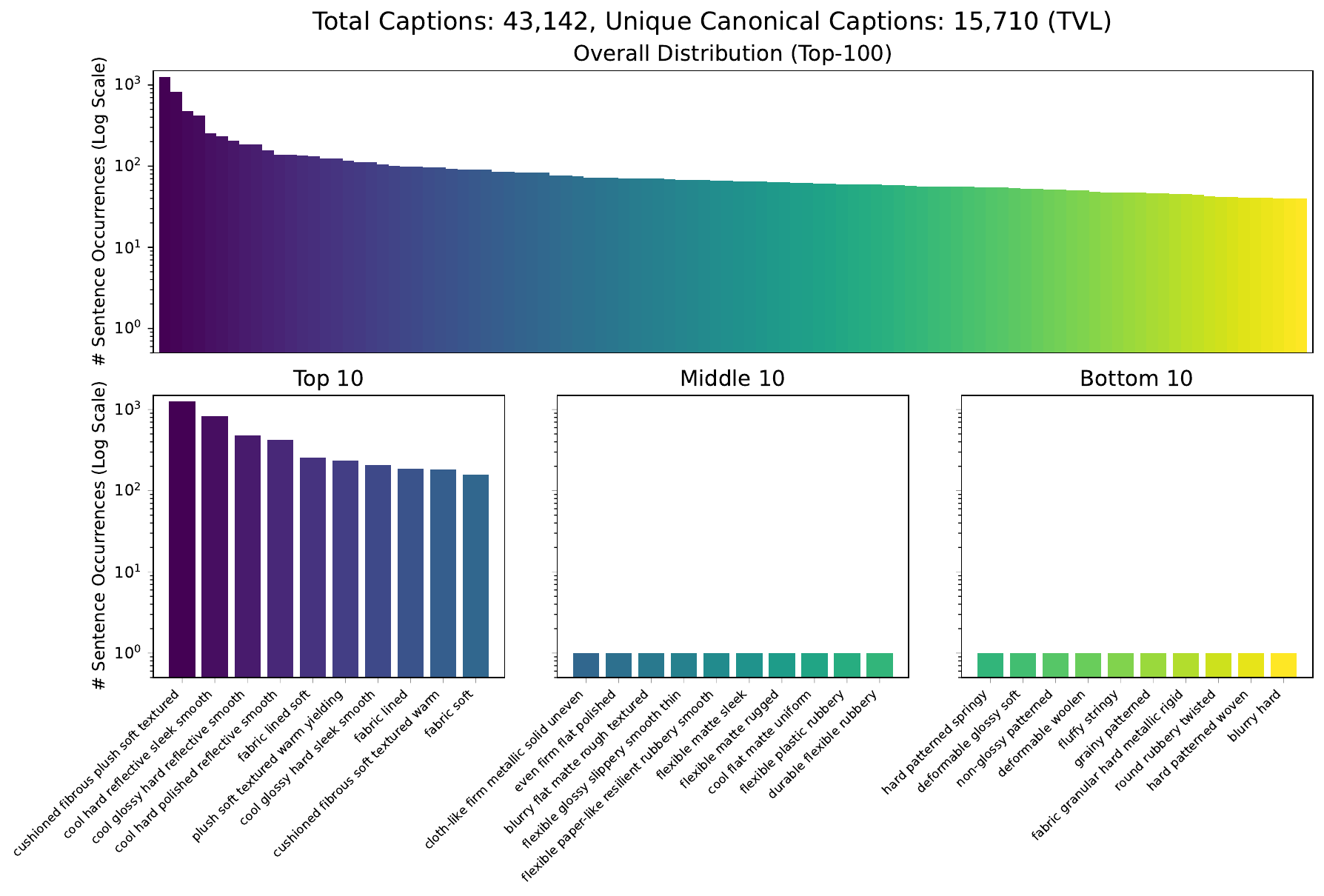}
    \vspace{-3mm}
    \caption{Distribution of Top-100 unique canonical captions of the TVL Dataset.}
    \label{fig:tvl_sentence_hist}
\end{figure*}

\begin{figure*}[t]
    \centering
    \includegraphics[height=0.425\textheight]{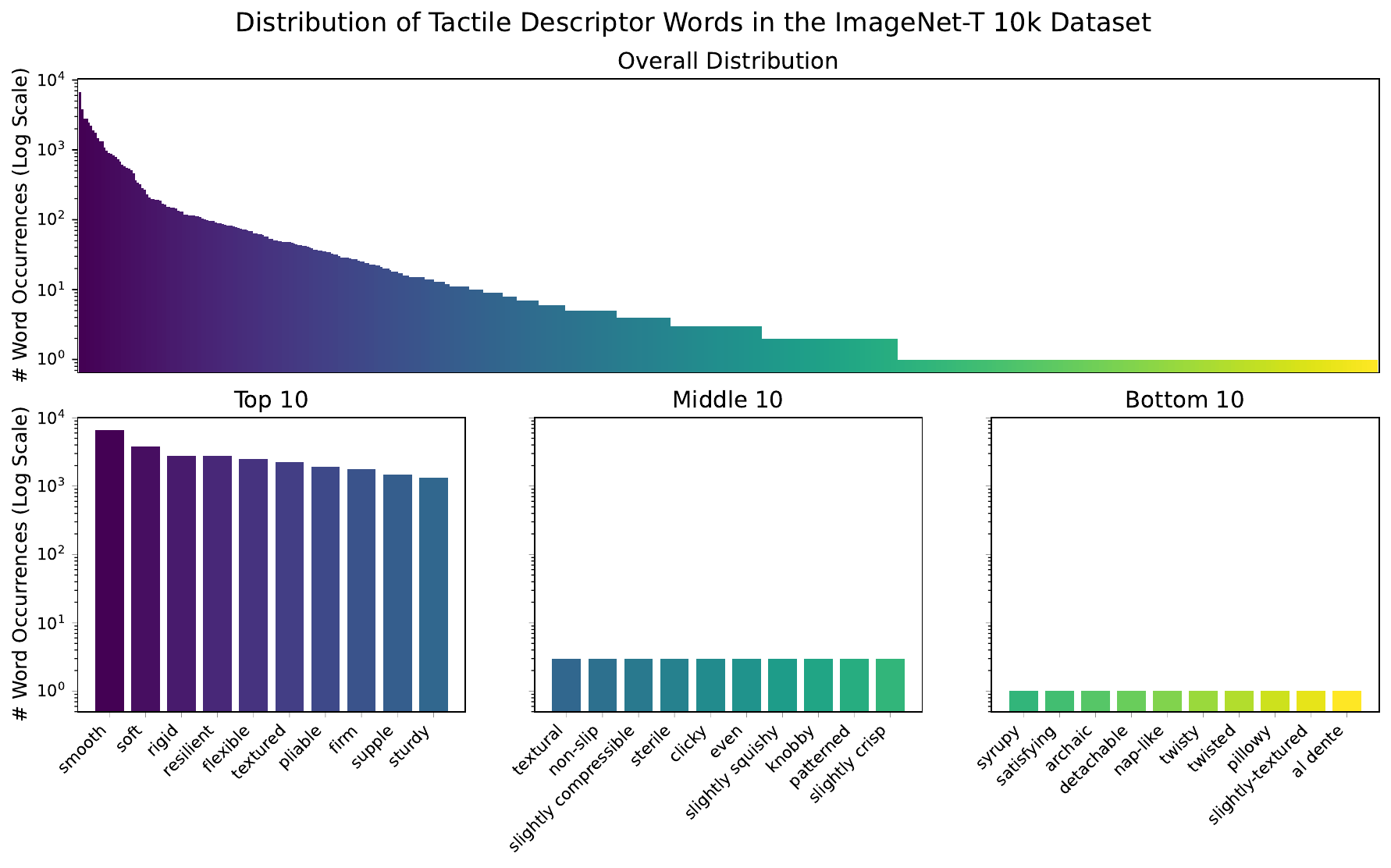}
    \vspace{-4mm}
    \caption{Distribution of words of the ImageNet-T 10k.}
    \label{fig:imagenet_t_10k_word_hist}
\end{figure*}

\begin{figure*}[t]
    \centering
    \hspace{-0.8cm}
    \includegraphics[height=0.495\textheight]{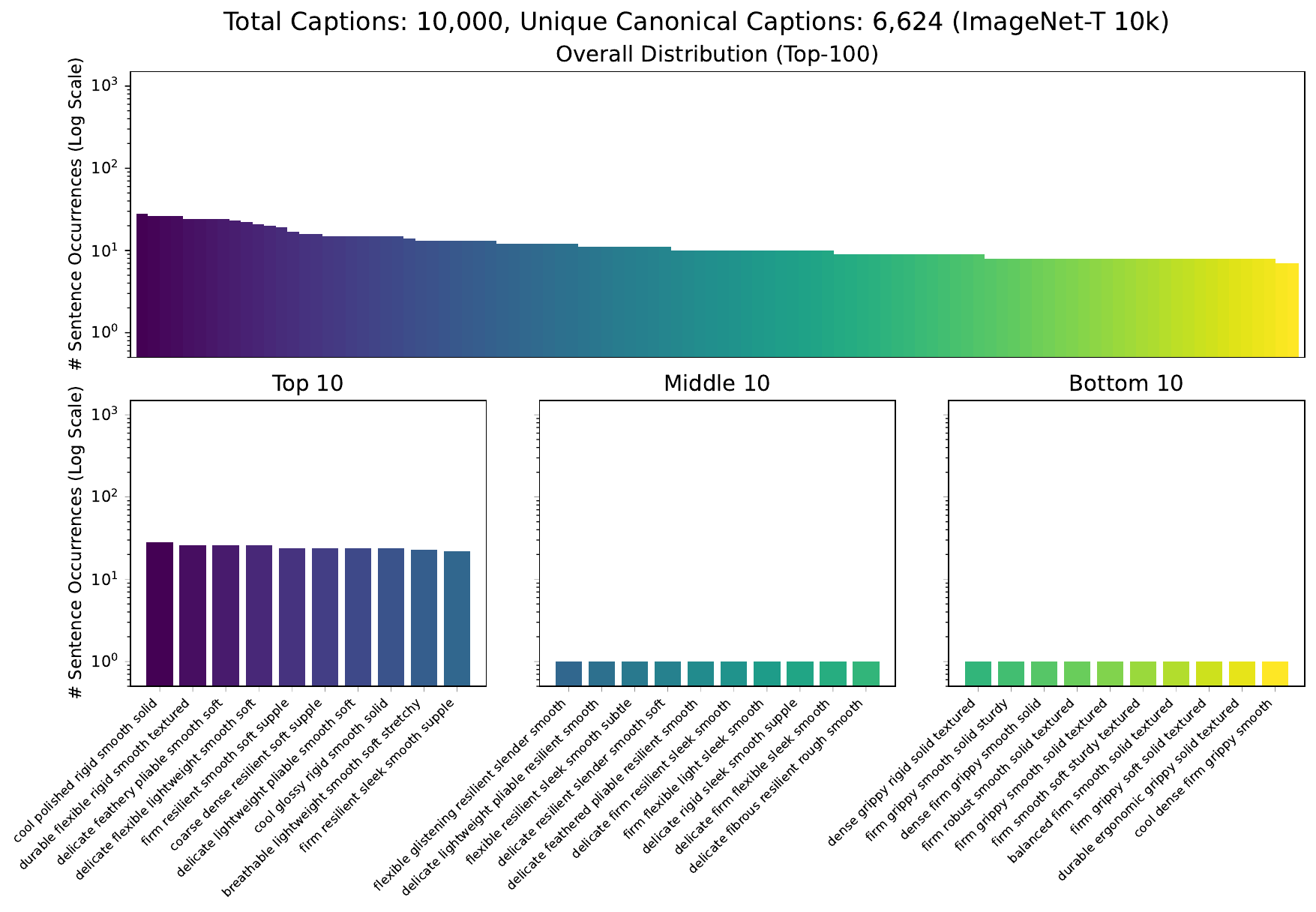}
    \vspace{-3mm}
    \caption{Distribution of Top-100 unique canonical captions of the ImageNet-T 10k.}
    \label{fig:imagenet_t_10k_sentence_hist}
\end{figure*}

\begin{figure*}[t]
    \centering
    \includegraphics[height=0.425\textheight]{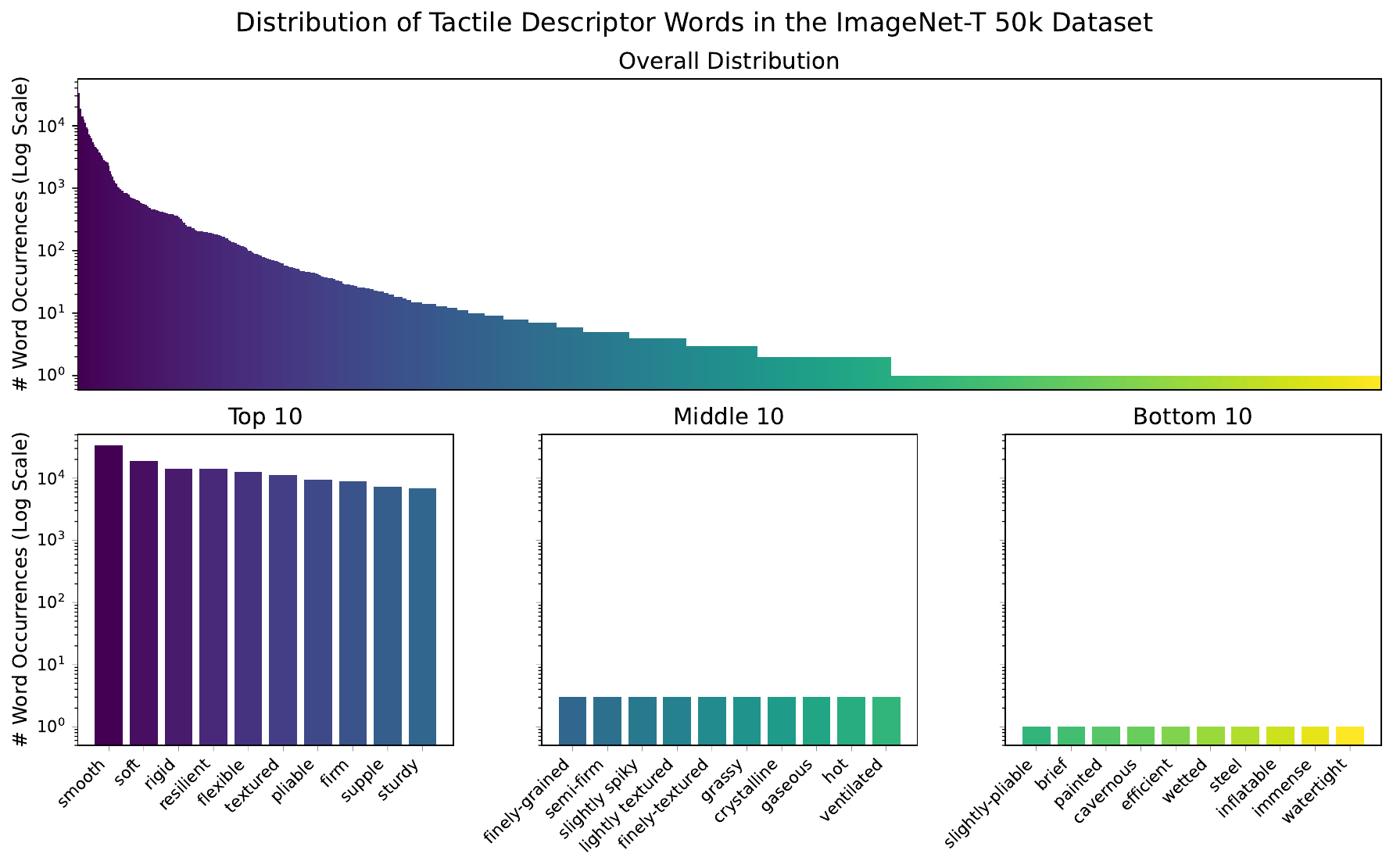}
    \vspace{-4mm}
    \caption{Distribution of words of the ImageNet-T 50k.}
    \label{fig:imagenet_t_50k_word_hist}
\end{figure*}

\begin{figure*}[t]
    \centering
    \hspace{-0.8cm}
    \includegraphics[height=0.495\textheight]{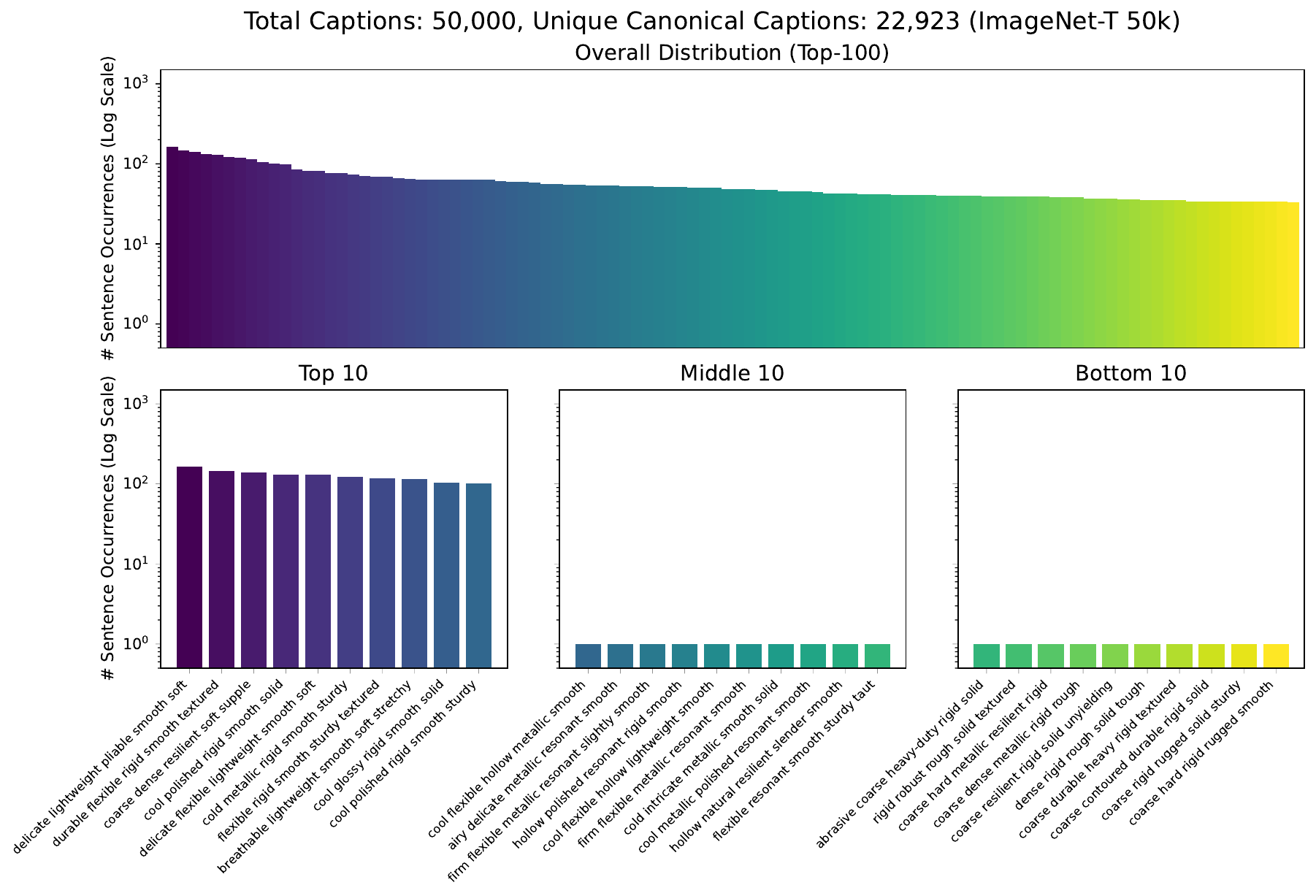}
    \vspace{-3mm}
    \caption{Distribution of Top-100 unique canonical captions of the ImageNet-T 50k.}
    \label{fig:imagenet_t_50k_sentence_hist}
\end{figure*}

\begin{figure*}[t]
    \centering
    \includegraphics[height=0.425\textheight]{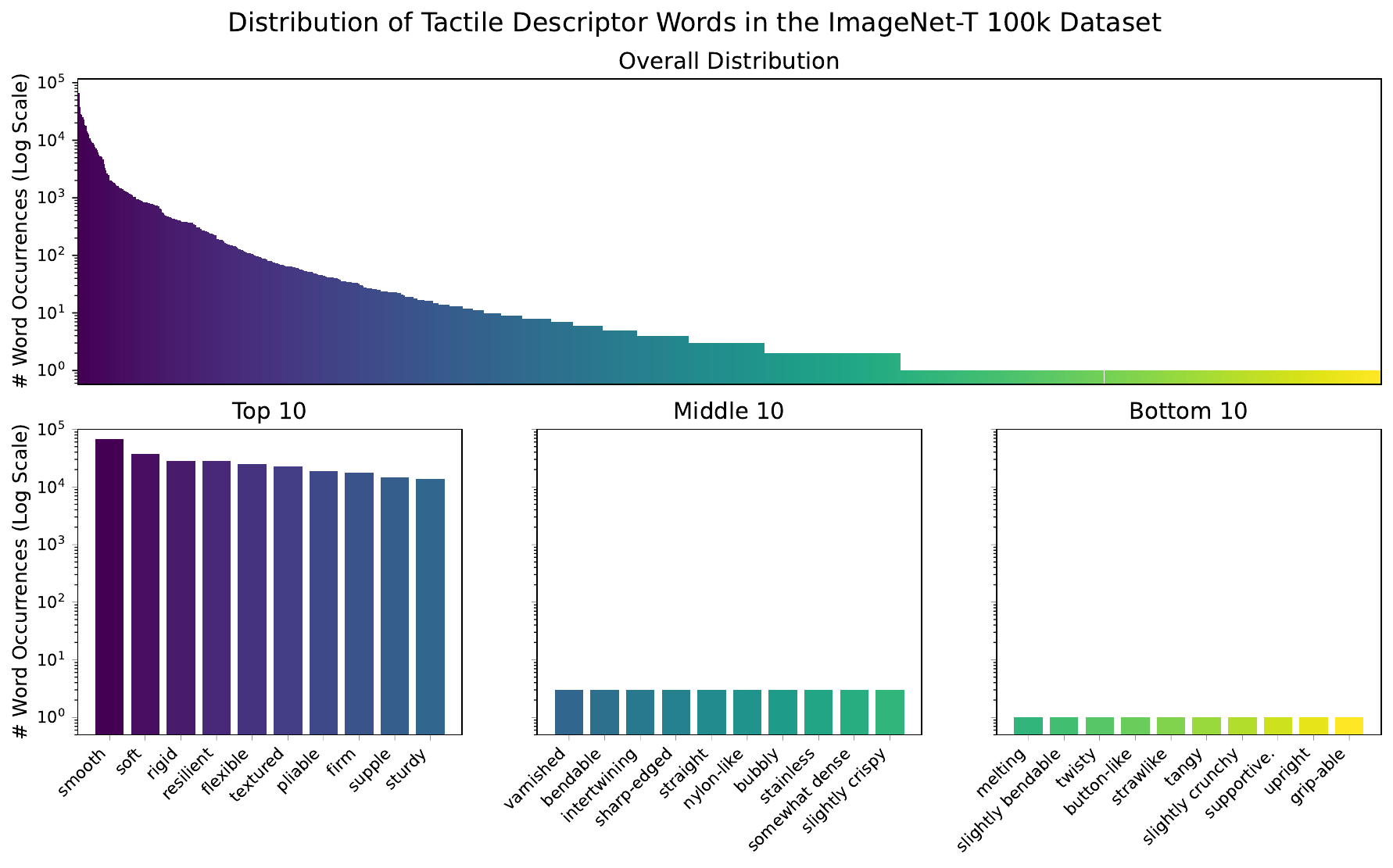}
    \vspace{-4mm}
    \caption{Distribution of words of the ImageNet-T 100k.}
    \label{fig:imagenet_t_100k_word_hist}
\end{figure*}

\begin{figure*}[t]
    \centering
    \hspace{-0.8cm}
    \includegraphics[height=0.495\textheight]{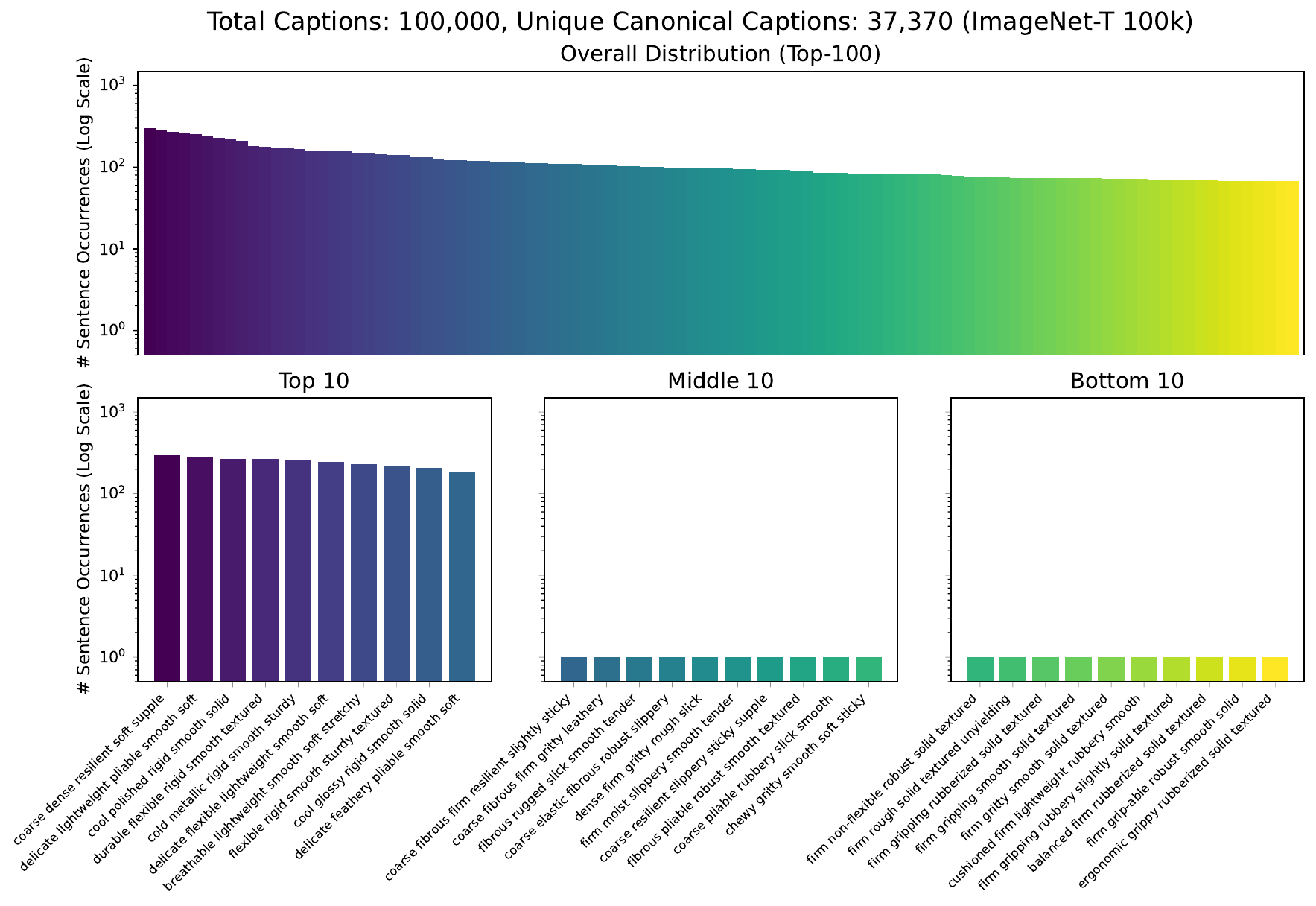}
    \vspace{-3mm}
    \caption{Distribution of Top-100 unique canonical captions of the ImageNet-T 100k.}
    \label{fig:imagenet_t_100k_sentence_hist}
\end{figure*}

\begin{figure*}[t]
    \centering
    \includegraphics[height=0.425\textheight]{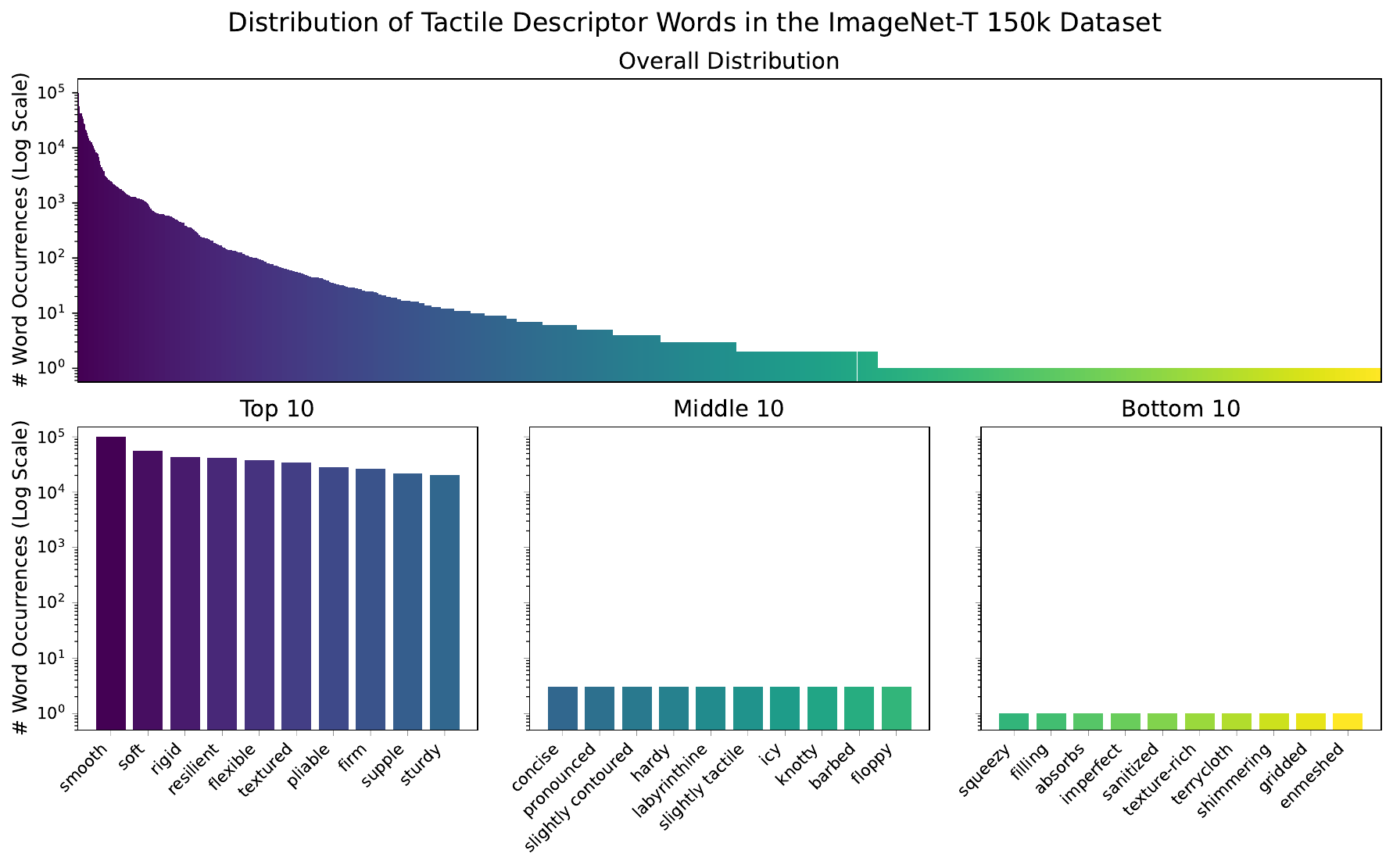}
    \vspace{-4mm}
    \caption{Distribution of words of the ImageNet-T 150k.}
    \label{fig:imagenet_t_150k_word_hist}
\end{figure*}

\begin{figure*}[t]
    \centering
    \hspace{-0.8cm}
    \includegraphics[height=0.495\textheight]{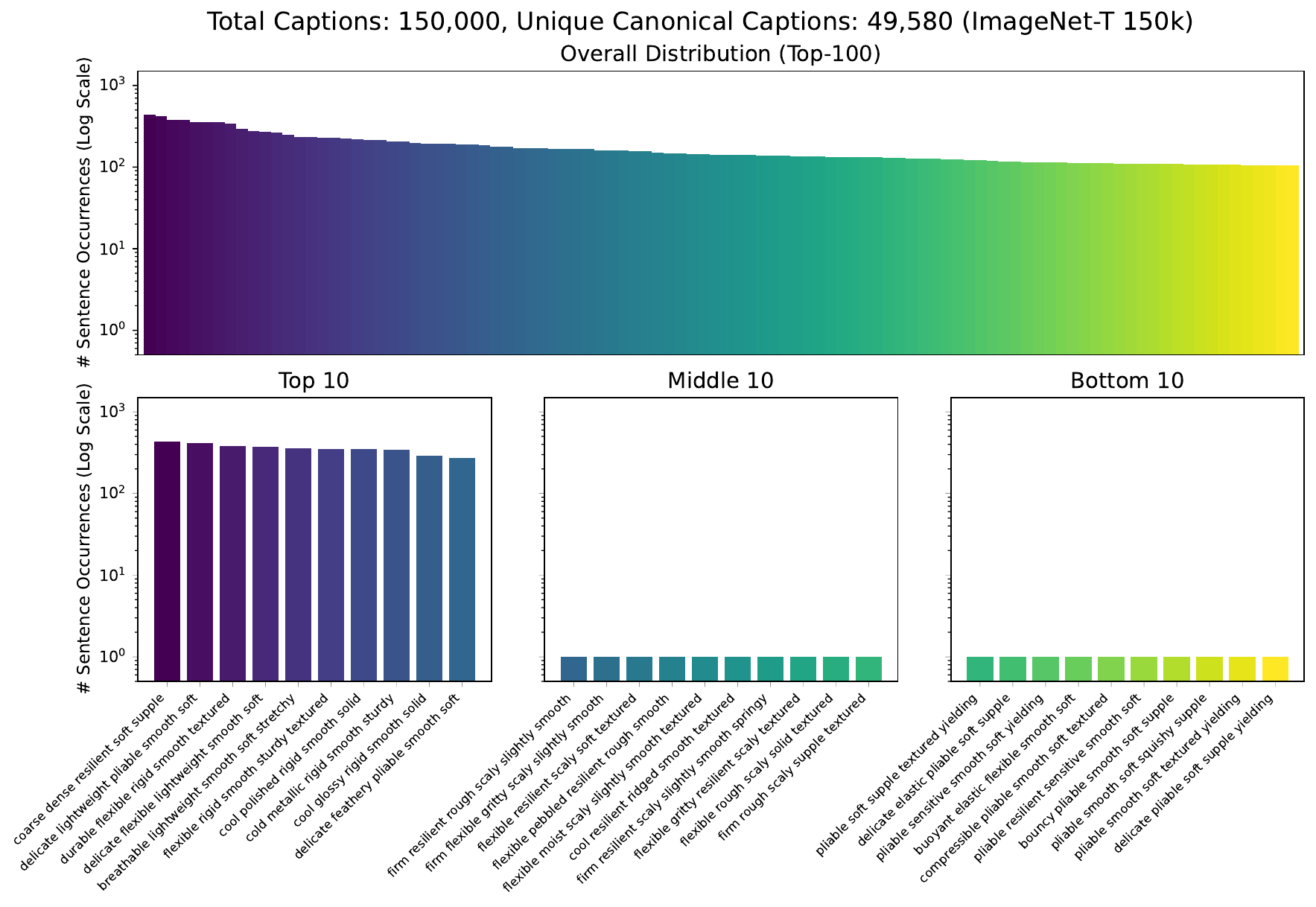}
    \vspace{-3mm}
    \caption{Distribution of Top-100 unique canonical captions of the ImageNet-T 150k.}
    \label{fig:imagenet_t_150k_sentence_hist}
\end{figure*}

In open-vocabulary tactile perception tasks, handling a wide range of words is important. Captions generated from external knowledge can help by offering more varied expressions. To improve how tactile concepts are described, we used ImageNet-T to produce diverse tactile-related phrases. We measured the number of unique words and sentences from the TVL~\cite{Fu2024ATV} test set and from the outputs retrieved by ImageNet-T. The results are shown in Figure~\ref{fig:word_count}. ImageNet-T produced more unique words and sentences than the original datasets. This is not just about quantity, as cosine similarity retrieves semantically similar but phrased differently. This helps describe tactile concepts in a richer and more flexible way, making ImageNet-T a useful source in open-vocabulary settings. Additional visualizations of the word and canonical sentence distributions are provided in Figure~\ref{fig:tvl_44k_word_hist} to Figure~\ref{fig:imagenet_t_150k_sentence_hist}.

\begin{figure*}[htbp]
    \centering
    \begin{subfigure}[b]{0.49\linewidth}
        \includegraphics[width=\linewidth]{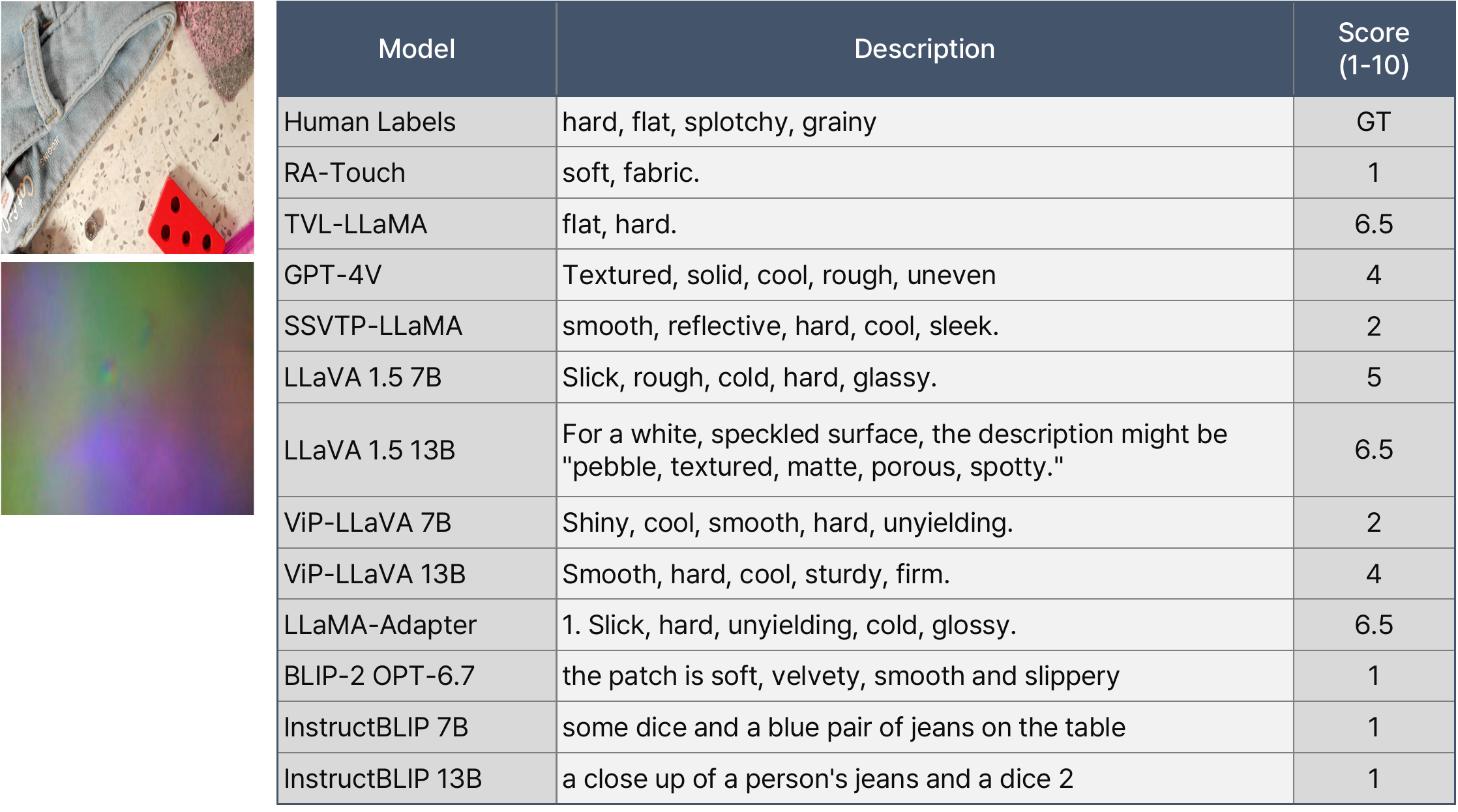}
    \end{subfigure}
    \hfill
    \begin{subfigure}[b]{0.49\linewidth}
        \includegraphics[width=\linewidth]{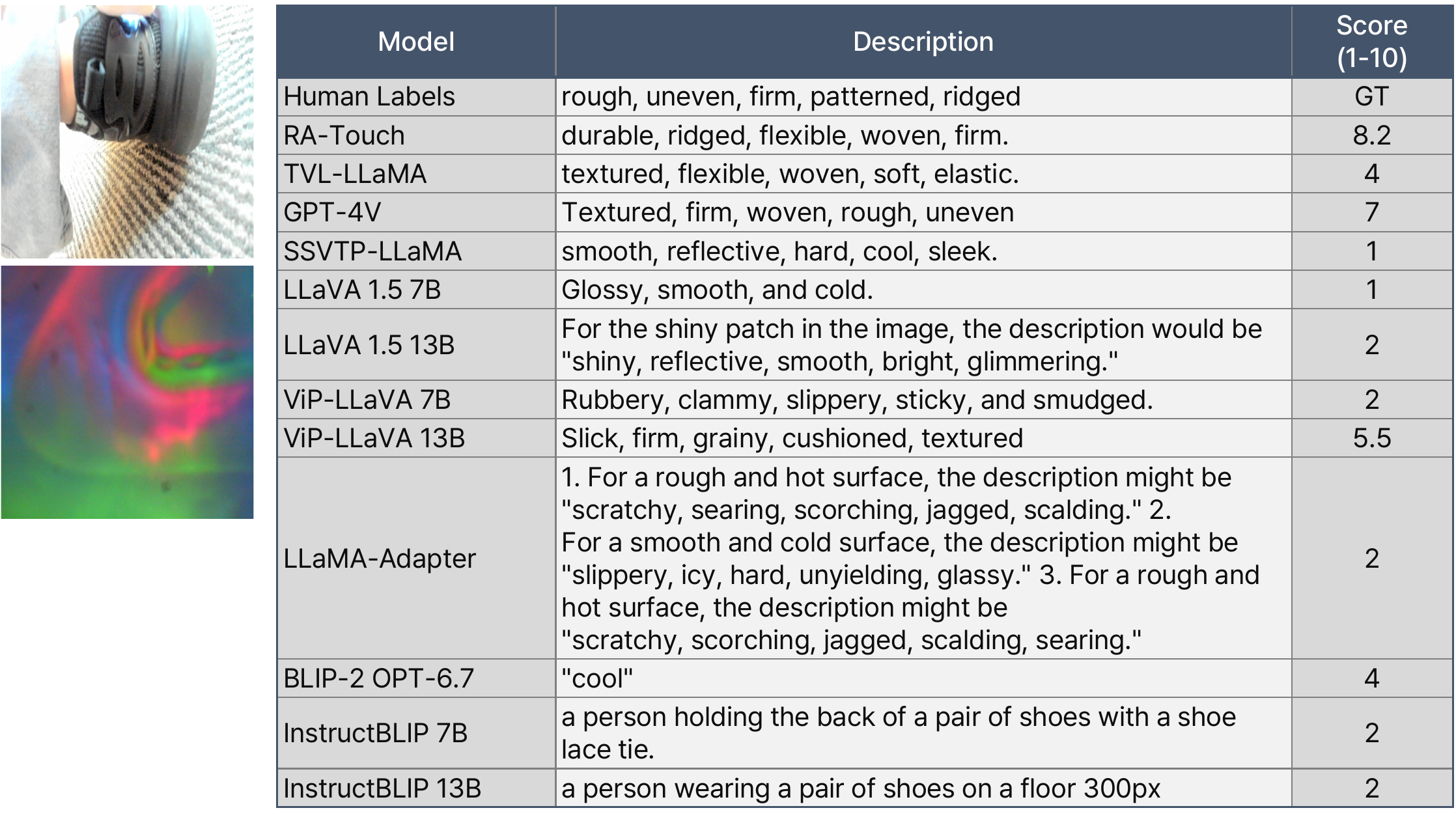}
    \end{subfigure}

    \vspace{8mm}

    \begin{subfigure}[b]{0.49\linewidth}
        \includegraphics[width=\linewidth]{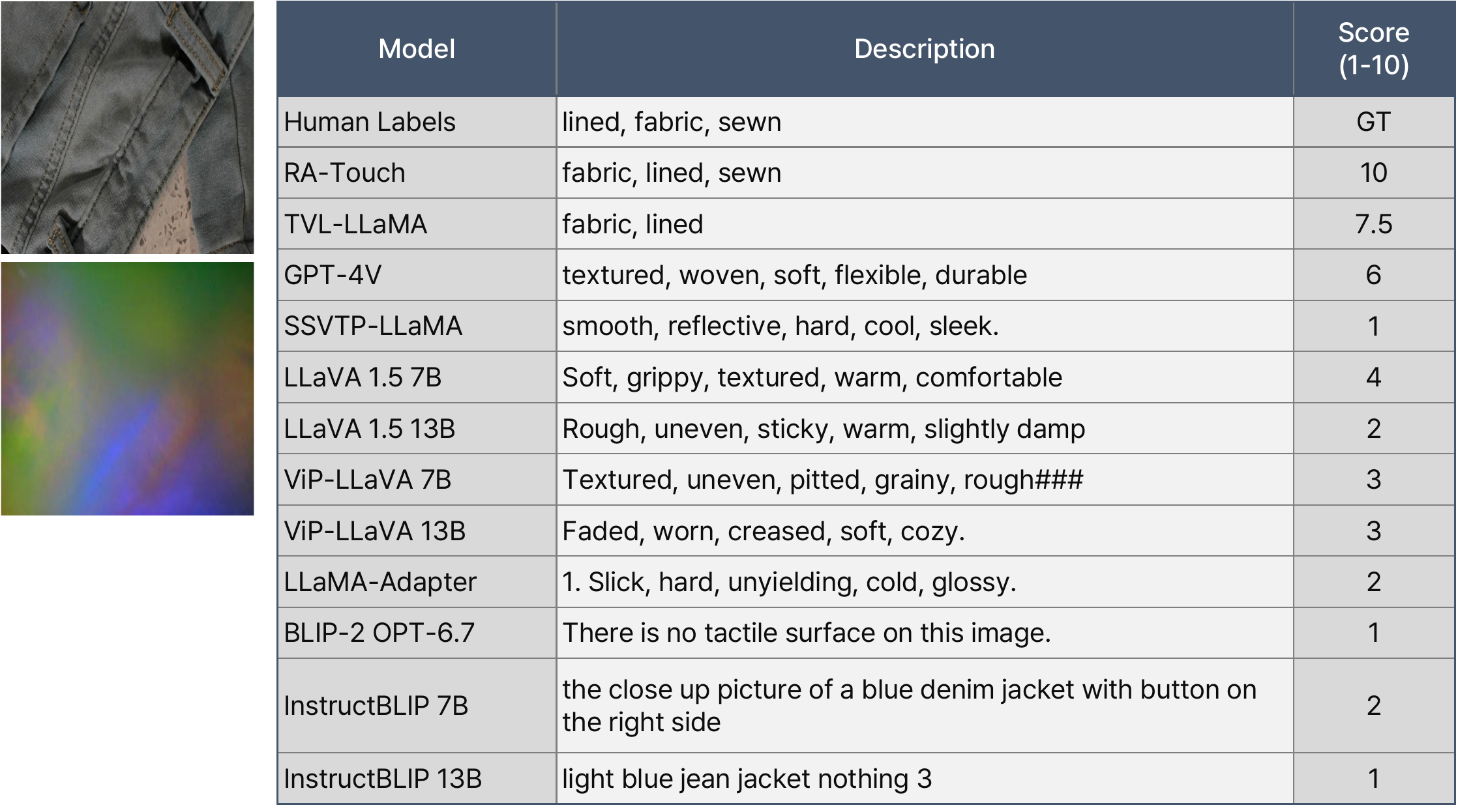}
    \end{subfigure}
    \hfill
    \begin{subfigure}[b]{0.49\linewidth}
        \includegraphics[width=\linewidth]{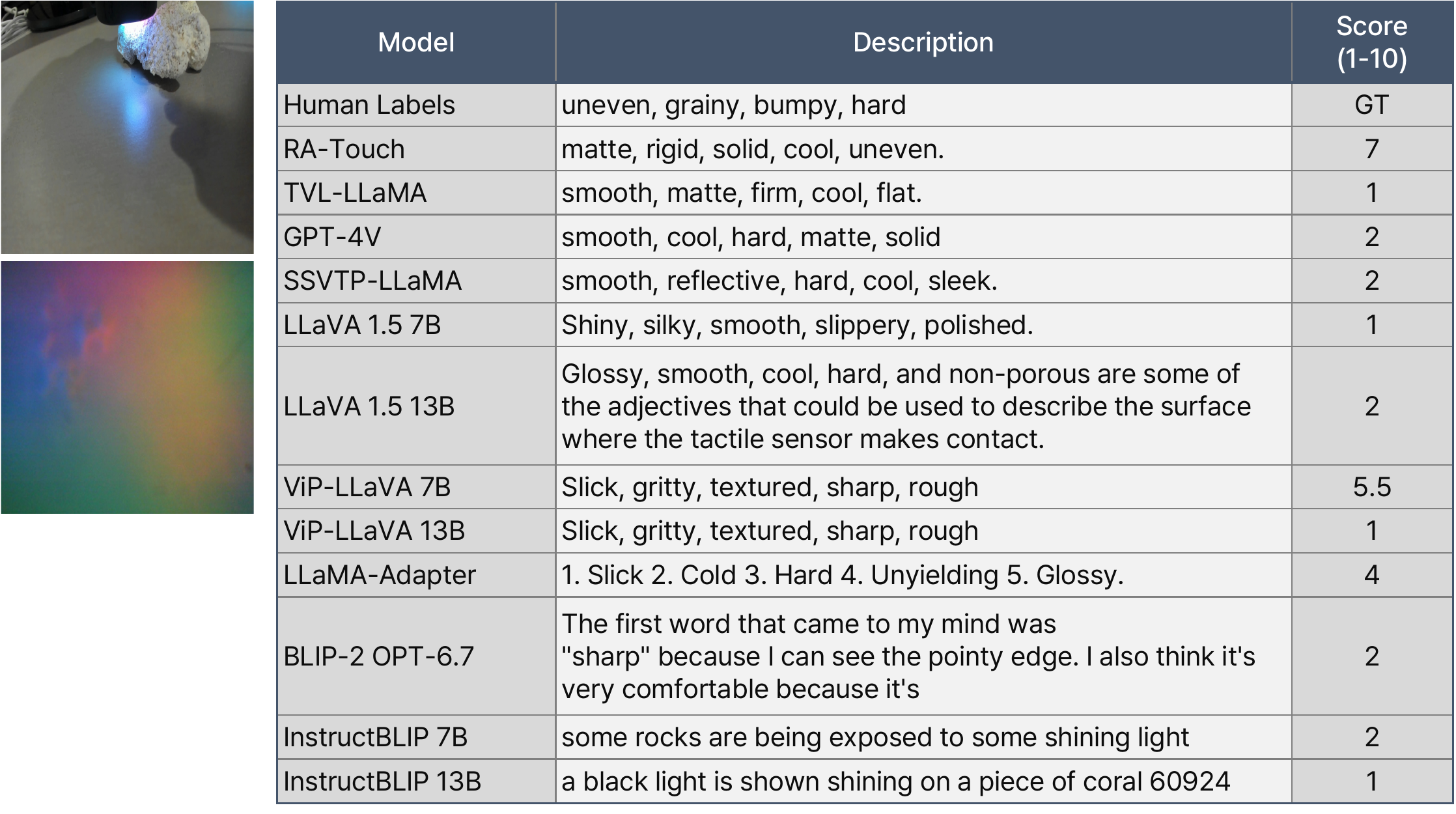}
    \end{subfigure}

    \vspace{8mm}

    \begin{subfigure}[b]{0.49\linewidth}
        \includegraphics[width=\linewidth]{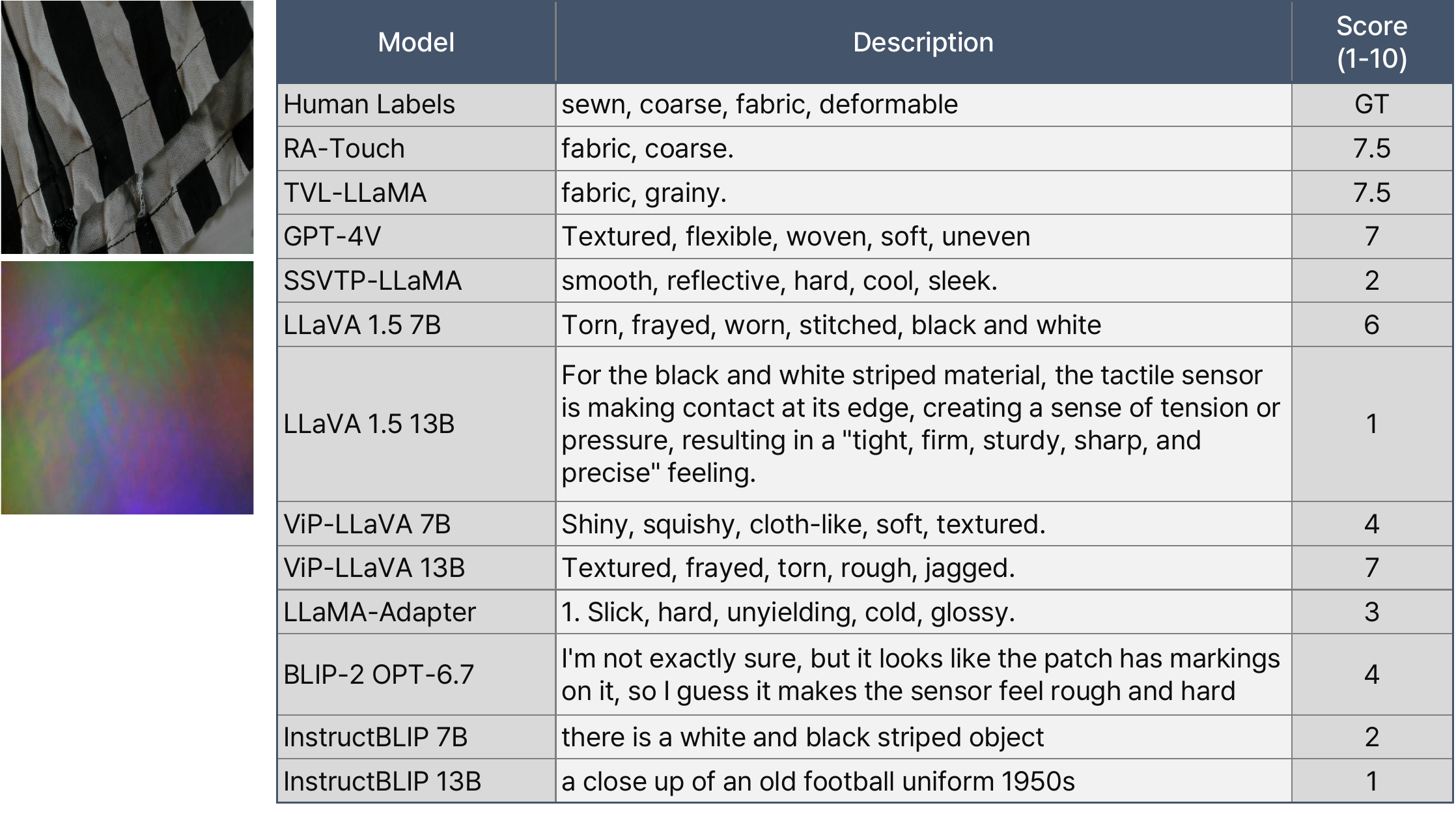}
    \end{subfigure}
    \hfill
    \begin{subfigure}[b]{0.49\linewidth}
        \includegraphics[width=\linewidth]{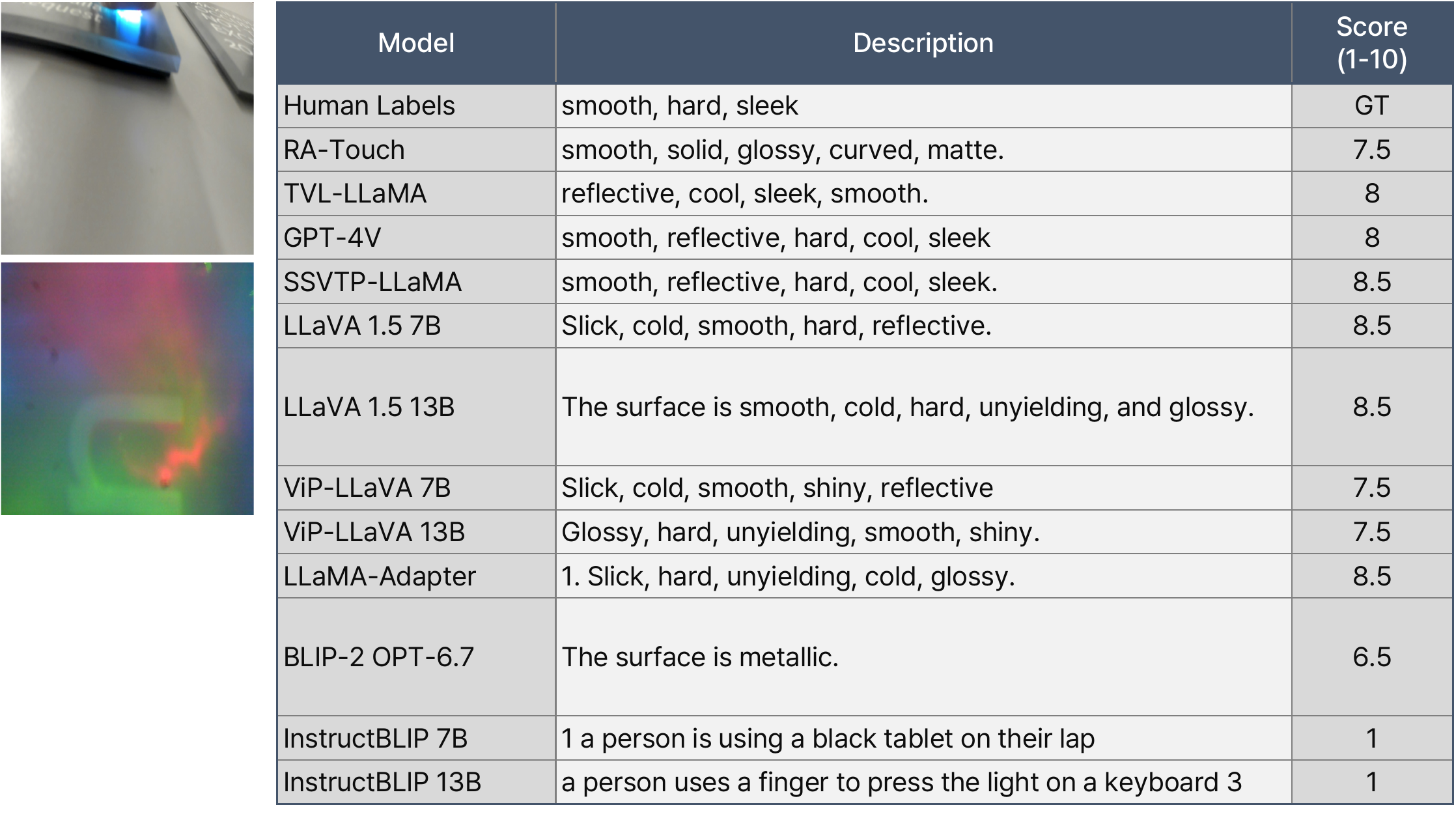}
    \end{subfigure}

    \caption{Qualitative comparison with various VLMs.}
    \label{fig:generation_examples}
\end{figure*}

\section{Generation Examples}
We present visuo-tactile examples from the TVL dataset~\cite{Fu2024ATV}, alongside descriptions generated by various models, including \ours, TVL-LLaMA, and other vision-language baselines as shown in Figure~\ref{fig:generation_examples}. Many samples overlap with those used in TVL-LLaMA to enable fair comparison. Human-labeled descriptions are used as ground truth for tactile attributes.
Overall, our method consistently produces more accurate and detailed captions that capture surface properties such as roughness, softness, firmness, or grain. In contrast, baseline models tend to focus on visual features like color or gloss, often generating vague or object-centric descriptions. Some models fail to attend to the tactile aspect and instead misinterpret contextual or semantic elements unrelated to texture.
These trends underscore the importance of tactile-aware retrieval and integration in producing grounded and semantically meaningful tactile descriptions. The results support the effectiveness of our framework in aligning language outputs with tactile perception.

\section{Discussion \& Future Work}
RA-Touch achieves the state-of-the-art performance on the tactile perception task, demonstrating effective tactile recognition even in data-scarce tactile environments. Using a retrieval-augmented approach, the model was able to incorporate a wide range of vision-language scenarios, highlighting a new direction for tactile perception and showing the potential for broader application to various downstream tasks involving tactile data. We also observed that the quality of external knowledge can significantly influence performance. In light of this, we constructed a new vision-language dataset with texture-centric captions, ImageNet-T, which may provide more suitable supervision for learning tactile-relevant representations. Despite these encouraging results, several technical aspects may benefit from further exploration. Our method uses encoders with a visual-centric bias, often prioritizing background over task-relevant cues, which can degrade retrieval when key content is ambiguous. One possible direction is to explore adaptive mechanisms that extract visual features conditioned on the given tactile input, enabling the model to focus on contextually relevant regions and better align with tactile semantics.

\end{document}